\pdfoutput=1 
\documentclass[twoside,11pt]{article}  

\usepackage{jair,theapa,rawfonts}

\usepackage[T1]{fontenc}    
\usepackage{url}            
\usepackage{booktabs}       
\usepackage{amsfonts}       
\usepackage{nicefrac}       
\usepackage{microtype}      
\usepackage{xcolor}         

\usepackage{amsmath}
\usepackage{amssymb}
\usepackage{bbm}
\usepackage{bm}
\usepackage{cancel}
\usepackage{mathtools}

\usepackage{ascmac}  
\usepackage{algorithm}
\usepackage{algpseudocode}

\usepackage{fancyhdr}
\usepackage{tcolorbox}
\usepackage{tikz}

\usepackage{booktabs}
\usepackage{comment}
\usepackage{lastpage}
\usepackage{listings}
\usepackage{multibib}
\usepackage{multirow}
\usepackage{adjustbox}
\usepackage[super]{nth}
\usepackage[caption=false]{subfig}

\tcbuselibrary{breakable, skins, theorems}

\algtext*{EndFor}
\algtext*{EndWhile}
\algtext*{EndIf}
\algtext*{EndProcedure}
\algtext*{EndFunction}
\algnewcommand{\LineComment}[1]{\State \(\triangleright\) #1}

\newcites{appx}{References}

\newif\ifunderreview
\newif\ifsubmission
\newif\ifappendix

\underreviewfalse

\submissiontrue

\appendixtrue

\ifsubmission
\newcommand{\todo}[1]{}
\newcommand{\replace}[2]{}
\newcommand{\sw}[1]{}

\else
\newcommand{\todo}[1]{\textbf{\textcolor{red}{[TODO: #1]}}}

\newcommand{\replace}[2]{\textbf{\textcolor{red}{[del: \cancel{#1}]}}\textbf{\textcolor{blue}{[new: #2]}}}

\newcommand{\sw}[1]{\textbf{\textcolor{cyan}{[SW: #1]}}}

\usepackage[inline]{showlabels}

\fi

\makeatletter
\newcommand{\customlabel}[2]{%
  \protected@write \@auxout {}{\string \newlabel {#1}{{#2}{\thepage}{#2}{#1}{}} }%
  \hypertarget{#1}{}
}
\makeatother

\newcommand{\argmin}{\mathop{\mathrm{argmin}}}
\newcommand{\argmax}{\mathop{\mathrm{argmax}}}

\newcommand{\ceil}[1]{\lceil #1 \rceil}

\newcommand{\blue}[1]{\textcolor{blue}{#1}}

\newcommand{\green}[1]{\textcolor{green}{#1}}

\newcommand{\purple}[1]{\textcolor{purple}{#1}}
\newcommand{\red}[1]{\textcolor{red}{#1}}

\newcommand{\violet}[1]{\textcolor{violet}{#1}}

\newcommand{\rank}{\stackrel{\mathrm{rank}}{\simeq}}

\newcommand{\xv}{\boldsymbol{x}}

\newcommand{\thetav}{\boldsymbol{\theta}}

\newcommand{\xopt}{\xv_{\mathrm{opt}}}

\newcommand{\EI}{\mathrm{EI}}
\newcommand{\prob}{\mathbb{P}}

\newcommand{\X}{\mathcal{X}}

\newcommand{\D}{\mathcal{D}}
\newcommand{\Dl}{\mathcal{D}^{(l)}}
\newcommand{\Dg}{\mathcal{D}^{(g)}}
\newcommand{\Nl}{N^{(l)}}
\newcommand{\Ng}{N^{(g)}}

\newcommand{\bl}{b^{(l)}}
\newcommand{\bg}{b^{(g)}}
\newcommand{\wl}{w^{(l)}_0}
\newcommand{\wg}{w^{(g)}_0}

\newcommand{\xnull}{x_{\mathrm{null}}}
\newcommand{\Rnull}{\mathbb{R}_{\mathrm{null}}}

\newcommand{\bmagic}{b_{\mathrm{magic}}}

\newcommand{\ok}{\green{$\bigcirc$}}
\newcommand{\soso}{\blue{$\triangle$}}
\newcommand{\bad}{\violet{$\times$}}

\DeclareFixedFont{\ttb}{T1}{txtt}{bx}{n}{12} 
\DeclareFixedFont{\ttm}{T1}{txtt}{m}{n}{12}  

\definecolor{deepblue}{rgb}{0,0,0.5}
\definecolor{deepred}{rgb}{0.6,0,0}
\definecolor{deepgreen}{rgb}{0,0.5,0}

\newcommand{\ackley}{
  \exp(1) + 20\biggl(
  1 - \exp\biggl(
  -\frac{1}{5} \sqrt{\frac{1}{D}\sum_{d=1}^D x_d^2}
  \biggr)
  \biggr)
  - \exp\biggl(
  \frac{1}{D}\sum_{d=1}^D \cos 2\pi x_d
  \biggr)
}
\newcommand{\griewank}{
  1 + \frac{1}{4000} \sum_{d=1}^D x_d^2
  - \prod_{d=1}^D \cos \frac{x_d}{\sqrt{d}}
}
\newcommand{\ktablet}{
  \sum_{d=1}^K x_d^2 +
  \sum_{d=K+1}^{D} (100 x_d)^2
}
\newcommand{\levy}{
  \sin^2 \pi w_1 +
  \sum_{d=1}^{D-1}
  (w_d - 1)^2
  (1 + 10 \sin^2 (\pi w_d + 1))
  + (w_D - 1)^2
  (1 + \sin^2 2\pi w_D)
}
\newcommand{\perm}{
  \sum_{d_1=1}^D\biggl(
  \sum_{d_2=1}^D
  (d_2 + 1) (x_{d_2}^{d_1} - \frac{1}{{d_2}^{d_1}})
  \biggr)^2
}
\newcommand{\rastrigin}{
  10D + \sum_{d=1}^D (x_d^2 - 10 \cos 2\pi x_d)
}
\newcommand{\rosenbrock}{
  \sum_{d=1}^{D-1}\biggl(
  100 (x_{d+1} - x_d^2)^2
  + (x_d - 1)^2
  \biggr)
}
\newcommand{\schwefel}{
  -\sum_{d=1}^D x_d \sin \sqrt{|x_d|}
}
\newcommand{\sphere}{
  \sum_{d=1}^D x_d^2
}
\newcommand{\styblinski}{
  \frac{1}{2}\sum_{d=1}^D (x_d^4 - 16x_d^2 + 5x_d)
}
\newcommand{\weightedsphere}{
  \sum_{d=1}^D d x_d^2
}
\newcommand{\xinsheyang}{
\sum_{d_1=1}^D |x_{d_1}|
\exp\biggl(
-\sum_{d_2=1}^D \sin x_{d_2}^2
\biggr)
}

\begin{document}
\ShortHeadings{A Tutorial of Tree-Structured Parzen Estimator}
{Watanabe}
\firstpageno{1}

\title{
  Tree-Structured Parzen Estimator:
  Understanding \\
  Its Algorithm Components and Their Roles
  \\ for Better Empirical Performance
}


\author{\name Shuhei Watanabe~\thanks{This work was done at the University of Freiburg.} \email shuheiwatanabe@preferred.jp \\
  \addr Preferred Networks Inc., Japan \\
}

\maketitle

\begin{abstract}
  Recent scientific advances require complex experiment design, necessitating the meticulous tuning of many experiment parameters.
  Tree-structured Parzen estimator (TPE) is a widely used Bayesian optimization method in recent parameter tuning frameworks such as Hyperopt and Optuna.
  Despite its popularity, the roles of each control parameter in TPE and the algorithm intuition have not been discussed so far.
  The goal of this paper is to identify the roles of each control parameter and their impacts on parameter tuning based on the ablation studies using diverse benchmark datasets.
  The recommended setting concluded from the ablation studies is demonstrated to improve the performance of TPE.
  Our TPE implementation used in this paper is available at \url{https://github.com/nabenabe0928/tpe/tree/single-opt}.
  \textcolor{magenta}{OptunaHub~\cite{ozaki2025optunahub} now provides our standalone TPE implementation~\footnote{The package page: \url{https://hub.optuna.org/samplers/tpe_tutorial/}}.}
\end{abstract}

\section{Introduction}

Recent scientific advances have seen the possibility of black-box optimization (BBO) in research fields such as drug discovery~\shortcite{schneider2020rethinking}, material discovery~\shortcite{xue2016accelerated,li2017rapid,vahid2018new}, financial applications~\shortcite{gonzalvez2019financial}, and hyperparameter optimization (HPO) of machine learning algorithms~\shortcite{loshchilov2016cma,chen2018bayesian,feurer2019hyperparameter}.
This trend has spurred the development of sample-efficient parameter-tuning frameworks such as Optuna~\shortcite{akiba2019optuna}, Ray~\shortcite{liaw2018tune}, BoTorch~\shortcite{balandat2020botorch}, and Hyperopt~\shortcite{bergstra2011algorithms,bergstra2013making,bergstra2013hyperopt,bergstra2015hyperopt}, enabling researchers to make significant strides in these domains.

Tree-structured Parzen estimator (TPE) is a widely used Bayesian optimization (BO) method in these frameworks and it has achieved various outstanding performances so far.
For example, TPE played a pivotal role for HPO of deep learning models in winning Kaggle competitions~\shortcite{open-images-2019-object-detection,happy-whale-and-dolphin} and \shortciteA{watanabe2023speeding} won the AutoML 2022 competition on ``Multiobjective Hyperparameter Optimization for Transformers'' using TPE.
Furthermore, TPE has been extended to multi-fidelity~\shortcite{falkner2018bohb}, multi-objective~\shortcite{ozaki2020multiobjective,ozaki2022multiobjective}, meta-learning~\shortcite{watanabe2022multi,watanabe2023speeding}, and constrained~\shortcite{watanabe2022ctpe,watanabe2023ctpe} settings to tackle diverse scenarios.
Despite its versatility, its algorithm intuition and the roles of each control parameter have not been discussed so far.
Therefore, we describe the algorithm intuition and empirically present the roles of each control parameter.
The rest of this paper is structured as follows:
\begin{enumerate}
  \vspace{-1mm}
  \item \textbf{Background}: explains the knowledge required for this paper,
  \vspace{-1mm}
  \item \textbf{Algorithm Details of TPE}: describes the TPE algorithm and empirically presents the roles of each control parameter, and
  \vspace{-1mm}
  \item \textbf{Ablation Study}: performs the ablation study of the control parameters in the original TPE and investigates the enhancement in the bandwidth selection on diverse benchmark datasets.
  \vspace{-1mm}
\end{enumerate}
The recommended setting drawn from the analysis is compared against recent baseline methods.
This paper narrows down our scope solely to single-objective optimization problems for simplicity.
We defer the details of the extensions and the applications of TPE to Appendix~\ref{appx:related-work:section} and some general tips for HPO to Appendix~\ref{appx:general-advice:section}.

\section{Background}
This section describes the knowledge required to read through this paper.

\subsection{Notations}
We first define notations in this paper.
\begin{itemize}
  \vspace{-1mm}
  \item $\X_d \subseteq \mathbb{R}$ (for $d = 1,\dots,D$), a domain of the $d$-th (transformed) hyperparameter,
  \vspace{-1mm}
  \item $\xv \in \X \coloneqq \X_1 \times \X_2 \times \dots \times \X_D \subseteq \mathbb{R}^D$, a (transformed) hyperparameter configuration,
  \vspace{-1mm}
  \item $y = f(\xv) + \varepsilon$, an observation of the objective function $f: \X \rightarrow \mathbb{R}$ with a noise $\varepsilon$,
  \vspace{-1mm}
  \item $\D \coloneqq \{(\xv_n, y_n)\}_{n=1}^N$, a set of observations (the size $N \coloneqq |\D|$),
  \vspace{-1mm}
  \item $\Dl, \Dg$, a better group and a worse group in $\D$ (the sizes $\Nl \coloneqq |\Dl|, \Ng \coloneqq |\Dg|$),
  \vspace{-6mm}
  \item $\gamma \in (0, 1]$, a top quantile used for the better group $\Dl$,
  \vspace{-1mm}
  \item $y^\gamma \in \mathbb{R}$, the top-$\gamma$ quantile objective value in $\D$,
  \vspace{-1mm}
  \item $p(\xv | \Dl), p(\xv | \Dg)$, the probability density functions (PDFs) of the better group and the worse group built by kernel density estimators (KDEs),
  \vspace{-1mm}
  \item $r(\xv |\D) \coloneqq p(\xv | \Dl) / p(\xv | \Dg)$, the density ratio (equivalent to acquisition function) used to judge the promise of a hyperparameter configuration,
  \vspace{-1mm}
  \item $k: \X \times \X \rightarrow \mathbb{R}_{\geq 0}$, a kernel function with bandwidth $b \in \mathbb{R}_+$ that changes based on a provided dataset,
  \vspace{-1mm}
  \item $\bl, \bg \in \mathbb{R}_+$, bandwidth (a control parameter) for the kernel function based on $\Dl$ and $\Dg$,
  \vspace{-1mm}
  \item $w_n \in [0, 1]$ (for $n = 1, \dots, N$), a weight for each basis in KDEs,
  \vspace{-1mm}
  \item $\rank$, the order isomorphism between left hand side and right hand side and $\phi(\xv) \rank \psi(\xv)$ means $\phi(\xv_1) < \phi(\xv_2) \Leftrightarrow \psi(\xv_1) < \psi(\xv_2)$.
  \vspace{-1mm}
\end{itemize}
Note that ``transformed'' implies that some parameters might be preprocessed by such as log transformation or logit transformation,
and the notations $l$ (\emph{lower} or better) and $g$ (\emph{greater} or worse) come from the original paper~\shortcite{bergstra2011algorithms}.

\subsection{Bayesian Optimization}

Bayesian optimization (BO)~\footnote{
  We encourage readers to check recent surveys~\shortcite{brochu2010tutorial,shahriari2016taking,garnett2022bayesian}.
} aims to minimize the objective function $f(\xv)$ as follows:
\begin{equation}
  \begin{aligned}
    \xopt \in \argmin_{\xv \in \X} f(\xv).
  \end{aligned}
  \label{main:background:eq:hpo-formulation}
\end{equation}
Note that this paper consistently considers
\textbf{minimization} problems.
For example, hyperparameter optimization (HPO)
of machine learning algorithms aims
to find an optimal hyperparameter configuration $\xopt$
(e.g., learning rate, dropout rate, and the number of layers)
that exhibits the best performance
(e.g., the error rate in classification tasks, and mean squared error in regression tasks).
BO iteratively searches for $\xopt$ using the so-called
acquisition function to trade off the degree of exploration
and exploitation.
Roughly speaking, while exploitation searches near promising observations, exploration searches unseen regions.
A common choice for the acquisition function is
the following expected improvement (\texttt{EI})~\shortcite{jones1998efficient}:
\begin{equation}
  \begin{aligned}
    \EI_{y^\star}[\xv|\D] \coloneqq \int_{-\infty}^{y^\star}
    (y^\star - y)p(y | \xv, \D)dy.
  \end{aligned}
  \label{main:background:eq:expected-improvement}
\end{equation}
Another choice is the probability of improvement (\texttt{PI})~\shortcite{kushner1964new}:
\begin{equation}
  \begin{aligned}
    \prob(y \leq y^\star | \xv, \D)
    \coloneqq \int_{-\infty}^{y^\star} p(y | \xv, \D)dy.
  \end{aligned}
  \label{main:background:eq:probability-of-improvement}
\end{equation}
Note that $y^\star$ is a control parameter specified by algorithms or users.
In principle, \texttt{PI} is exploitative and \texttt{EI} is explorative.
TPE is an exploitative BO method because the acquisition function of TPE is equivalent to \texttt{PI} as shown by \shortciteA{watanabe2022ctpe,watanabe2023ctpe,song2022general}~\footnote{
  \shortciteA{bergstra2011algorithms} originally state that the acquisition function of TPE is \texttt{EI}, but \texttt{EI} is equivalent to \texttt{PI} in the TPE formulation.
}.
Owing to the exploitative nature, TPE is inclined to search locally.
The posterior $p(y | \xv, \D)$ computation varies depending on the BO methods.
Although a typical choice is Gaussian process regression~\shortcite{williams2006gaussian}, practical methods such as SMAC~\shortcite{hutter2011sequential} and TPE use random forests and KDEs, respectively.
The next section describes the posterior modeling $p(y | \xv, D)$ for TPE by KDEs.

\begin{figure}[t]
  \centering
  \includegraphics[width=0.98\textwidth]{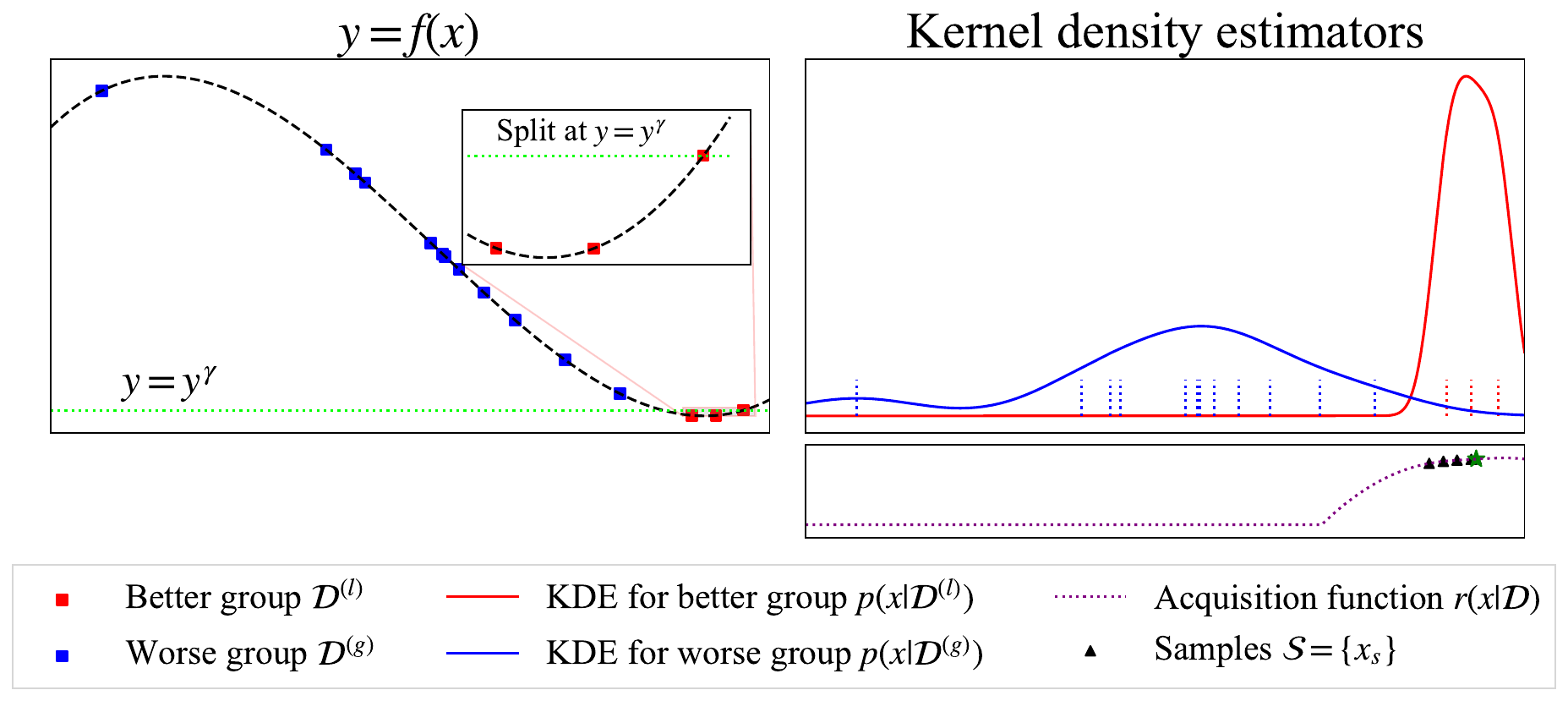}
  \vspace{-3mm}
  \caption{
    The conceptual visualization of TPE.
    \textbf{Left}:
    the objective function $y = f(\xv)$ (black dashed line) and its observations $\D$.
    The magnified figure shows the boundary $y = y^\gamma$ (\green{green dotted line}) of
    $\Dl$ (\red{red squares}) and $\Dg$ (\blue{blue squares}).
    \textbf{Top right}:
    the KDEs built by $\Dl$ (\red{red solid line}) and $\Dg$ (\blue{blue solid line}).
    \textbf{Bottom right}:
    the density ratio $p(\xv |\Dl) / p(\xv | \Dg)$ (\purple{purple dotted line}) used for the acquisition function.
    We pick the configuration with the best acquisition function value
    (\green{green star})
    in the samples (black triangles) from $p(\xv |\Dl)$.
  }
  \vspace{-3mm}
  \label{main:background:fig:tpe-conceptual}
\end{figure}

\begin{algorithm}[tb]
  \caption{Tree-structured Parzen estimator (TPE)}
  \label{main:background:alg:tpe-algo}
  \begin{algorithmic}[1]
    \Statex{$N_{\mathrm{init}}$
    (The number of initial configurations, \texttt{n\_startup\_trials} in Optuna),
    $N_{s}$
    (The number of candidates to consider in the optimization of the acquisition function,
    \texttt{n\_ei\_candidates} in Optuna),
    $\Gamma$
    (A function to compute the top quantile $\gamma$, \texttt{gamma} in Optuna),
    $W$
    (A function to compute weights $\{w_n\}_{n=0}^{N+1}$, \texttt{weights} in Optuna),
    $k$
    (A kernel function),
    $B$
    (A function to compute a bandwidth $b$ for $k$).
    }
    \State{$\D \leftarrow \emptyset$}
    \For{$n = 1, 2, \dots, N_{\mathrm{init}}$}
    \Comment{Initialization}
    \State{Randomly pick $\xv_n$}
    \State{$y_n \coloneqq f(\xv_n) + \epsilon_n$}
    \Comment{Evaluate the (expensive) objective function}
    \State{$\D \leftarrow \D \cup \{(\xv_n, y_n)\}$}
    \EndFor
    \While{Budget is left}
    \label{main:background:line:tpe-algo-start}
    \State{Compute $\gamma \leftarrow \Gamma(N)$ with $N \coloneqq |\D|$}
    \Comment{Section~\ref{main:algorithm-detail:section:gamma}~(Splitting algorithm)}
    \State{Split $\D$ into $\Dl$ and $\Dg$}
    \State{Compute $\{w_n\}_{n=0}^{N+1} \leftarrow W(\D)$}
    \Comment{See Section~\ref{main:algorithm-detail:section:weight}~(Weighting algorithm)}
    \State{Compute $\bl \leftarrow B(\Dl), \bg \leftarrow B(\Dg)$}
    \Comment{Section~\ref{main:algorithm-detail:section:bandwidth}~(Bandwidth selection)}
    \State{Build $p(\xv | \Dl), p(\xv|\Dg)$ based on Eq.~(\ref{main:background:eq:kernel-density-estimation})}
    \Comment{Use $\{w_n\}_{n=0}^{N+1}$ and $\bl, \bg$}
    \State{Sample $\mathcal{S} \coloneqq \{\xv_s\}_{s=1}^{N_s} \sim p(\xv|\Dl)$}
    \State{Pick $\xv_{N + 1} \coloneqq \xv^\star \in \argmax_{\xv \in \mathcal{S}} r(\xv |\D)$}
    \label{main:background:line:tpe-algo-greedy}
    \Comment{The evaluations by the acquisition function}
    \State{$y_{N + 1} \coloneqq f(\xv_{N + 1}) + \epsilon_{N+1}$}
    \Comment{Evaluate the (expensive) objective function}
    \State{$\D \leftarrow \D \cup \{\xv_{N + 1}, y_{N + 1}\}$}
    \EndWhile
    \label{main:background:line:tpe-algo-end}
  \end{algorithmic}
\end{algorithm}

\subsection{Tree-Structured Parzen Estimator}
\label{main:background:section:tpe}
TPE is a variant of BO methods first invented by \shortciteA{bergstra2011algorithms}.
The name comes from the method being able to handle a tree-structured search space, and using Parzen estimators, aka kernel density estimators (KDEs).
Notice that a tree-structured search space is a search space that includes some conditional parameters~\footnote{
  For example, when we optimize the dropout rates at each layer in an $L$-layered neural network where $L \in \{2,3\}$, the dropout rate at the third layer does not exist for $L = 2$.
  Therefore, we call the dropout rate at the third layer a \emph{conditional parameter}.
  Note that this paper tests TPE only on non tree-structured spaces.
}.
TPE models the posterior $p(y | \xv, \D)$ using the following assumption:
\begin{eqnarray}
  p(\xv|y, \D) \coloneqq \left\{
  \begin{array}{ll}
     p(\xv|\Dl) & (y \leq y^\gamma) \\
     p(\xv|\Dg) & (y > y^\gamma)
  \end{array}
  \right.,
  \label{main:background:eq:tpe-transform-assumption}
\end{eqnarray}
where the top-quantile $\gamma$ is computed at each iteration based on the number of observations $N (=|\D|)$ (see Section~\ref{main:algorithm-detail:section:gamma}), and $y^\gamma$ is the top-$\gamma$-quantile objective value in the set of observations $\D$; see Figure~\ref{main:background:fig:tpe-conceptual} for the intuition.
For simplicity, we assume that $\D$ is already sorted by $y_n$ such that $y_1 \leq y_2 \leq \dots \leq y_N$.
Then the better group and the worse group are obtained as $\Dl = \{(\xv_n, y_n)\}_{n=1}^{\Nl}$ and $\Dg = \{(\xv_n, y_n)\}_{n=\Nl+1}^{N}$ where $\Nl = \ceil{\gamma N}$.
The KDEs in Eq.~(\ref{main:background:eq:tpe-transform-assumption}) are estimated via:
\begin{equation}
\begin{aligned}
  p(\xv|\Dl) &= w_0^{(l)} p_0(\xv) + \sum_{n=1}^{\Nl} w_n k(\xv, \xv_n | \bl), \\
  p(\xv|\Dg) &= w_0^{(g)} p_0(\xv) + \sum_{n=\Nl+1}^{N} w_n k(\xv, \xv_n | \bg) \\
\end{aligned}
\label{main:background:eq:kernel-density-estimation}
\end{equation}
where the weights $\{w_n\}_{n=1}^N$ are determined at each iteration (see Section~\ref{main:algorithm-detail:section:weight}), $k$ is a kernel function (see Section~\ref{main:algorithm-detail:section:kernel}), $\bl, \bg \in \mathbb{R}_+$ are the bandwidth (see Section~\ref{main:algorithm-detail:section:bandwidth}) and $p_0$ is non-informative prior (see Section~\ref{main:algorithm-detail:section:prior}).
Note that the summations of weights are $1$, meaning that $w_0^{(l)} + \sum_{n=1}^{\Nl} w_n = 1$ and $w_0^{(g)} + \sum_{n=\Nl+1}^{N} w_n = 1$ hold.
Using the assumption in Eq.~(\ref{main:background:eq:tpe-transform-assumption}), we obtain the following acquisition function:
\begin{equation}
\begin{aligned}
  \prob(y \leq y^\gamma |\xv, \D) \rank r(\xv | \D) \coloneqq
  \frac{p(\xv|\Dl)}{p(\xv|\Dg)}.
\end{aligned}
\label{main:background:eq:density-ratio}
\end{equation}
The intermediate process is provided in Appendix~\ref{appx:theoretical-details-of-tpe:section}.
The main routine of the TPE algorithm lies in Lines~\ref{main:background:line:tpe-algo-start}--\ref{main:background:line:tpe-algo-end} of Algorithm~\ref{main:background:alg:tpe-algo}.
Each iteration repeats the TPE algorithm parameter calculations and the selection of the next configuration based on the acquisition function.
The selection candidates are sampled from the KDE built by the better group $p(\xv|\Dl)$.
Although the global convergence of TPE is guaranteed when the $\epsilon$-greedy algorithm is used in Line~\ref{main:background:line:tpe-algo-greedy} as shown by \shortciteA{watanabe2023speeding}, this paper focuses on the greedy algorithm due to our assumption of a restrictive budget ($\sim 200$ evaluations).

\section{Algorithm Details of Tree-Structured Parzen Estimator}
This section describes each component of TPE and elucidates the roles of each control parameter.
More specifically, we highlight the effects of each control parameter on the trade-off between exploitation and exploration.
Roughly speaking, exploitation searches near promising observations and exploration searches unseen regions.
Each subsection title accompanies (\texttt{arg\_name}), which refers to the corresponding argument's name of \texttt{TPESampler} in Optuna.
Each visualization uses the default setting except \texttt{multivariate=True} of Optuna v4.0.0 if not specified.
Table~\ref{main:algorithm-detail:tab:component-comparison} presents the implementational differences in TPE variants.

\begin{figure}[t]
  \centering
  \includegraphics[width=0.98\textwidth]{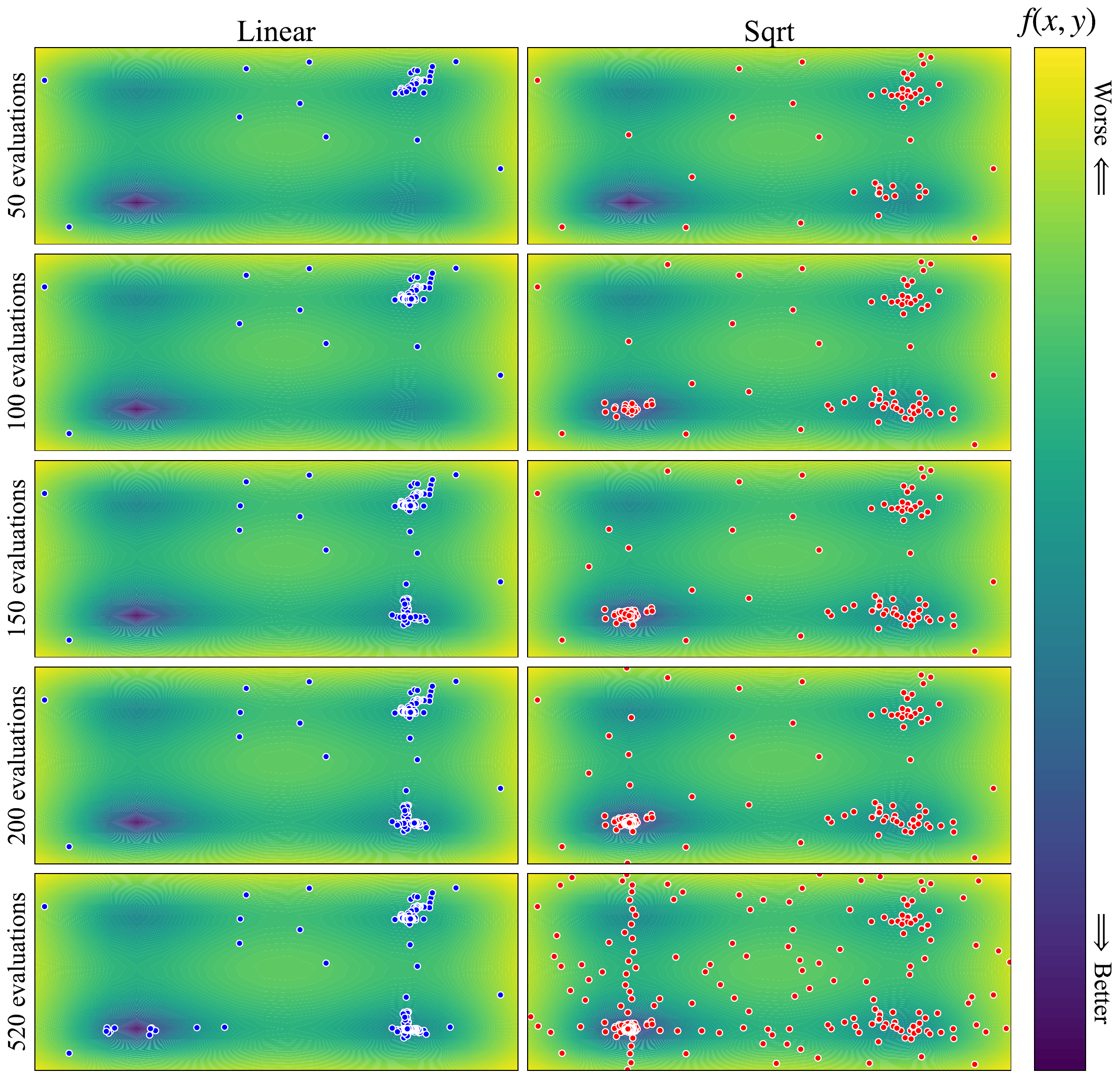}
  \caption{
    The optimizations of the Styblinski function using the splitting algorithm \texttt{linear} and \texttt{sqrt}.
    The \red{red} and \blue{blue} dots show the observations till each ``X evaluations''.
    The lower left blue shade in each figure is the optimal point
    and this area should be found with as few observations as possible.
    \textbf{Left column}: the optimization using \texttt{linear}.
    The optimal area is found with around $500$ evaluations owing to strong exploitation.
    \textbf{Right column}: the optimization using \texttt{sqrt}.
    The optimal area is found with around $100$ evaluations thanks to exploration.
    Although there is no observation near the optimal area for both methods at $50$ evaluations, \texttt{sqrt} finds the optimal area thanks to its exploration nature.
  }
  \label{main:algorithm-detail:fig:exploration-by-gamma-in-styblinski}
\end{figure}

\begin{figure}[t]
  \centering
  \includegraphics[width=0.98\textwidth]{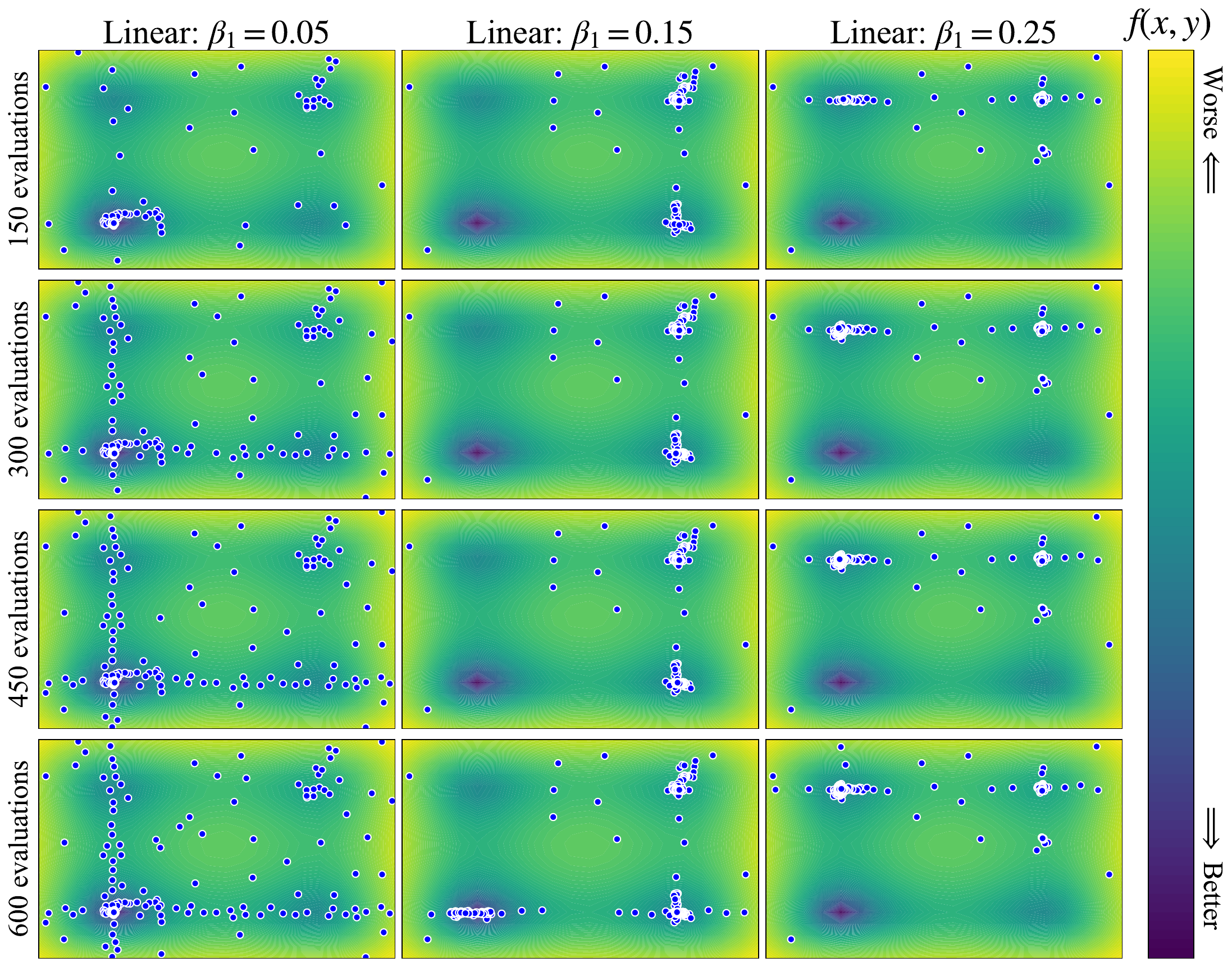}
  \caption{
    The optimizations of the Styblinski function using the splitting algorithm \texttt{linear} with different $\beta_1$ (\textbf{Left column}: $\beta_1 = 0.05$, \textbf{Center column}: $\beta_1 = 0.15$, \textbf{Right column}: $\beta_1 = 0.25$).
    The \blue{blue dots} show the observations till each ``X evaluations''.
    The lower left blue shade in each figure is the optimal point and this area should be found with as few observations as possible.
    We see scattered dots (more explorative) for a small $\beta_1$ and concentrated dots (more exploitative) for a large $\beta_1$.
  }
  \label{main:algorithm-detail:fig:gamma-linear-ablation}
\end{figure}

\begin{figure}[t]
  \centering
  \includegraphics[width=0.98\textwidth]{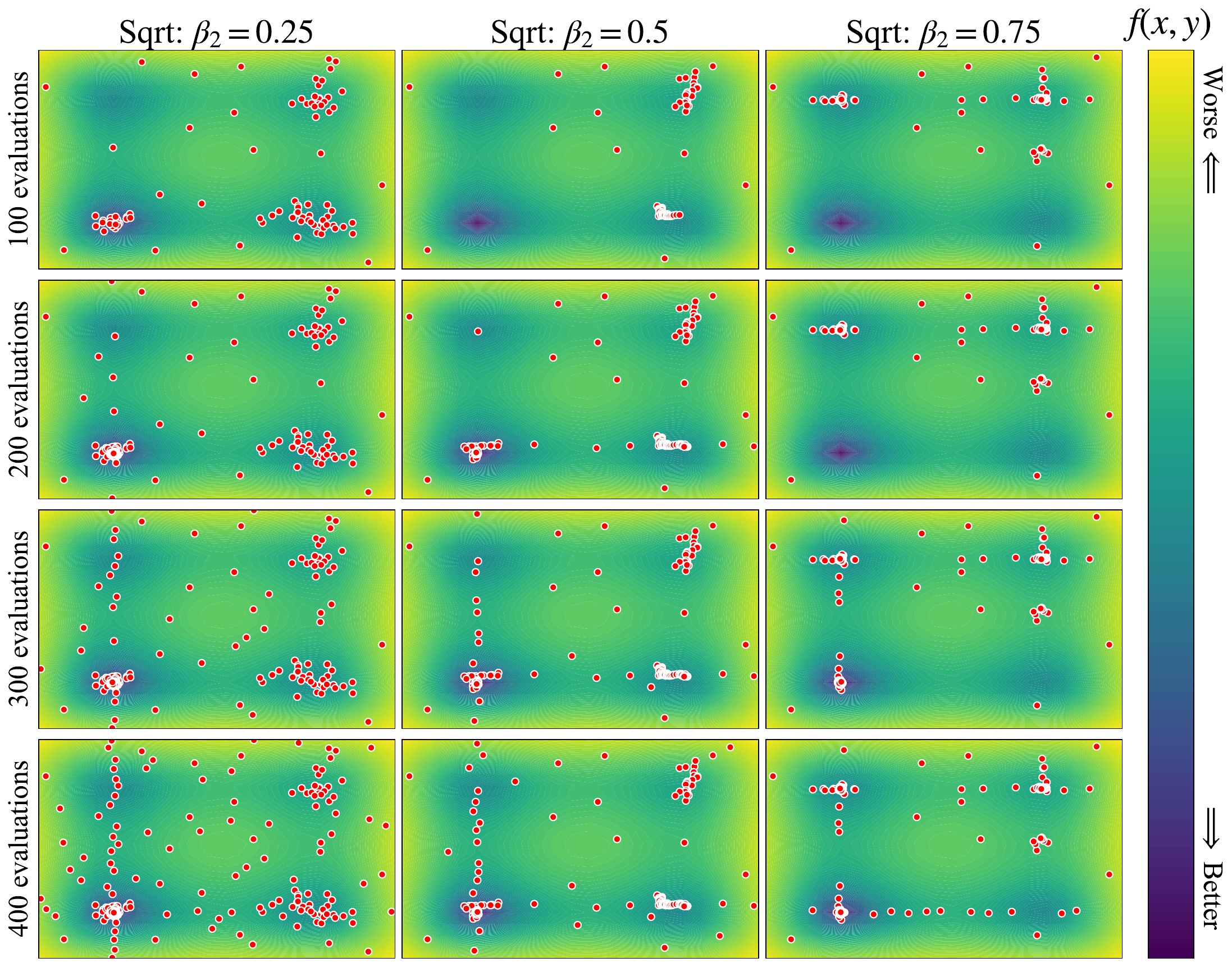}
  \caption{
    The optimizations of the Styblinski function using the splitting algorithm \texttt{sqrt}
    with different $\beta_2$
    (\textbf{Left column}: $\beta_2 = 0.25$, \textbf{Center column}: $\beta_2 = 0.5$, \textbf{Right column}: $\beta_2 = 0.75$).
    The \red{red dots} show the observations till each ``X evaluations''.
    The lower left blue shade in each figure is the optimal point
    and this area should be found with as few observations as possible.
    We see scattered dots (more explorative) for a small $\beta_2$
    and concentrated dots (more exploitative) for a large $\beta_2$.
  }
  \label{main:algorithm-detail:fig:gamma-sqrt-ablation}
\end{figure}

\subsection{Splitting Algorithm (\texttt{gamma})}
\label{main:algorithm-detail:section:gamma}
The splitting algorithm of $\D$ into the better group $\Dl$ and the worse group $\Dg$ is controlled by the quantile $\gamma$ and a function $\Gamma(N)$.
The following two variants of $\Gamma$ are proposed separately by \shortciteA{bergstra2011algorithms} and \shortciteA{bergstra2013making}, respectively:
\begin{itemize}
  \vspace{-1mm}
  \item (\texttt{linear}) $\gamma \coloneqq \Gamma(N) = \beta_1 \in (0, 1]$,
  \vspace{-1mm}
  \item (Square root (\texttt{sqrt})) $\gamma \coloneqq \Gamma(N) = \beta_2 / \sqrt{N}, \beta_2 \in (0, \sqrt{N}]$,
  \vspace{-1mm}
\end{itemize}
where $\beta_1 = 0.15$ and $\beta_2 = 0.25$ are used in the original papers and the number of observations in the better group $\Dl$ is limited to $25$ at maximum.
Figure~\ref{main:algorithm-detail:fig:exploration-by-gamma-in-styblinski}
shows that \texttt{sqrt} promotes more exploration and suppresses exploitation, and Figures~\ref{main:algorithm-detail:fig:gamma-linear-ablation}, \ref{main:algorithm-detail:fig:gamma-sqrt-ablation}
demonstrate smaller $\beta_1$ and $\beta_2$ lead to more exploration and less exploitation.
Small $\beta_1$ and $\beta_2$ are explorative because:
\begin{enumerate}
  \vspace{-1mm}
  \item $p(\xv|\Dl)$ would have a narrower mode that requires few observations for $p(\xv|\Dg)$ to cancel out the contribution from $p(\xv|\Dl)$ in the density ratio $r(\xv|\D)$,
  and thus it takes less time to switch to exploration, and
  \vspace{-1mm}
  \item Since the prior weight $w_0^{(l)}$ becomes larger due to a smaller number of observations in the better group $\Dl$, the prior effect becomes more dominant in $p(\xv|\Dl)$ of Eq.~(\ref{main:background:eq:kernel-density-estimation})
  and it promotes exploration.
  \vspace{-1mm}
\end{enumerate}
On the other hand, when the objective function has multiple modes,
the multimodality in $p(\xv|\Dl)$ allows to explore all modalities,
and large $\beta_1$ or $\beta_2$ does not necessarily lead to poor performance.
It explains why \shortciteA{ozaki2020multiobjective,ozaki2022multiobjective} use \texttt{linear}, which gives a larger $\gamma$,
for multi-objective settings.

\begin{figure}[t]
  \centering
  \includegraphics[width=0.98\textwidth]{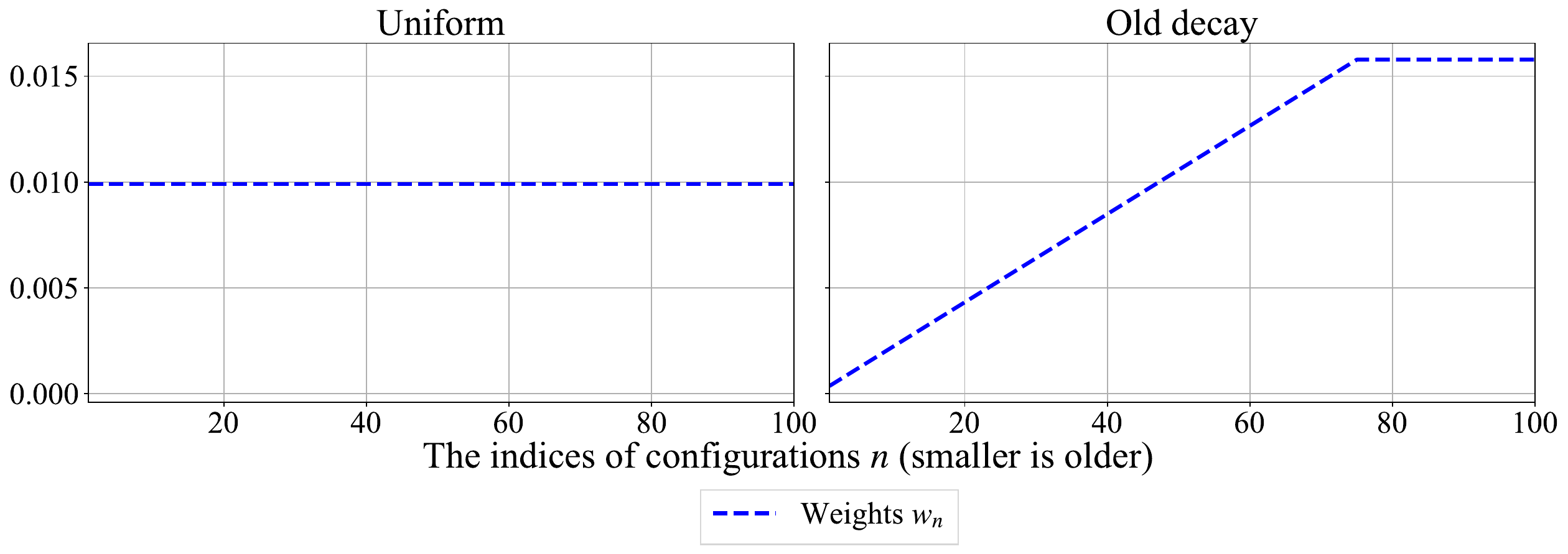}
  \vspace{-3mm}
  \caption{
    The distributions of each weighting algorithm when using $\Ng = 100$.
    \textbf{Left}: the weight distribution for the uniform.
    \textbf{Right}: the weight distribution for the old decay.
    Older observations get lower weights
    and the latest 25 observations get the uniform weight.
  }
  \vspace{-3mm}
  \label{main:algorithm-detail:fig:weight-distribution}
\end{figure}

\begin{table}
  \begin{center}
    \caption{
      The advantages and disadvantages of each weighting algorithm.
    }
    \vspace{2mm}
    \label{main:algorithm-detail:tab:benefits-drawbacks-of-weighting-strategy}
    \makebox[1 \textwidth][c]{       
    \resizebox{1 \textwidth}{!}{   
    \begin{tabular}{lll}
      \toprule
      Weighting algorithm & Advantages & Disadvantages \\
      \midrule
      Uniform & - Use all observations equally & - Take time to account the recent observations \\
      \vspace{2mm}
      & - Need no careful preprocessing of $y$ & - Not consider the ranking in each group \\
      Old decay & - Take less time to switch to exploration & - Might waste the knowledge from the past \\
      \vspace{2mm}
      & - Need no careful preprocessing of $y$ & - Not consider the ranking in each group \\
      Expected improvement & - Consider the ranking in the better group & - Need careful preprocessing of $y$ \\
      \bottomrule
    \end{tabular}
    }}
  \end{center}
\end{table}

\subsection{Weighting Algorithm (\texttt{weights}, \texttt{prior\_weight})}
\label{main:algorithm-detail:section:weight}
The weighting algorithm $W$ is used to determine the
weights $\{w_n\}_{n=1}^N$ for KDEs.
For simplicity, we denote the prior weights as
$w_0^{(l)} \coloneqq w_0$ and
$w_0^{(g)} \coloneqq w_{N + 1}$,
and we consider only \texttt{prior\_weight=1.0},
which is the default value;
see Section~\ref{main:algorithm-detail:section:prior} for more details
about \texttt{prior\_weight}.
\shortciteA{bergstra2011algorithms} use the following \emph{uniform} weighting algorithm:
\begin{eqnarray}
  w_n \coloneqq \left\{
  \begin{array}{ll}
    \frac{1}{\Nl + 1} & (n = 0, \dots, \Nl)     \\
    \frac{1}{\Ng + 1} & (n = \Nl + 1, \dots, N + 1)
  \end{array}
  \right.,
\end{eqnarray}
In contrast, \shortciteA{bergstra2013making} use the following \emph{old decay} weighting algorithm:
\begin{eqnarray}
  w_n \coloneqq \left\{
  \begin{array}{ll}
    \frac{1}{\Nl + 1}                                   & (n = 0, \dots, \Nl)     \\
    \frac{w_n^\prime}{\sum_{n=\Nl +1}^{N+1} w_n^\prime} & (n = \Nl + 1, \dots, N + 1)
  \end{array}
  \right..
\end{eqnarray}
where $w_n^\prime$ (for $n = \Nl + 1, \dots, N + 1$) is defined as:
\begin{eqnarray}
  w_n^\prime \coloneqq \left\{
  \begin{array}{ll}
    1                                   & (t_n > \Ng + 1 - T_{\mathrm{old}}) \\
    \tau(t_n) + \frac{1-\tau(t_n)}{\Ng + 1} & (\mathrm{otherwise})
  \end{array}
  \right..
\end{eqnarray}
Note that $t_n \in \{1,\dots, \Ng + 1\}$ for $n = \Nl + 1, \dots, N + 1$ is the query order of the $n$-th observation
in $\Dg$, which means $t_n = 1$ is the oldest and $t_n = \Ng + 1$ is the youngest, and the decay rate is $\tau(t_n) \coloneqq (t_n - 1) / (\Ng - T_{\mathrm{old}})$ where $T_{\mathrm{old}} = 25$ is used by \shortciteA{bergstra2013making}.
We take $t_{N+1} = 1$ to view the prior as the oldest information.
We visualize the weight distribution in Figure~\ref{main:algorithm-detail:fig:weight-distribution}.
The old decay assigns smaller weights to older observations
as the search should focus on the current region of interest but not on the old regions of interest.
Furthermore, the following computation makes the acquisition function \emph{expected improvement} (\texttt{EI})
more strictly~\shortcite{song2022general}\footnote{This statement is not exact. See Erratum in Appendix~\ref{appx:erratum:section}.}:
\begin{eqnarray}
  w_n \coloneqq \left\{
  \begin{array}{ll}
    \frac{y^\gamma - y_n}{\sum_{n^\prime=1}^{\Nl} (1 + 1/\Nl) (y^\gamma - y_{n^\prime})} & (n = 1, \dots, \Nl) \\
    \frac{1}{\Ng + 1}                                   & (n = \Nl + 1, \dots, N + 1)     \\
  \end{array}
  \right.
  \label{main:algorithm-detail:eq:weighting-expected-improvement}
\end{eqnarray}
where $w^{\prime}_0$ is the mean of $y^\gamma - y_n$ for $n = 1, \dots, \Nl$
and 
\begin{equation}
\begin{aligned}
  w_0 \coloneqq \frac{
    \sum_{n=1}^{\Nl} (y^\gamma - y_n) / \Nl
  }{
    \sum_{n=1}^{\Nl} (1 + 1/\Nl) (y^\gamma - y_n)
  }.
\end{aligned}
\end{equation}
Note that the weighting algorithm used in MOTPE~\shortcite{ozaki2022multiobjective} is proven to be \texttt{EI} by \shortciteA{song2022general} although this fact is not explicitly mentioned by \shortciteA{ozaki2022multiobjective}.\footnote{This statement is not exact. See Erratum in Appendix~\ref{appx:erratum:section}.}

Table~\ref{main:algorithm-detail:tab:benefits-drawbacks-of-weighting-strategy} lists the advantages and disadvantages of each weighting algorithm.
While \texttt{EI} can consider the scale of $y$, $y$ must be carefully preprocessed such as by standardization and log transformation.
Although uniform and old decay should be used when an abundant budget is available, multiple independent runs of TPE are preferred in such cases.

\subsection{Kernel Functions}
\label{main:algorithm-detail:section:kernel}

This section explains the control parameters in the kernel functions.
Notice that this section consistently denotes the kernel function for the $d$-th dimension as $k_d: \X_d \times \X_d \rightarrow \mathbb{R}_{\geq 0}$ and the uniform weight is used for simplicity.

\subsubsection{Kernel for Numerical Parameters}
The Gaussian kernel is used for numerical parameters:
\begin{equation}
\begin{aligned}
  g(x,x_n|b) \coloneqq \frac{1}{\sqrt{2\pi b^2}} \exp{
    \biggl[        
      -\frac{1}{2} \biggl(\frac{x - x_n}{b}\biggr)^2
    \biggr]
  }.
\end{aligned}
\label{main:algorithm-detail:eq:gauss-kernel}
\end{equation}
\shortciteA{bergstra2011algorithms} employ the truncated Gaussian kernel $k_d(x, x_n) \coloneqq g(x, x_n | b) / Z(x_n)$ for $x$ defined on $\X_d \coloneqq [L, R]$ where $Z(x_n) \coloneqq \int_{L}^{R} g(x, x_n | b) dx$ is a normalization constant.
The parameter $b \in \mathbb{R}_+$ in the Gaussian kernel is called \emph{bandwidth}, and \shortciteA{falkner2018bohb} use Scott's rule \shortcite{scott2015multivariate}~(see Appendix~\ref{appx:algorithm-detail:section:scott-bandwidth-selection}) and \shortciteA{bergstra2011algorithms} use a heuristic to determine the bandwidth $b$ as described in Appendix~\ref{appx:algorithm-detail:section:hyperopt-bandwidth-selection}.
This paper uses Scott's rule as a main algorithm to be consistent with the classical KDE basis.
The kernel function for a numerical discrete parameter $x \in \{L, L + q, \dots, R\}$ is computed as:
\begin{equation}
\begin{aligned}
  k_d(x, x_n) \coloneqq \frac{1}{Z(x_n)}\int_{x - q/2}^{x + q/2}
  g(x^\prime, x_n | b) dx^\prime
\end{aligned}
\end{equation}
where we defined $R \coloneqq L + (K - 1)q$, $q \in \mathbb{R}$, and $K \in \mathbb{Z}_+$.
The normalization constant for the discrete kernel is computed as $Z(x_n) \coloneqq \int_{L - q/2}^{R + q/2} g(x^\prime, x_n | b) dx^\prime$.
In general, large and small bandwidths are explorative and exploitative, respectively, as discussed in Section~\ref{main:algorithm-detail:section:bandwidth}.

\subsubsection{Kernel for Categorical Parameters}
The Aitchison-Aitken kernel~\shortcite{aitchison1976multivariate} is used for categorical parameters:
\begin{eqnarray}
  k_d(x, x_n|b) = \left\{
  \begin{array}{ll}
     1 - b & (x = x_n) \\
     \frac{b}{C - 1} & (\mathrm{otherwise})
  \end{array}
  \right.
  \label{main:algorithm-detail:eq:categorical-kernel}
\end{eqnarray}
where $C$ is the number of choices in the categorical parameter and $b \in [0, 1)$ is the \emph{bandwidth} for this kernel.
Optuna v4.0.0 uses a heuristic for bandwidth computation shown in Appendix~\ref{appx:algorithm-detail:section:bandwidth-selection}, which we refer to \texttt{optuna} in the ablation study.
The bandwidth in this kernel also controls the degree of exploration and a large bandwidth is explorative.

\begin{figure}[t]
  \centering
  \includegraphics[width=0.98\textwidth]{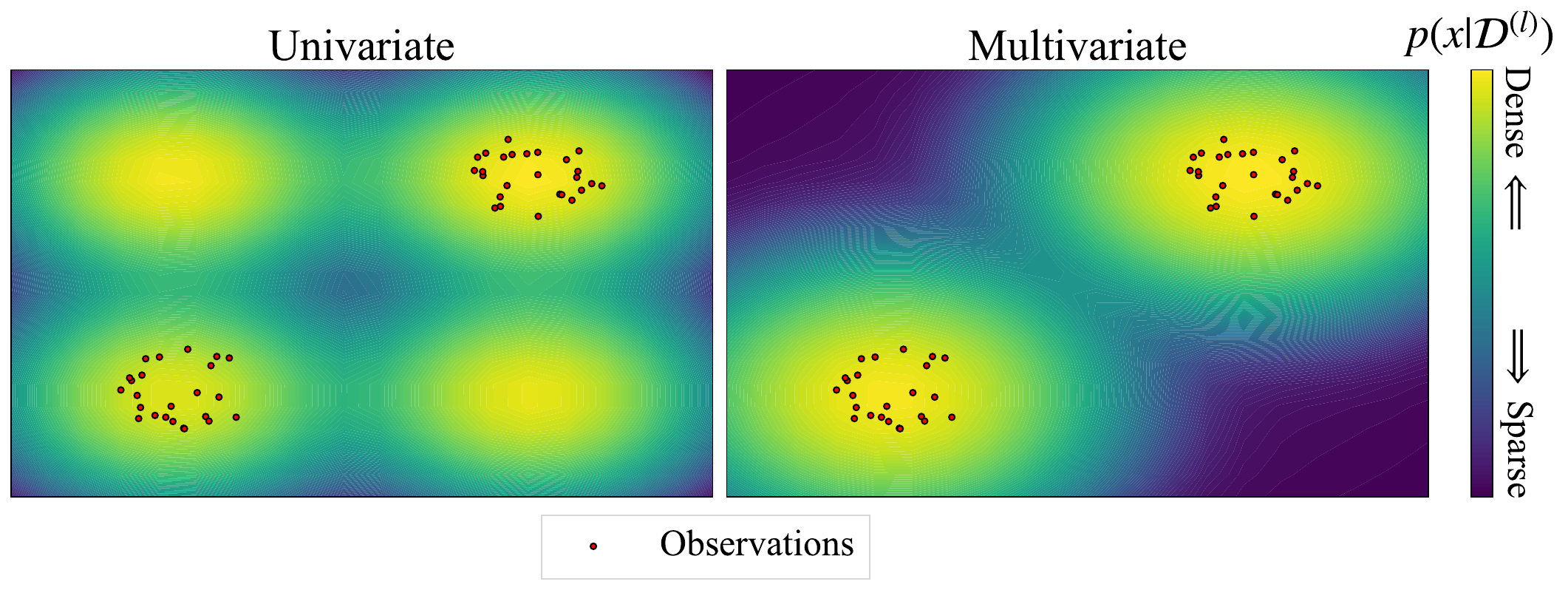}
  \vspace{-3mm}
  \caption{
    The difference between KDEs by univariate and multivariate kernels.
    The brighter color shows high density.
    \textbf{Left}: the contour plot of the KDE by the univariate kernel.
    As the univariate kernel cannot capture the interaction effects,
    we see bright colors even at the top-left and bottom-right regions.
    \textbf{Right}: the contour plot of the KDE by the multivariate kernel.
    As the multivariate kernel can capture the interaction effects,
    we see bright colors only near the observations.
  }
  \vspace{-3mm}
  \label{main:algorithm-detail:fig:uv-vs-mv}
\end{figure}

\begin{figure}[t]
  \centering
  \includegraphics[width=0.98\textwidth]{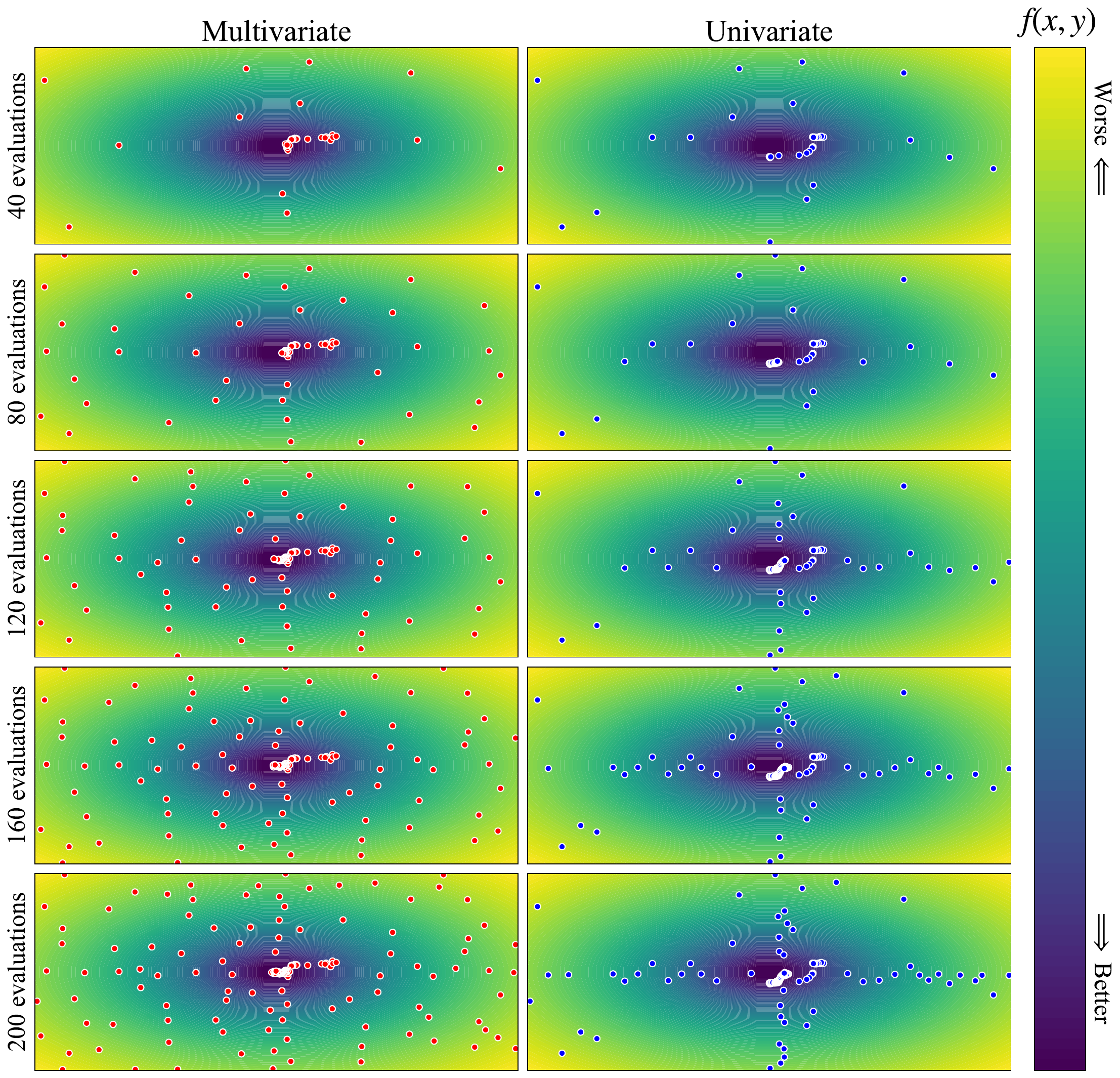}
  \vspace{-3mm}
  \caption{
    The optimizations of the Sphere function using the multivariate and univariate kernels.
    The \red{red} and \blue{blue} dots show the observations till each ``X evaluations''.
    The blue shade in each figure is the optimal point
    and this area should be found with as few observations as possible.
    \textbf{Left column}: the optimization using the multivariate kernel.
    Exploration starts after the center part is covered.
    \textbf{Right column}: the optimization using the univariate kernel.
    The observations gather close to the two lines $x_1 = 0$ and $x_2 = 0$
    because the univariate kernel cannot account for interaction effects.
  }
  \label{main:algorithm-detail:fig:exploration-by-multivariate-in-sphere}
  \vspace{-3mm}
\end{figure}

\subsubsection{Univariate Kernel vs Multivariate Kernel (\texttt{multivariate})}
\shortciteA{bergstra2011algorithms,bergstra2013making} use the so-called \emph{univariate} KDEs:
\begin{equation}
  \begin{aligned}
    p(\xv|\{\xv_n\}_{n=1}^N) \coloneqq
    \frac{1}{N} \prod_{d=1}^D \sum_{n = 1}^N k_d(x_d, x_{n,d}| b),
  \end{aligned}
  \label{main:algorithm-detail:eq:univariate-kde}
\end{equation}
where $x_{n,d} \in \X_d$ is the $d$-th dimension of the $n$-th observation $\xv_n$.
Although the independence assumption of each dimension allows the univariate kernel to handle conditional parameters, this assumption limits the performance if no conditional parameters exist.
\shortciteA{falkner2018bohb} tackle this issue using the following \emph{multivariate} KDEs:
\begin{equation}
  \begin{aligned}
    p(\xv|\{\xv_n\}_{n=1}^N) \coloneqq
    \frac{1}{N} \sum_{n = 1}^N \prod_{d=1}^D k_d(x_d, x_{n,d}| b).
  \end{aligned}
  \label{main:algorithm-detail:eq:multivariate-kde}
\end{equation}
Although the multivariate kernel cannot be applied to the tree-structured search space as is, \texttt{group=True} in Optuna overcomes this limitation as explained in Appendix~\ref{appx:algorithm-detail:section:group-param}.
As mentioned earlier, the multivariate kernel is important to improve the performance.
According to Eq.~(\ref{main:algorithm-detail:eq:univariate-kde}),
since $\sum_{n=1}^{N}k_d(x_d, x_{n,d}|b)$ is independent~\footnote{
  If $f(\xv) = \prod_{d=1}^D f_d(x_d)$ where $f_d: \X_d \rightarrow \mathbb{R}_{\geq 0}$, it is obvious that $\min_{\xv \in \X} f(\xv) = \prod_{d=1}^{D} \min_{x_d \in \X_d} f_d(x_d)$.
  Therefore, the individual optimization of each dimension leads to the optimality.
  Notice that if $f_d$ can map to a negative number, the statement is not necessarily true.
} of that of another dimension $d^\prime (\neq d)$,
the optimization of each dimension can be separately performed.
However, as Figure~\ref{main:algorithm-detail:fig:uv-vs-mv} visualizes, the univariate kernel cannot capture the interaction effects.
Meanwhile, the multivariate kernel considers interaction effects, making it unlikely to be misguided.
Figure~\ref{main:algorithm-detail:fig:exploration-by-multivariate-in-sphere} shows the multivariate kernel recognizes the exact location of the mode while the univariate kernel ends up searching the axes $x_1 = 0$ and $x_2 = 0$ separately due to the incapability to recognize the exact mode location.
Interestingly, however, the separate search of each dimension is effective for objective functions with many modes such as the Xin-She-Yang function and the Rastrigin function.

\subsubsection{Bandwidth Modification (\texttt{consider\_magic\_clip})}
\label{main:algorithm-detail:section:bandwidth}

\begin{figure}[t]
  \centering
  \includegraphics[width=0.88\textwidth]{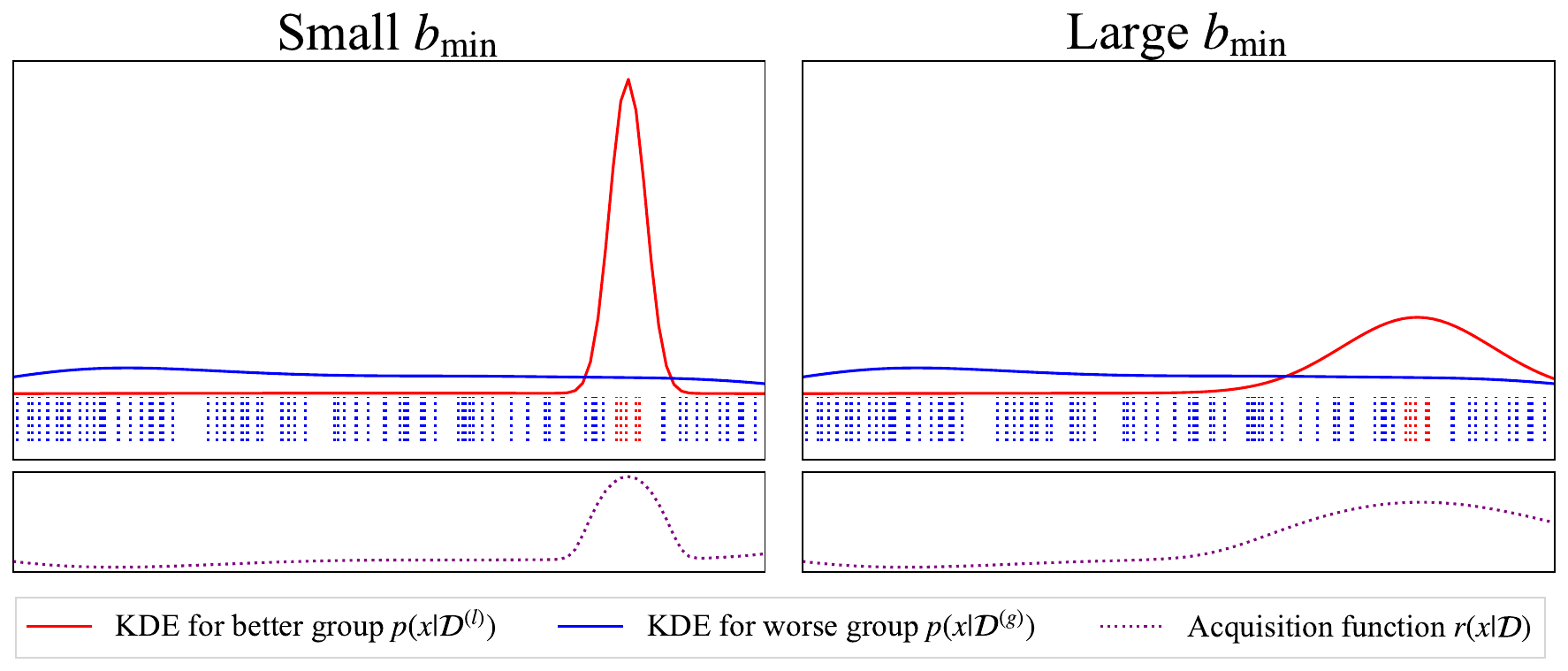}
  \vspace{-3mm}
  \caption{
    The change in the density ratio $r(\xv|\D)$ (\purple{purple dotted lines} in \textbf{Bottom row})
    with respect to the minimum bandwidth $b_{\min}$.
    Both settings use the same observations (\red{red} and \blue{blue} dotted lines in \textbf{Top row}).
    \textbf{Left}: a small $b_{\min}$.
    Since the KDE for the better group (\red{red line}) has a sharp peak,
    the density ratio is sharply peaked.
    \textbf{Right}: a large $b_{\min}$.
    The density ratio is horizontally distributed.
  }
  \vspace{-3mm}
  \label{main:algorithm-detail:fig:bandwidth-vs-density-ratio}
\end{figure}

\begin{figure}[t]
  \centering
  \includegraphics[width=0.88\textwidth]{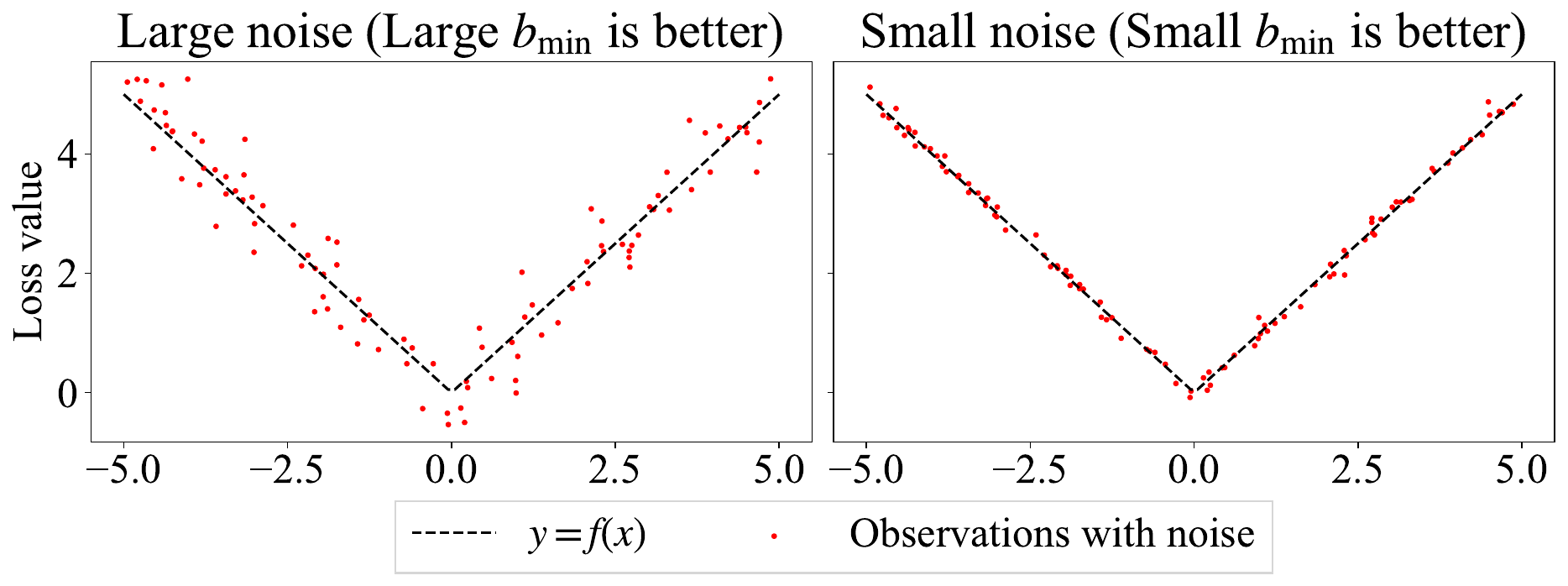}
  \vspace{-3mm}
  \caption{
    A concrete example where the minimum bandwidth matters.
    The black dashed lines are the true objective function without noise and the \red{red dots} are the observations with noise.
    \textbf{Left}: a function that has a large noise compared to the variation with respect to $x$.
    As can be seen, some observations near $x = 1$ are better than the others at $x = 1$ due to the noise.
    In such cases, smaller $b_{\min}$, which leads to more precise optimizations, does not make much difference.
    \textbf{Right}: a function that has a small noise compared to the variation with respect to $x$.
    Since the noise is small, smaller $b_{\min}$ helps to yield precise solutions.
  }
  \vspace{-3mm}
  \label{main:algorithm-detail:fig:noise-and-cardinality}
\end{figure}

The bandwidth $b$ of the numerical kernel defined on $[L, R]$ is first computed by a heuristic, e.g., Scott's rule~\shortcite{scott2015multivariate}, and then is clipped by the so-called \emph{magic clipping} invented by \shortciteA{bergstra2011algorithms}:
\begin{equation}
\begin{aligned}
  b_{\mathrm{new}} \coloneqq \max(\{b, b_{\min}\})~\mathrm{where}~
    b_{\min} \coloneqq \frac{R - L}{\min(\{100, N\})}
\end{aligned}
\label{main:algorithm-detail:eq:magic-clipping}
\end{equation}
where $N = \Nl + 1$ is used for $p(\xv|\Dl)$ and $\Ng + 1$ is used for $p(\xv|\Dg)$.
Note that if $p(\xv | \D^{(\cdot)})$ does not include the prior $p_0$, $N = N^{(\cdot)}$ is used instead.
Appendix~\ref{appx:algorithm-detail:section:bandwidth-selection} discusses more details of the bandwidth selection algorithms used in TPE.
In principle, $b_{\min}$ becomes exploitative when it is small and becomes explorative when it is large as shown in Figure~\ref{main:algorithm-detail:fig:bandwidth-vs-density-ratio}.
The magic clipping mostly expands the bandwidth, changing the behavior of TPE from exploitative to explorative.
For example, when we search for the best dropout rate of neural networks from the range of $[0, 1]$, the difference between $0.4$ and $0.5$ is important but that between $0.40$ and $0.41$ is not so important because the noise is likely to be dominant in the performance variation.
Figure~\ref{main:algorithm-detail:fig:noise-and-cardinality} intuitively illustrates this point.
Since the performance variation is explained by the noise rather than the parameter (\textbf{Left}), more precise optimization by a small bandwidth is not necessary.
In contrast, when the noise is negligible (\textbf{Right}), a small bandwidth is essential.
The noisy example may be optimized sufficiently by picking a value from $\{-5, -4, \dots, 4, 5\}$, which we call intrinsic set.
We denote the size of an intrinsic set as \emph{intrinsic cardinality}, which is $11$ in this example.
The scale of $b_{\min}$ should be inversely proportional to the \emph{intrinsic cardinality} of each parameter.
The bandwidth modification is individually analyzed in the ablation study for better performance.

\begin{figure}[t]
  \centering
  \includegraphics[width=0.98\textwidth]{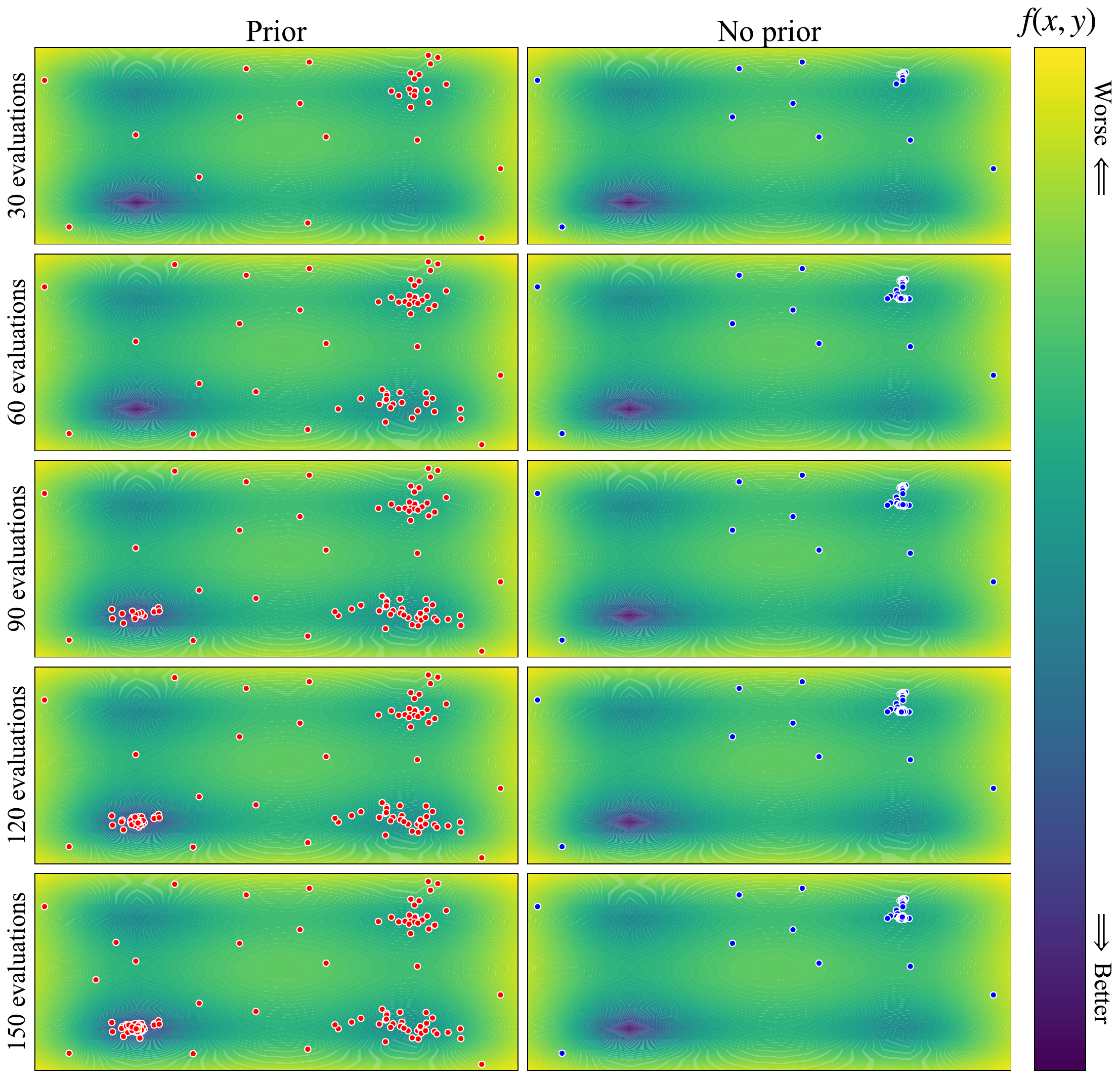}
  \vspace{-3mm}
  \caption{
    The optimizations of the Styblinski function with and without prior $p_0$.
    The \red{red} and \blue{blue} dots show the observations till each ``X evaluations''.
    The lower left blue shade in each figure is the optimal point
    and this area should be found with as few observations as possible.
    \textbf{Left column}: the optimization with prior.
    Unseen regions are explored every time each mode is covered by many observations.
    \textbf{Right column}: the optimization without prior.
    No exploration happens after one of the modes is found.
  }
  \vspace{-5mm}
  \label{main:algorithm-detail:fig:exploration-by-prior-in-styblinski}
\end{figure}

\subsubsection{Non-Informative Prior (\texttt{consider\_prior}, \texttt{prior\_weight})}
\label{main:algorithm-detail:section:prior}

Prior $p_0$ in Eq.~(\ref{main:background:eq:kernel-density-estimation}) essentially controls the regularization effect in KDEs.
The PDF of Gaussian distribution $\mathcal{N}((R + L) / 2, (R - L)^2)$ is used for a numerical parameter defined on $[L, R]$, and the probability mass function of uniform categorical distribution $\mathcal{U}(\{1,\dots,C\})$ is used for a categorical parameter $\{1,\dots,C\}$.
Prior is especially important for $p(\xv|\Dl)$ to prevent strong exploitation, which is seen in Figure~\ref{main:algorithm-detail:fig:exploration-by-prior-in-styblinski}, due to a small $\Nl$ throughout an optimization.
Prior is mostly indispensable to TPE as discussed later.
The regularization effect of prior can be controlled by \texttt{prior\_weight} as well.
For example, \texttt{prior\_weight=2.0} doubles $\wl, \wg$, promoting exploration.

\begin{table}
  \begin{center}
    \caption{
      The implementational differences in each component of TPE variants.
      Other components such as prior and the number of initial configurations are shared across all the implementations.
      TPE (2011), TPE (2013), BOHB, MOTPE, c-TPE, and Optuna refer to TPE by \shortciteA{bergstra2011algorithms}, TPE (Hyperopt) by \shortciteA{bergstra2013making}, TPE in BOHB~\shortcite{falkner2018bohb}, TPE in MOTPE~\shortcite{ozaki2022multiobjective}, TPE in c-TPE~\shortcite{watanabe2023ctpe}, and TPE in Optuna v4.0.0, respectively.
      Note that \texttt{uniform} of BOHB is computed by Eq.~(\ref{appx:algorithm-detail:eq:bohb-weighting}), and c-TPE uses a fixed $b_{\min}$ instead of magic clipping.
      For the bandwidth selection, \texttt{hyperopt}, \texttt{scott}, and \texttt{optuna} refer to Eqs.~(\ref{appx:algorithm-detail:eq:hyperopt-bandwidth-selection}),
      (\ref{appx:algorithm-detail:eq:scott-bandwidth}), and
      (\ref{appx:algorithm-detail:eq:optuna-numerical-bandwidth}) in Appendix~\ref{appx:algorithm-detail:section:bandwidth-selection}, respectively.
      BOHB calculates the bandwidth of categorical parameters using \texttt{scott} as if they are numerical parameters.
    }
    \vspace{2mm}
    \label{main:algorithm-detail:tab:component-comparison}
    \makebox[1 \textwidth][c]{       
      \resizebox{1 \textwidth}{!}{   
        \begin{tabular}{lcccccc}
          \toprule
          \multirow[]{2}{*}{Version} & Splitting                         & Weighting                & \multicolumn{2}{c}{Bandwidth} & Multivariate                                                      & Magic clipping                                             \\
                                     & (\texttt{gamma})                  & (\texttt{weights})       & Numerical                     & Categorical                                                       & (\texttt{multivariate}) & (\texttt{consider\_magic\_clip}) \\
          \midrule
          TPE (2011)                 & \texttt{linear}, $\beta_1 = 0.15$ & \texttt{uniform}         & \texttt{hyperopt}             & $b=0$                                                             & \texttt{False}          & \texttt{True}                    \\
          TPE (2013)                 & \texttt{sqrt}, $\beta_2 = 0.25$   & \texttt{old decay}       & \texttt{hyperopt}             & $b=0$                                                             & \texttt{False}          & \texttt{True}                    \\
          BOHB                       & \texttt{linear}, $\beta_1 = 0.15$ & \texttt{uniform}$^\star$ & \texttt{scott}                & \texttt{scott}                                                    & \texttt{True}           & \texttt{False}                   \\
          MOTPE                      & \texttt{linear}, $\beta_1 = 0.10$ & \texttt{EI}              & \texttt{hyperopt}             & $b=0$                                                             & \texttt{False}          & \texttt{True}                    \\
          c-TPE                      & \texttt{sqrt}, $\beta_2 = 0.25$   & \texttt{uniform}         & \texttt{scott}                & $b=0.2$                                                           & \texttt{True}           & \texttt{False}$^\star$           \\
          Optuna                     & \texttt{linear}, $\beta_1 = 0.10$ & \texttt{old decay}       & \texttt{optuna}               & Eq.~(\ref{appx:algorithm-detail:eq:optuna-categorical-bandwidth}) & \texttt{True}           & \texttt{True}                    \\
          \bottomrule
        \end{tabular}
      }
    }
  \end{center}
  \vspace{-3mm}
\end{table}

\section{Experiments}
\label{main:experiments:section:ablation-study}

\begin{table}[t]
  \begin{center}
    \caption{
      The search space of the control parameters used in the ablation study.
      Note that $\beta_1, \beta_2$ are conditional parameters
      and we have $2 \times 2 \times 2 \times (4 + 4) \times 4 \times 4 = 1024$ possible combinations in total.
    }
    \vspace{2mm}
    \label{main:experiments:tab:search-space-ablation}
    \begin{tabular}{ll}
      \toprule
      Component                                                                            & Choices                                                                  \\
      \midrule
      Multivariate (\texttt{multivariate})                                                 & \{\texttt{True}, \texttt{False}\}                                        \\
      Use prior $p_0$ (\texttt{consider\_prior})                                           & \{\texttt{True}, \texttt{False}\}                                        \\
      Use magic clipping (\texttt{consider\_magic\_clip})                                  & \{\texttt{True}, \texttt{False}\}                                        \\
      Splitting algorithm $\Gamma$ (\texttt{gamma})                                        & \{\texttt{linear}, \texttt{sqrt}\}                                       \\
      ~~~~1. $\beta_1$ in \texttt{linear}                                                  & \{0.05, 0.10, 0.15, 0.20\}                                               \\
      ~~~~2. $\beta_2$ in \texttt{sqrt}                                                    & \{0.25, 0.50, 0.75, 1.0\}                                                \\
      Weighting algorithm $W$ (\texttt{weights})                                           & \{\texttt{uniform}, \texttt{old-decay}, \texttt{old-drop}, \texttt{EI}\} \\
      Categorical bandwidth $b$ in Eq.~(\ref{main:algorithm-detail:eq:categorical-kernel}) & \{0.0, 0.1, 0.2, \texttt{optuna}\}                                   \\
      \bottomrule
    \end{tabular}
  \end{center}
  \vspace{-3mm}
\end{table}

This section first provides the ablation study of each control parameter and presents the recommended default values.
Then, we further investigate the enhancement for bandwidth selection.
Finally, we compare the enhanced TPE with various baseline methods.

\subsection{Ablation Study}
This section aims to identify the importance of each control parameter discussed in the previous section via the ablation study and provides the recommended default setting.

\subsubsection{Setup}
\label{main:experiments:section:ablation-setup}
\begin{figure}[t]
  \begin{center}
    \vspace{-10mm}
    \subfloat[Benchmark functions with 5D\vspace{-3mm}]{
      \includegraphics[width=0.8\textwidth]{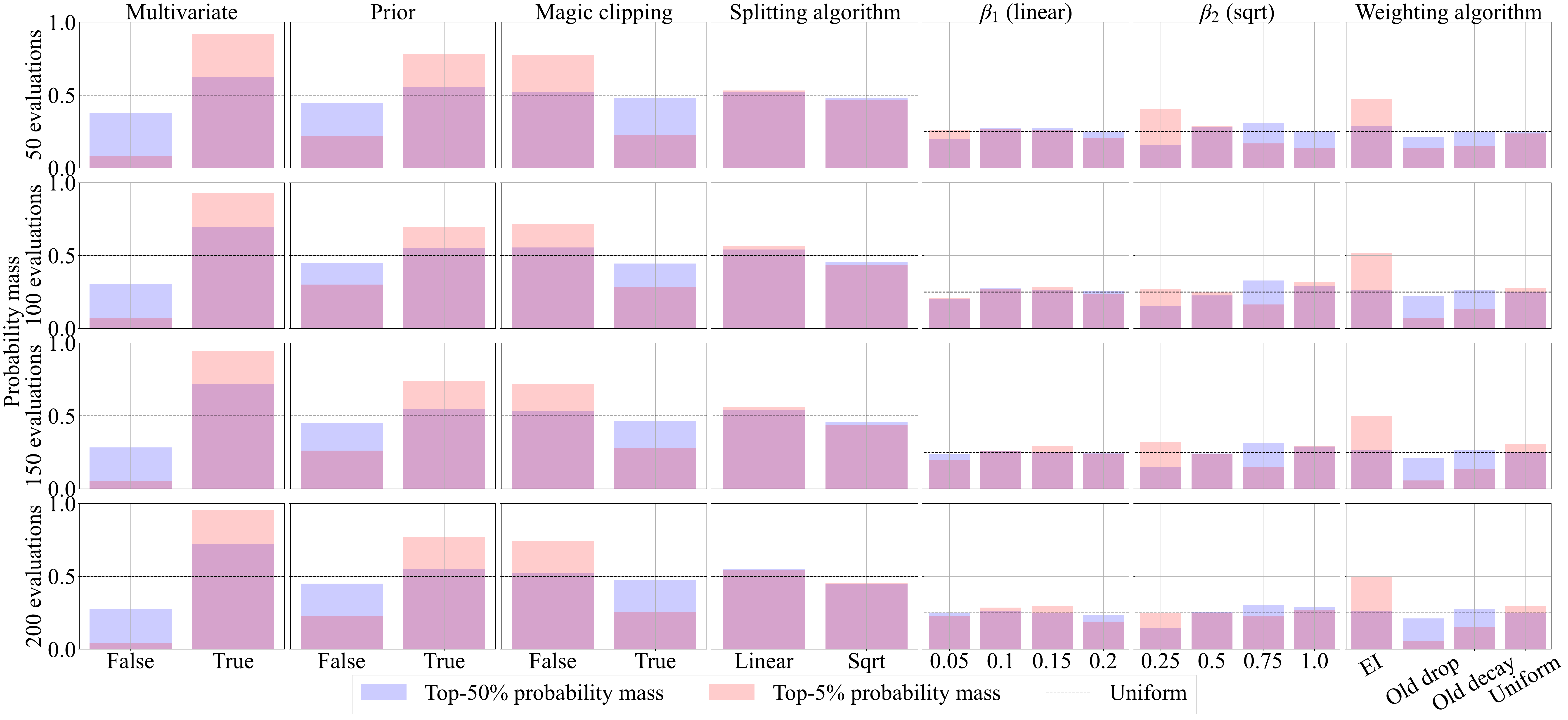}
    } \\
    \subfloat[Benchmark functions with 10D\vspace{-3mm}]{
      \includegraphics[width=0.8\textwidth]{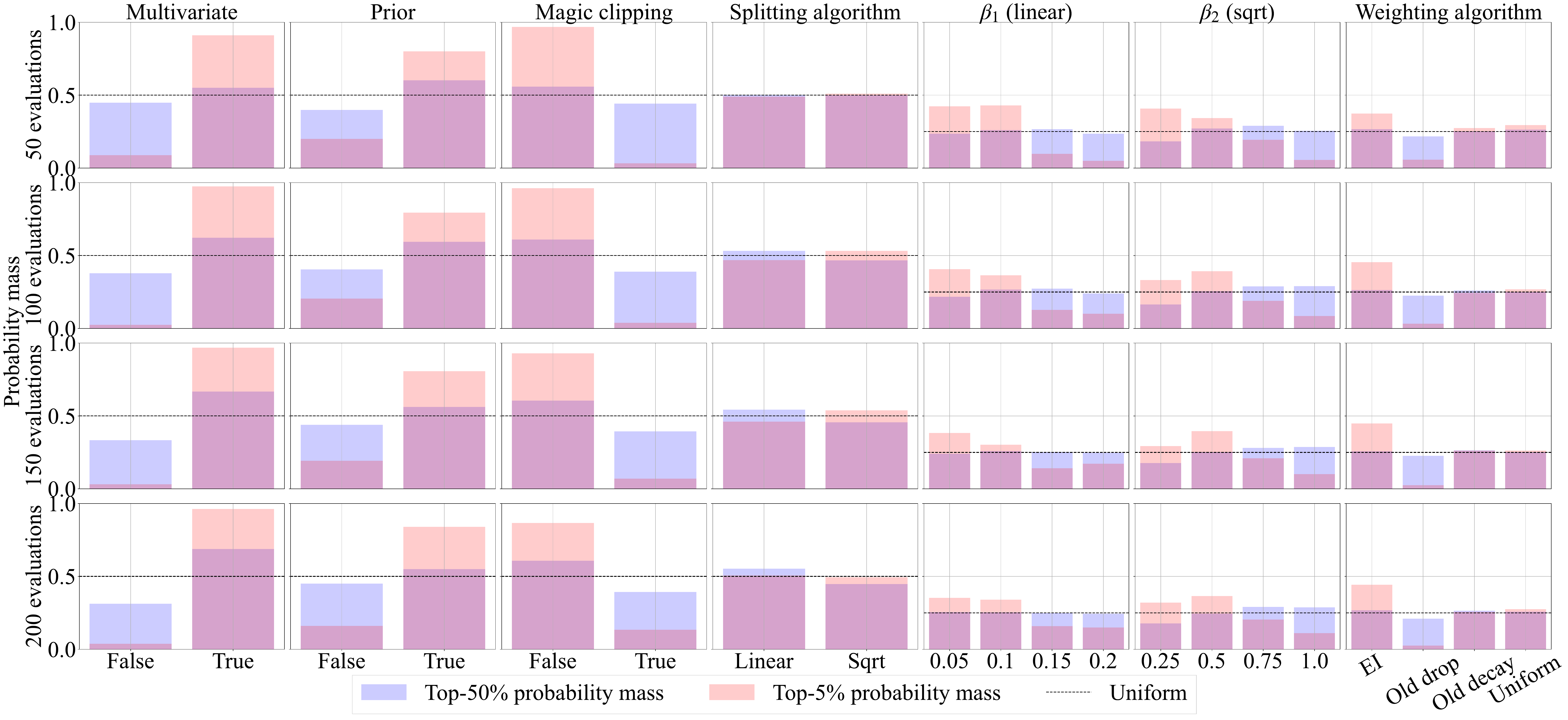}
    } \\
    \subfloat[Benchmark functions with 30D\vspace{-3mm}]{
      \includegraphics[width=0.8\textwidth]{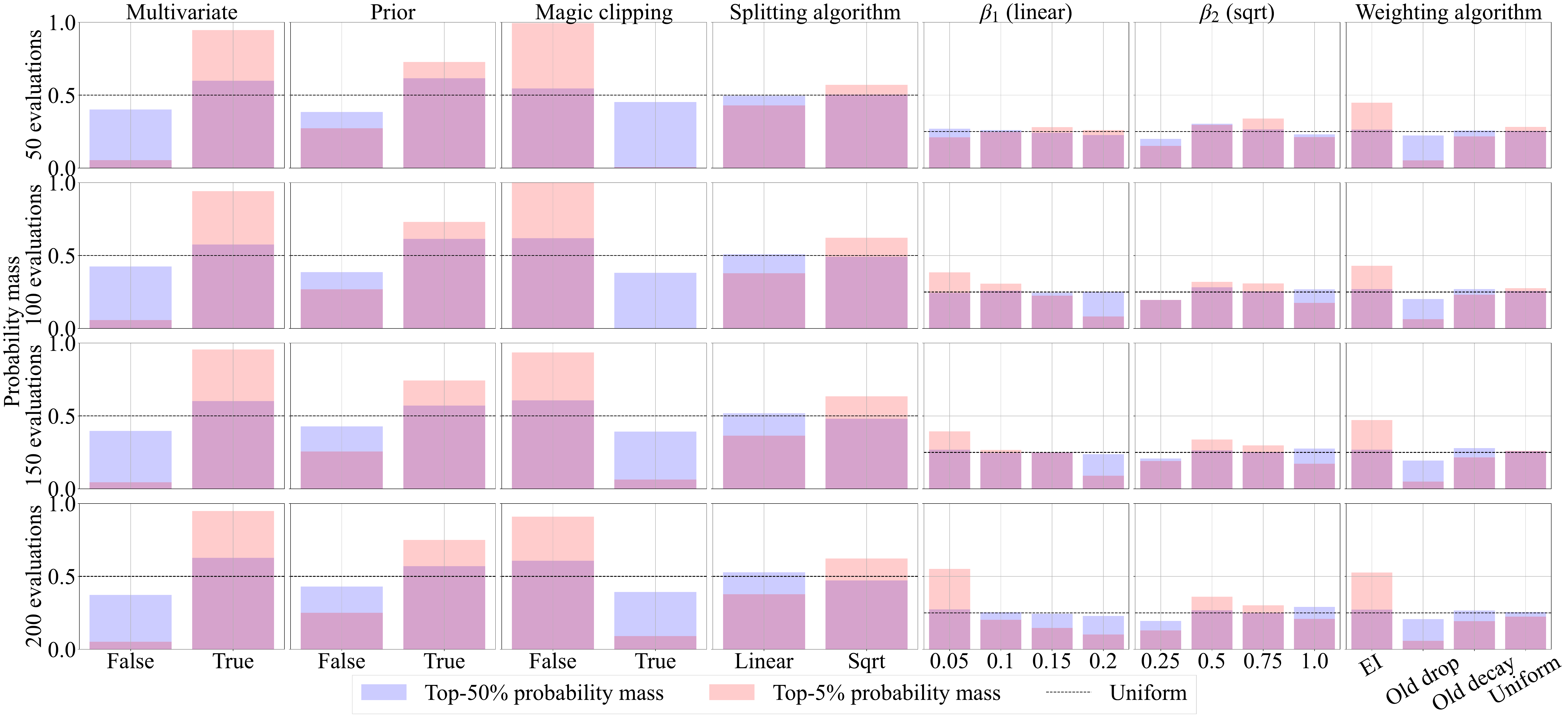}
    }
    \caption{
      The probability mass values of each control parameter
      in the top-$5\%$ and top-$50\%$ observations
      at \{50,100,150,200\} evaluations
      on the benchmark functions.
      High probability mass in a specific value implies that
      we are likely to yield top-$5\%$ or top-$50\%$
      performance with the specific value.
    }
    \label{main:experiments:fig:bench-anova}
    \vspace{-10mm}
  \end{center}
\end{figure}

\begin{figure}[t]
  \begin{center}
    \vspace{-10mm}
    \subfloat[HPOBench\vspace{-3mm}]{
      \includegraphics[width=0.85\textwidth]{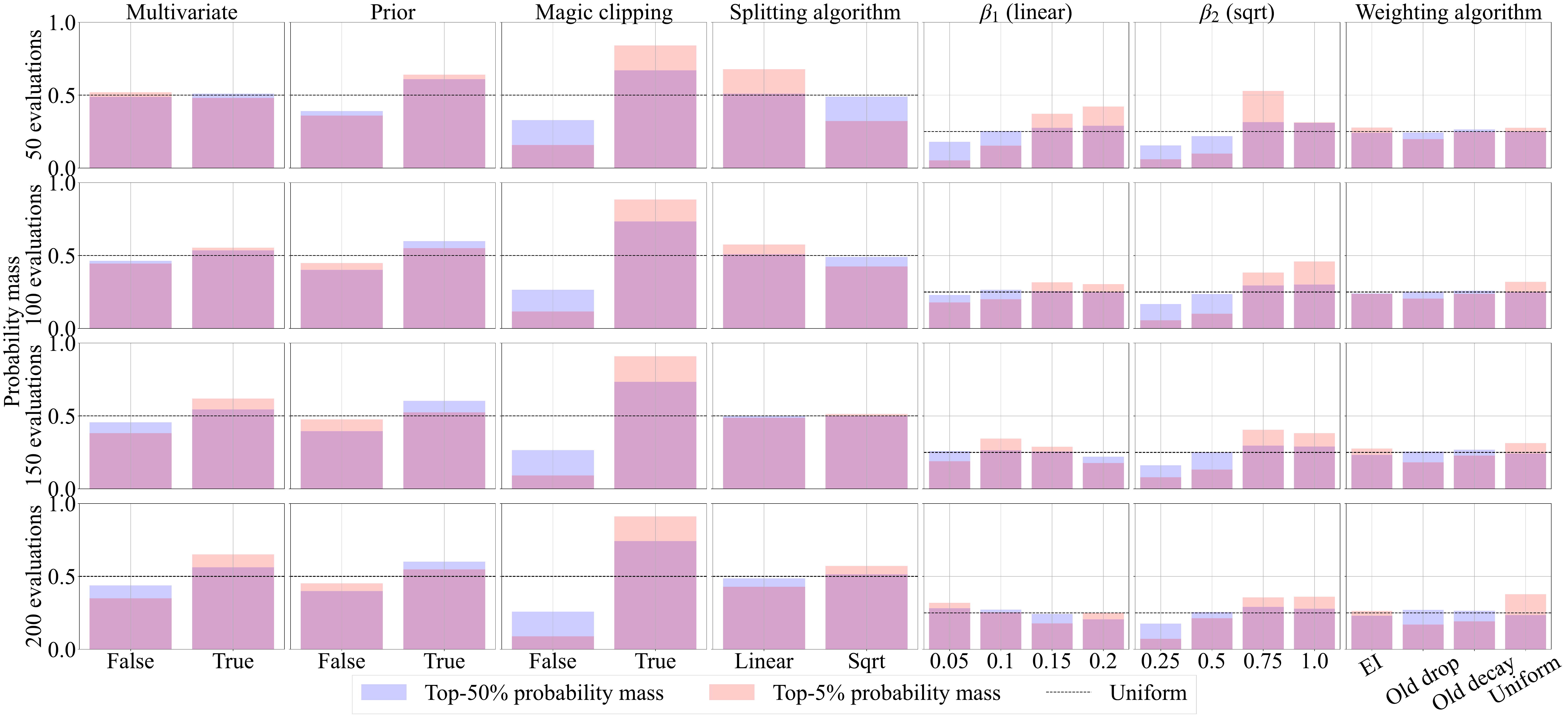}
    } \\
    \subfloat[HPOlib\vspace{-3mm}]{
      \includegraphics[width=0.85\textwidth]{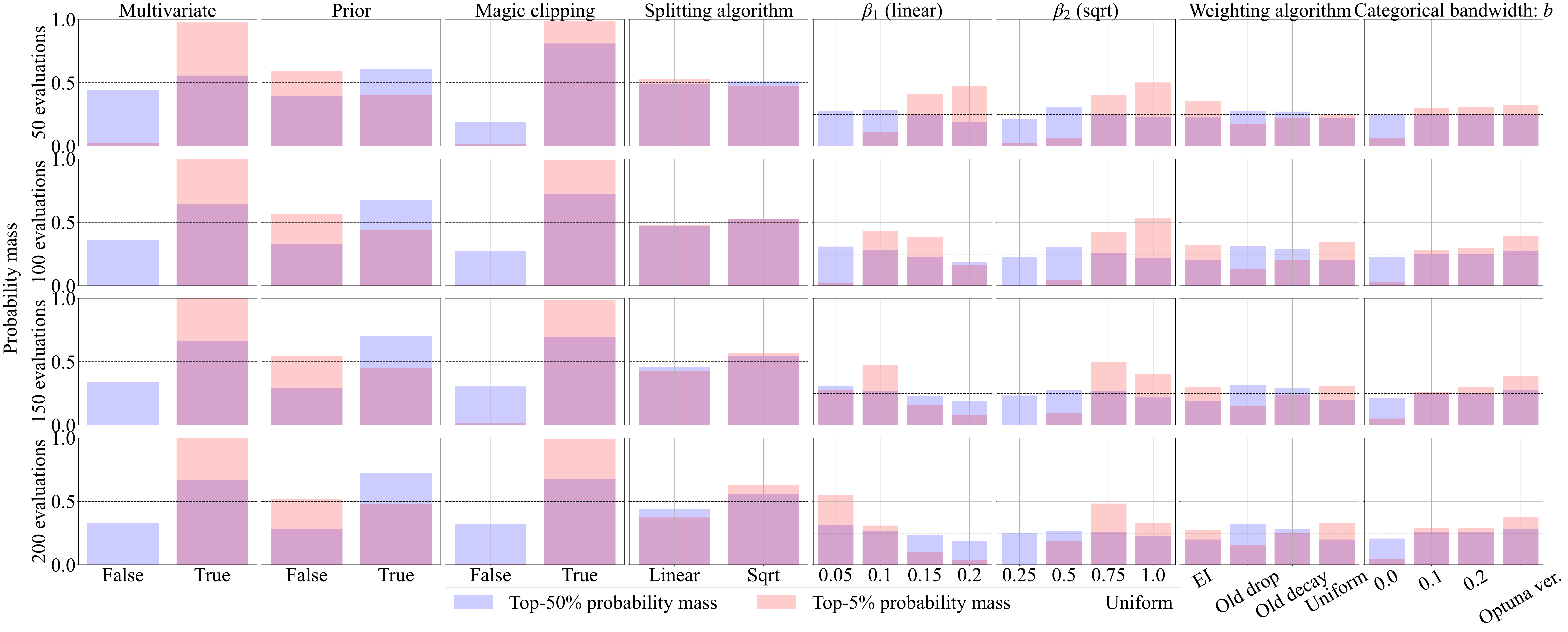}
    } \\
    \subfloat[JAHS-Bench-201\vspace{-3mm}]{
      \includegraphics[width=0.85\textwidth]{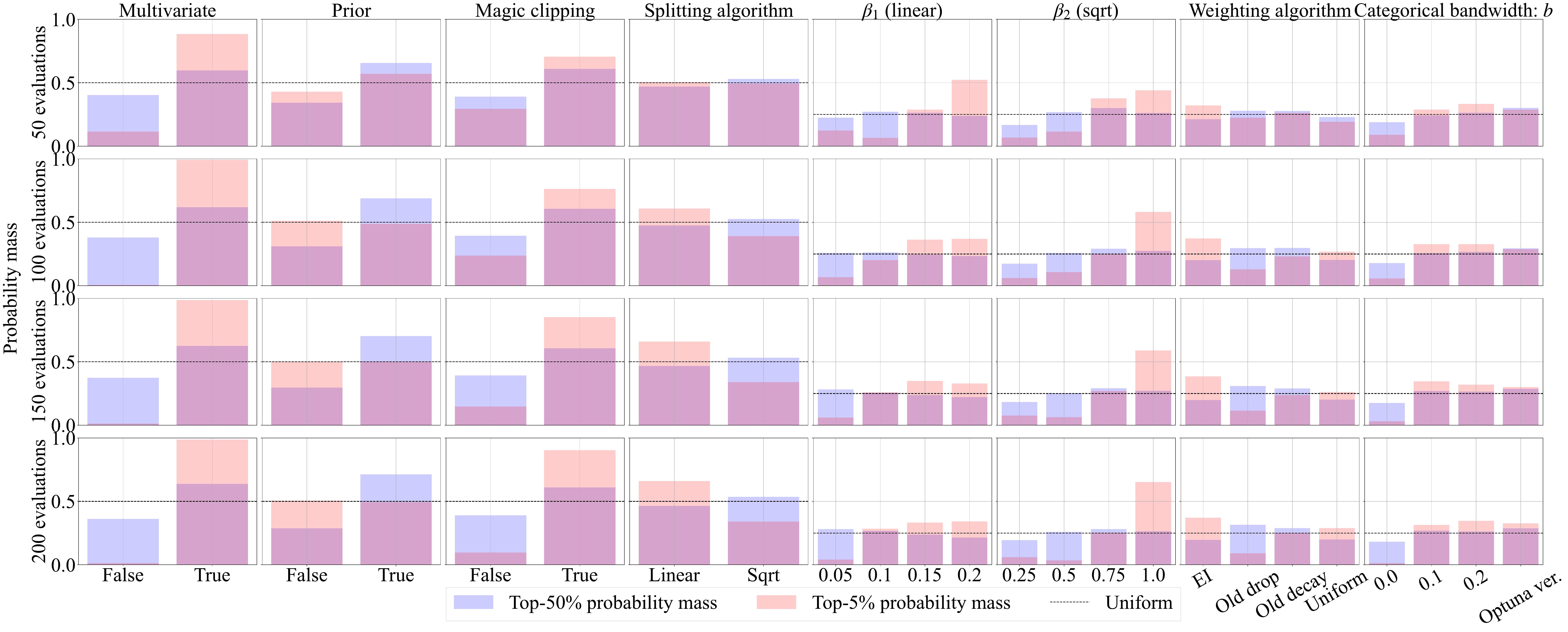}
    }
    \caption{
      The probability mass values of each control parameter
      in the top-$5\%$ and top-$50\%$ observations
      at \{50,100,150,200\} evaluations
      on the HPO benchmarks.
      High probability mass in a specific value implies that
      we are likely to yield top-$5\%$ or top-$50\%$
      performance with the specific value.
    }
    \label{main:experiments:fig:hpobench-anova}
    \vspace{-10mm}
  \end{center}
\end{figure}

We exhaustively evaluate all the possible combinations of the control parameters specified in Table~\ref{main:experiments:tab:search-space-ablation} to find performant default values.
Note that \texttt{old-drop} is added to the choices of the weighting algorithm to verify whether old information is necessary.
\texttt{old-drop} gives uniform weights to the recent $T_{\mathrm{old}} = 25$ observations and drops the weights (i.e., to give zero weights) for the rest.
The other parameters not specified in the table are fixed to the default values of Optuna v4.0.0 except Scott's rule in Eq.~(\ref{appx:algorithm-detail:eq:scott-bandwidth}) is used for the numerical bandwidth selection.
The experiments are conducted over $51$ objective functions listed in Appendix~\ref{appx:experiments:section:detail-of-funcs} to strengthen the reliability of the analysis.
The objective functions include $36$ benchmark functions ($12$ different functions $\times$ $3$ different dimensionalities),
$8$ tasks in HPOBench~\shortcite{eggensperger2021hpobench},
$4$ tasks in HPOlib~\shortcite{klein2019tabular},
and $3$ tasks in JAHS-Bench-201~\shortcite{bansal2022jahs}.
The default scale of $y$ is used in \texttt{EI} except on HPOlib where the log scale of the validation MSE is used.
Each optimization observes $200$ configurations and is repeated $10$ times each with a different random seed.
The initialization follows the default setting of Optuna v4.0.0, meaning that $10$ random configurations are evaluated before each optimization (i.e., \texttt{n\_startup\_trials=10}).
The analysis for Figures~\ref{main:experiments:fig:bench-anova} -- \ref{main:experiments:fig:hpobench-anova-bw} is performed based on \shortciteA{watanabe2023pedanova}.
Roughly speaking, parameter choices above the black-dotted lines are likely to achieve good performance.
More specifically, the red and blue shades above the black-dotted lines contribute to the top-5\% and the top-50\% among all the possible configurations, respectively.
For more details, see Appendix~\ref{appx:experiments:section:detail-of-funcs}.

\begin{table}[t]
  \begin{center}
    \caption{
      The hyperparameter importance (HPI) of each control parameter on the benchmark functions to achieve
      top $50\%$ and to improve to top $5\%$ from top $50\%$.
      HPI is measured at \{50, 100, 150, 200\} evaluations.
      We bold the top-2 HPI.
      Note that while the HPI in $\beta$ quantifies whether varying $\beta$ given either \texttt{linear} or \texttt{sqrt} matters, that in ``Splitting algorithm'' quantifies whether switching between \texttt{linear} and \texttt{sqrt} matters.
    }
    \vspace{2mm}
    \label{main:experiments:tab:bench-hpi}
    \makebox[1 \textwidth][c]{       
      \resizebox{1 \textwidth}{!}{   
        \begin{tabular}{c|lcc|cc|cc|cc}
          \toprule
          \multirow[]{3}{*}{Dimension} & \multirow[]{3}{*}{Control parameter} & \multicolumn{8}{c}{The number of function evaluations}                                                                                                                                                                                               \\
          \cmidrule{3-10}
                                       &                                      & \multicolumn{2}{c}{50 evaluations}                     & \multicolumn{2}{c}{100 evaluations} & \multicolumn{2}{c}{150 evaluations} & \multicolumn{2}{c}{200 evaluations}                                                                             \\
                                       &                                      & Top $50\%$                                             & Top $5\%$                           & Top $50\%$                          & Top $5\%$                           & Top $50\%$       & Top $5\%$        & Top $50\%$       & Top $5\%$        \\
          \midrule
          \multirow[]{7}{*}{5D}        & Multivariate                         & \textbf{23.67\%}                                       & 12.77\%                             & \textbf{41.39\%}                    & 14.71\%                             & \textbf{48.67\%} & 10.90\%          & \textbf{50.81\%} & 11.28\%          \\
                                       & Prior                                & 6.21\%                                                 & 9.45\%                              & 4.65\%                              & 10.47\%                             & 6.57\%           & 8.87\%           & 8.45\%           & 10.19\%          \\
                                       & Magic clipping                       & \textbf{36.11\%}                                       & 10.35\%                             & \textbf{21.65\%}                    & 9.33\%                              & 15.19\%          & 8.77\%           & 10.45\%          & 12.60\%          \\
                                       & Splitting algorithm                  & 1.09\%                                                 & 0.97\%                              & 1.78\%                              & 3.33\%                              & 1.95\%           & 2.95\%           & 3.05\%           & 3.09\%           \\
                                       & ~~~~$\beta_1$ (\texttt{linear})               & 9.80\%                                                 & \textbf{17.95\%}                    & 7.10\%                              & 11.04\%                             & 5.80\%           & 8.37\%           & 7.24\%           & 16.86\%          \\
                                       & ~~~~$\beta_2$ (\texttt{sqrt})                 & 19.24\%                                                & \textbf{39.12\%}                    & 20.80\%                             & \textbf{27.89\%}                    & \textbf{17.64\%} & \textbf{39.32\%} & \textbf{16.08\%} & \textbf{22.36\%} \\
                                       & Weighting algorithm                  & 3.87\%                                                 & 9.39\%                              & 2.62\%                              & \textbf{23.22\%}                    & 4.18\%           & \textbf{20.82\%} & 3.92\%           & \textbf{23.62\%} \\
          \midrule
          \multirow[]{7}{*}{10D}       & Multivariate                         & 8.39\%                                                 & 16.64\%                             & \textbf{26.23\%}                    & 14.87\%                             & \textbf{44.48\%} & 14.38\%          & \textbf{48.94\%} & 14.24\%          \\
                                       & Prior                                & \textbf{25.69\%}                                       & 11.48\%                             & 12.96\%                             & 14.76\%                             & 6.31\%           & \textbf{18.81\%} & 4.50\%           & \textbf{19.72\%} \\
                                       & Magic clipping                       & \textbf{38.79\%}                                       & 20.09\%                             & \textbf{31.63\%}                    & 14.98\%                             & \textbf{25.45\%} & 14.60\%          & \textbf{21.81\%} & 11.44\%          \\
                                       & Splitting algorithm                  & 0.87\%                                                 & 0.69\%                              & 2.34\%                              & 2.23\%                              & 3.32\%           & 2.48\%           & 3.76\%           & 3.23\%           \\
                                       & ~~~~$\beta_1$ (\texttt{linear})               & 7.83\%                                                 & \textbf{20.42\%}                    & 8.49\%                              & \textbf{17.92\%}                    & 5.77\%           & 15.04\%          & 5.68\%           & 13.81\%          \\
                                       & ~~~~$\beta_2$ (\texttt{sqrt})                 & 14.70\%                                                & \textbf{24.06\%}                    & 15.79\%                             & \textbf{22.84\%}                    & 11.83\%          & \textbf{21.78\%} & 11.51\%          & \textbf{25.76\%} \\
                                       & Weighting algorithm                  & 3.74\%                                                 & 6.61\%                              & 2.55\%                              & 12.41\%                             & 2.84\%           & 12.91\%          & 3.78\%           & 11.81\%          \\
          \midrule
          \multirow[]{7}{*}{30D}       & Multivariate                         & \textbf{32.83\%}                                       & 17.15\%                             & 18.62\%                             & 18.37\%                             & \textbf{31.01\%} & 16.89\%          & \textbf{36.80\%} & 15.23\%          \\
                                       & Prior                                & \textbf{29.53\%}                                       & 14.52\%                             & \textbf{24.91\%}                    & 15.55\%                             & 12.38\%          & \textbf{19.74\%} & 9.52\%           & \textbf{20.66\%} \\
                                       & Magic clipping                       & 16.31\%                                                & \textbf{26.46\%}                    & \textbf{31.82\%}                    & \textbf{18.99\%}                    & \textbf{28.84\%} & 14.69\%          & \textbf{27.98\%} & 12.38\%          \\
                                       & Splitting algorithm                  & 1.06\%                                                 & 1.87\%                              & 2.12\%                              & 3.17\%                              & 1.58\%           & 5.57\%           & 2.45\%           & 5.69\%           \\
                                       & ~~~~$\beta_1$ (\texttt{linear})               & 4.82\%                                                 & \textbf{17.80\%}                    & 4.54\%                              & \textbf{23.22\%}                    & 8.66\%           & \textbf{22.77\%} & 6.26\%           & \textbf{22.58\%} \\
                                       & ~~~~$\beta_2$ (\texttt{sqrt})                 & 12.17\%                                                & 8.56\%                              & 11.24\%                             & 8.94\%                              & 8.56\%           & 7.93\%           & 9.86\%           & 8.71\%           \\
                                       & Weighting algorithm                  & 3.28\%                                                 & 13.65\%                             & 6.75\%                              & 11.75\%                             & 8.96\%           & 12.40\%          & 7.13\%           & 14.75\%          \\
          \bottomrule
        \end{tabular}
      }
    }
  \end{center}
  \vspace{-7mm}
\end{table}

\begin{table}[t]
  \begin{center}
    \caption{
      The hyperparameter importance (HPI) of each control parameter on the HPO benchmarks to achieve
      top $50\%$ and to improve to top $5\%$ from top $50\%$.
      HPI is measured at \{50, 100, 150, 200\} evaluations.
      We bold the top-2 HPI.
      Note that while the HPI in $\beta$ quantifies whether varying $\beta$ given either \texttt{linear} or \texttt{sqrt} matters, that in ``Splitting algorithm'' quantifies whether switching between \texttt{linear} and \texttt{sqrt} matters.
    }
    \vspace{2mm}
    \label{main:experiments:tab:hpobench-hpi}
    \makebox[1 \textwidth][c]{       
      \resizebox{1 \textwidth}{!}{   
        \begin{tabular}{c|lcc|cc|cc|cc}
          \toprule
          \multirow[]{3}{*}{Benchmark}      & \multirow[]{3}{*}{Control parameter} & \multicolumn{8}{c}{The number of function evaluations}                                                                                                                                                                                               \\
          \cmidrule{3-10}
                                            &                                      & \multicolumn{2}{c}{50 evaluations}                     & \multicolumn{2}{c}{100 evaluations} & \multicolumn{2}{c}{150 evaluations} & \multicolumn{2}{c}{200 evaluations}                                                                             \\
                                            &                                      & Top $50\%$                                             & Top $5\%$                           & Top $50\%$                          & Top $5\%$                           & Top $50\%$       & Top $5\%$        & Top $50\%$       & Top $5\%$        \\
          \midrule
          \multirow[]{7}{*}{HPOBench}       & Multivariate                         & 1.41\%                                                 & 6.79\%                              & 2.99\%                              & 8.35\%                              & 3.29\%           & 4.78\%           & 5.11\%           & 6.36\%           \\
                                            & Prior                                & 15.43\%                                                & 0.60\%                              & 11.11\%                             & 1.95\%                              & 11.78\%          & 2.90\%           & \textbf{10.93\%} & 3.30\%           \\
                                            & Magic clipping                       & \textbf{38.76\%}                                       & 10.16\%                             & \textbf{63.38\%}                    & 11.95\%                             & \textbf{60.96\%} & 12.77\%          & \textbf{62.16\%} & 12.43\%          \\
                                            & Splitting algorithm                  & 2.00\%                                                 & 7.61\%                              & 1.29\%                              & 12.26\%                             & 1.11\%           & 4.44\%           & 1.02\%           & 2.98\%           \\
                                            & ~~~~$\beta_1$ (\texttt{linear})               & 12.61\%                                                & \textbf{37.74\%}                    & 3.12\%                              & \textbf{29.52\%}                    & 6.00\%           & \textbf{36.62\%} & 7.94\%           & \textbf{42.21\%} \\
                                            & ~~~~$\beta_2$ (\texttt{sqrt})                 & \textbf{27.87\%}                                       & \textbf{29.51\%}                    & \textbf{17.18\%}                    & \textbf{28.03\%}                    & \textbf{15.49\%} & \textbf{23.06\%} & 10.87\%          & 15.18\%          \\
                                            & Weighting algorithm                  & 1.93\%                                                 & 7.59\%                              & 0.93\%                              & 7.94\%                              & 1.38\%           & 15.43\%          & 1.97\%           & \textbf{17.53\%} \\
          \midrule
          \multirow[]{8}{*}{HPOlib}         & Multivariate                         & 3.06\%                                                 & 20.88\%                             & 15.51\%                             & 16.10\%                             & 19.25\%          & \textbf{17.85\%} & 21.00\%          & 17.09\%          \\
                                            & Prior                                & \textbf{8.88\%}                                        & 6.11\%                              & \textbf{23.51\%}                    & 7.56\%                              & \textbf{31.62\%} & 10.95\%          & \textbf{35.41\%} & 10.44\%          \\
                                            & Magic clipping                       & \textbf{72.53\%}                                       & 5.89\%                              & \textbf{38.75\%}                    & 10.62\%                             & \textbf{28.31\%} & 13.90\%          & \textbf{22.96\%} & 16.06\%          \\
                                            & Splitting algorithm                  & 0.30\%                                                 & 0.51\%                              & 0.53\%                              & 0.20\%                              & 1.51\%           & 0.30\%           & 2.52\%           & 0.80\%           \\
                                            & ~~~~$\beta_1$ (\texttt{linear})               & 6.32\%                                                 & \textbf{28.93\%}                    & 7.62\%                              & \textbf{17.43\%}                    & 6.33\%           & 15.05\%          & 6.58\%           & \textbf{19.45\%} \\
                                            & ~~~~$\beta_2$ (\texttt{sqrt})                 & 6.70\%                                                 & \textbf{26.03\%}                    & 4.92\%                              & \textbf{29.81\%}                    & 2.34\%           & \textbf{26.18\%} & 0.97\%           & \textbf{19.84\%} \\
                                            & Weighting algorithm                  & 1.98\%                                                 & 4.91\%                              & 7.99\%                              & 9.79\%                              & 8.92\%           & 8.40\%           & 8.36\%           & 8.23\%           \\
                                            & Categorical bandwidth: $b$           & 0.23\%                                                 & 6.73\%                              & 1.16\%                              & 8.50\%                              & 1.72\%           & 7.37\%           & 2.20\%           & 8.08\%           \\
          \midrule
          \multirow[]{8}{*}{JAHS-Bench-201} & Multivariate                         & 11.47\%                                                & \textbf{20.27\%}                    & \textbf{16.16\%}                    & \textbf{23.69\%}                    & \textbf{17.05\%} & \textbf{20.26\%} & \textbf{19.79\%} & \textbf{16.73\%} \\
                                            & Prior                                & \textbf{33.05\%}                                       & 5.63\%                              & \textbf{38.34\%}                    & 8.17\%                              & \textbf{40.41\%} & 8.09\%           & \textbf{41.88\%} & 7.62\%           \\
                                            & Magic clipping                       & \textbf{19.53\%}                                       & 3.75\%                              & 13.59\%                             & 5.26\%                              & 12.29\%          & 9.45\%           & 11.90\%          & 11.42\%          \\
                                            & Splitting algorithm                  & 1.16\%                                                 & 1.22\%                              & 0.94\%                              & 3.98\%                              & 1.45\%           & 6.03\%           & 1.52\%           & 5.28\%           \\
                                            & ~~~~$\beta_1$ (\texttt{linear})               & 3.24\%                                                 & \textbf{34.47\%}                    & 1.34\%                              & 12.62\%                             & 2.68\%           & 10.23\%          & 2.94\%           & 11.18\%          \\
                                            & ~~~~$\beta_2$ (\texttt{sqrt})                 & 12.84\%                                                & 20.15\%                             & 10.78\%                             & \textbf{28.97\%}                    & 8.81\%           & \textbf{25.91\%} & 5.29\%           & \textbf{28.18\%} \\
                                            & Weighting algorithm                  & 5.80\%                                                 & 5.28\%                              & 10.22\%                             & 11.52\%                             & 9.89\%           & 12.87\%          & 10.38\%          & 12.66\%          \\
                                            & Categorical bandwidth: $b$           & 12.91\%                                                & 9.22\%                              & 8.62\%                              & 5.79\%                              & 7.41\%           & 7.16\%           & 6.30\%           & 6.93\%           \\
          \bottomrule
        \end{tabular}
      }
    }
  \end{center}
  \vspace{-5mm}
\end{table}

\subsubsection{Results \& Discussion}
\label{main:experiments:section:ablation-results}
Figures~\ref{main:experiments:fig:bench-anova},\ref{main:experiments:fig:hpobench-anova} present the probability mass of each choice in the top-$50\%$ and -$5\%$ observations and Tables~\ref{main:experiments:tab:bench-hpi},\ref{main:experiments:tab:hpobench-hpi} present the HPI of each control parameter.
Appendix~\ref{appx:experiments:section:additional-results} provides the individual results to supplement the information loss by the summarization in the figures.

\paragraph{Multivariate (\texttt{multivariate}).}
The settings with the multivariate kernel yield a strong peak in the top-$5\%$ probability mass for almost all settings and the mean performance in the individual results shows that the multivariate kernel outperforms the univariate kernel except on the 10- and 30-dimensional Xin-She-Yang function.
For the benchmark functions, the multivariate kernel is one of the most dominant factors and the HPI goes up as the number of evaluations increases in the low-dimensional problems.
It matches the intuition that the multivariate kernel needs more observations to be able to exploit useful information.
Although the multivariate kernel is not essential for HPOBench, it is recommended to use the multivariate kernel.

\paragraph{Prior (\texttt{consider\_prior}).}
While the settings with the prior $p_0$ yield a strong peak in the top-$50\%$ probability mass for all the settings, it is not the case in the top-$5\%$ probability mass for the HPO benchmarks.
For HPOBench and JAHS-Bench-201, although the settings with the prior are more likely to achieve the top-$5\%$ performance in the early stage of optimizations, the likelihood decreases over time.
It implies that the prior (more exploration) is more effective in the beginning.
On the other hand, for HPOlib, the likelihood of achieving the top-$5\%$ performance is higher in the settings without the prior.
It implies that the prior (more exploitation) should be reduced in the beginning for HPOlib.
According to the individual results, while the performance distributions of the settings without the prior are close to uniform, those with the prior have a stronger mode.
It means that the prior is primarily important for the top-$50\%$ as can be seen in Tables~\ref{main:experiments:tab:bench-hpi},\ref{main:experiments:tab:hpobench-hpi} as well and the settings without the prior require more careful tuning.
Although some settings on HPOlib without the prior outperform those with the prior, we recommend using the prior because the likelihood difference is not striking.

\paragraph{Magic Clipping (\texttt{consider\_magic\_clip}).}
According to the probability mass, the magic clipping has a negative impact on the benchmark functions and a positive impact on the HPO benchmarks.
Almost no settings achieve the top-$5\%$ performance with the magic clipping for the high-dimensional benchmark functions.
The results relate to the noise level and the intrinsic cardinality discussed in Section~\ref{main:algorithm-detail:section:bandwidth}.
Since the magic clipping affects the performance strongly, we investigate it further in the next section.

\paragraph{Splitting Algorithm and $\beta$ (\texttt{gamma}).}
The choice of either \texttt{linear} or \texttt{sqrt} does not strongly affect the likelihood of achieving the top-$50\%$.
Although the HPI of the splitting algorithm is dominated by the other HPs, the splitting algorithm choice slightly affects the results.
For example, while \texttt{linear} is more effective for the 5D benchmark functions, \texttt{sqrt} is more effective for the 30D benchmark functions.
Since \texttt{linear} and \texttt{sqrt} promotes exploitation and exploration, respectively, as discussed in Section~\ref{main:algorithm-detail:section:gamma},
it might be useful to change the splitting algorithm depending on the dimensionality of the search space.
However, the choice of $\beta$ is much more important according to the tables and we, unfortunately, cannot see similar patterns in each figure although the following findings are observed:
\begin{itemize}
  \vspace{-3mm}
  \item peaks of $\beta_1$ in \texttt{linear} largely change over time,
  \vspace{-3mm}
  \item peaks of $\beta_2$ in \texttt{sqrt} do not change drastically,
  \vspace{-3mm}
  \item \texttt{linear} with a small $\beta_1$ is effective for the high-dimensional benchmark functions,
  \vspace{-3mm}
  \item \texttt{sqrt} with a large $\beta_2$ is effective for the HPO benchmarks, and
  \vspace{-3mm}
  \item the peaks change from larger $\beta$ to smaller $\beta$ over time (exploitation to exploration).
  \vspace{-3mm}
\end{itemize}
Another finding is that while the magic clipping needs to be adjusted to promote exploitation for the benchmark functions and exploration for the HPO benchmarks, $\beta$ needs to be adjusted to promote exploration (small $\beta$) for the benchmark functions and exploitation (large $\beta$) for the HPO benchmarks.
It implies that each component controls the trade-off between exploration and exploitation differently, necessitating a careful tuning of these control parameters.
However, \texttt{linear} with $\beta_1 = 0.1$ and \texttt{sqrt} with $\beta_2 = 0.75$ exhibit relatively stable performance for each problem.

\paragraph{Weighting Algorithm (\texttt{weights}).}
Although the weighting algorithm is not an important factor to attain the top-$50\%$ performance, the results show that \texttt{EI} is effective to attain the top-$5\%$ performance, and \texttt{uniform} comes next.
Note that since the behavior of \texttt{EI} heavily depends on the distribution of the objective value $p(y)$, \texttt{EI} might require special treatment on $y$ as in the scale of HPOlib.
For example, if the objective returns infinity, the weights in Eq.~(\ref{main:algorithm-detail:eq:weighting-expected-improvement}) cannot be defined anymore.
According to the top-$50\%$ probability mass of the HPO benchmarks, \texttt{old-decay} and \texttt{old-drop} are the most frequent choices, implying that they are relatively robust to the choice of other control parameters.
However, the regularization effect caused by dropping past observations limits the performance, leading to less probability mass at the top-$5\%$.

\paragraph{Categorical Bandwidth $b$.}
The categorical bandwidth with $b = 0$ achieves the top-$5\%$ performance in nearly no cases due to the overfitting to one category.
Otherwise, any choices exhibit more or less similar performance while \texttt{optuna} slightly outperforms the other choices.

\paragraph{Ablation Study Summary.} To sum up the discussion, our recommendation is to use:
\begin{itemize}
  \vspace{-2mm}
  \item \texttt{multivariate=True},
  \vspace{-2mm}
  \item \texttt{consider\_prior=True},
  \vspace{-2mm}
  \item \texttt{consider\_magic\_clip=True} for HPO problems and \texttt{False} for benchmark functions,
  \vspace{-2mm}
  \item \texttt{gamma=linear} with $\beta_1 = 0.1$ or \texttt{gamma=sqrt} with $\beta_2 = 0.75$,
  \vspace{-2mm}
  \item \texttt{weights=EI} with some processing on $y$ or \texttt{weights=uniform}, and
  \vspace{-2mm}
  \item \texttt{optuna} of the categorical bandwidth selection.
  \vspace{-2mm}
\end{itemize}
The recommended setting allows TPE to outperform Optuna v4.0.0 except for some tasks of HPOBench and HPOlib.
The next section further discusses enhancements to the bandwidth selection.

\subsection{Analysis of Bandwidth Selection}
This section investigates the effect of various bandwidth selection algorithms on the performance and provides a recommended default setting.

\subsubsection{Setup}
The experiments investigate the effect of modifications on Eq.~(\ref{main:algorithm-detail:eq:magic-clipping}).
Recall that the bandwidth $b_{\mathrm{new}} = \max(\{b, b_{\min}\})$ is used and the modified minimum bandwidth $b_{\min}$ is computed as:
\begin{equation}
  \begin{aligned}
    b_{\min} \coloneqq \max(\{\Delta (R - L), \bmagic\}),
  \end{aligned}
  \label{main:experiments:eq:magic-clipping}
\end{equation}
In the experiments, we will modify:
\begin{enumerate}
  \vspace{-2mm}
  \item the bandwidth selection heuristic, which computes $b$, discussed in Appendix~\ref{appx:algorithm-detail:section:bandwidth-selection},
  \vspace{-2mm}
  \item the minimum bandwidth factor $\Delta$,
  \vspace{-2mm}
  \item the algorithm to compute $\bmagic$, and
  \vspace{-2mm}
  \item whether to use the magic clipping (we use $\bmagic = 0$ for the non magic-clipping setting).
  \vspace{-2mm}
\end{enumerate}
The algorithm of $\bmagic$ uses $\bmagic = (R - L) / N^\alpha$ where $\alpha = 1$ is used in Optuna v4.0.0 by default.
Table~\ref{main:experiments:tab:search-space-bandwidth} shows the search space of the control parameters.
Note that $\bmagic = 0$ included in the category of Modification 4 corresponds to $\alpha = \infty$.
As the magic clipping uses a function of the observation size determined by the splitting algorithm, all eight choices in Section~\ref{main:experiments:section:ablation-study} regarding the splitting algorithm is included in the search space along with the weighting algorithms \texttt{uniform}, \texttt{EI}, and \texttt{old-decay}.
The other control parameters are fixed to the recommended setting in Section~\ref{main:experiments:section:ablation-results}.

\begin{table}[t]
  \begin{center}
    \caption{
      The search space of the control parameters used in the investigation of better bandwidth selection.
      We provide the details of the bandwidth selection heuristic in Appendix~\ref{appx:algorithm-detail:section:bandwidth-selection}.
      We have $6 \times 4 \times 3 = 72$ possible combinations for the bandwidth selection
      and $8 \times 3 = 24$ for the splitting and the weighting algorithms.
      Therefore, we evaluate $72 \times 24 = 1728$ possible combinations.
    }
    \vspace{2mm}
    \label{main:experiments:tab:search-space-bandwidth}
    \begin{tabular}{ll}
      \toprule
      Component                             & Choices                                                \\
      \midrule
      The bandwidth selection heuristic     & \{\texttt{hyperopt}, \texttt{optuna}, \texttt{scott}\} \\
      The minimum bandwidth factor $\Delta$ & \{$0.01$, $0.03$, $0.1$, $0.3$\}                               \\
      The exponent $\alpha$ for $\bmagic$            & \{$2^{-2}$, $2^{-1}$, $2^0$, $2^1$, $2^{2}$, $\infty$\}                      \\
      \bottomrule
    \end{tabular}
  \end{center}
  \vspace{-5mm}
\end{table}

\begin{figure}[t]
  \begin{center}
    \vspace{-10mm}
    \subfloat[Benchmark functions with 5D\vspace{-3mm}]{
      \includegraphics[width=0.8\textwidth]{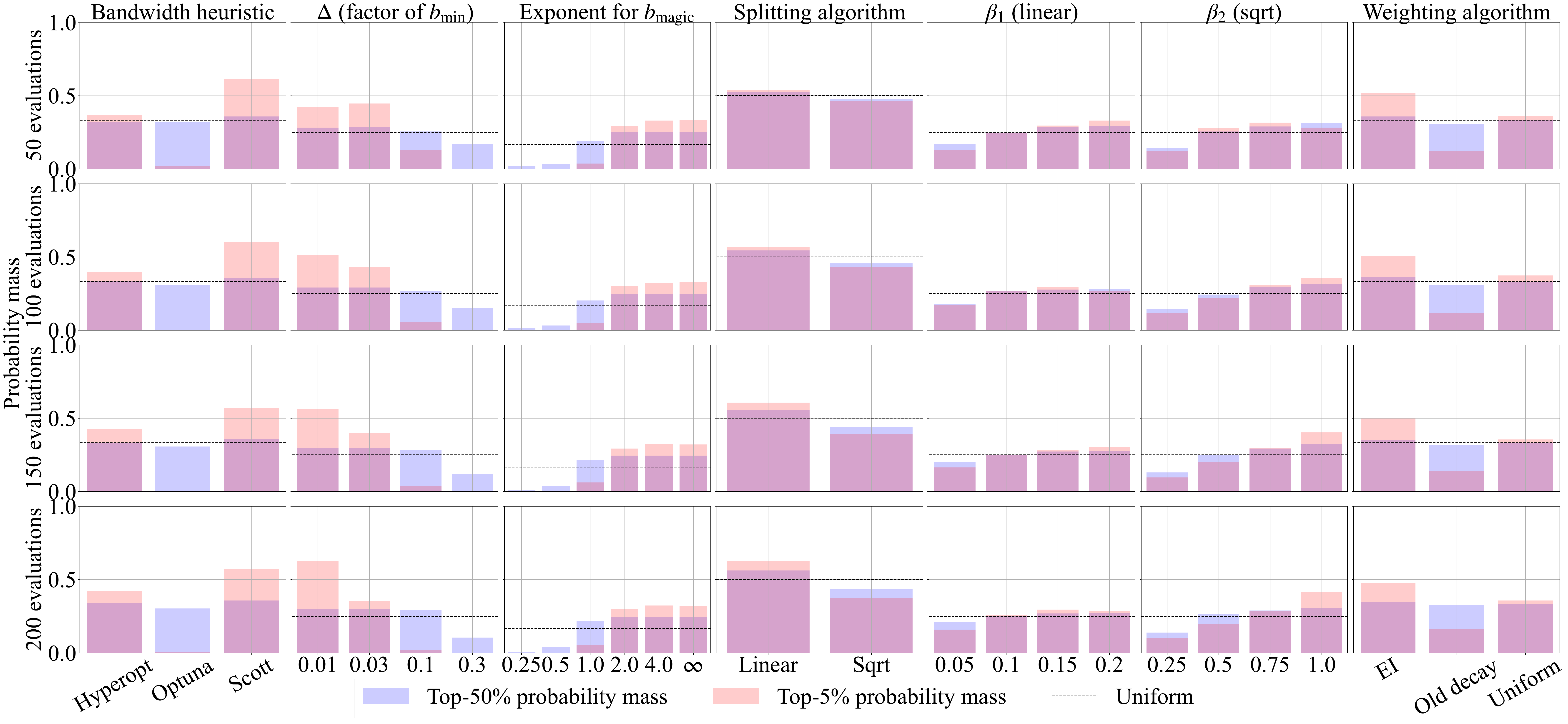}
    } \\
    \subfloat[Benchmark functions with 10D\vspace{-3mm}]{
      \includegraphics[width=0.8\textwidth]{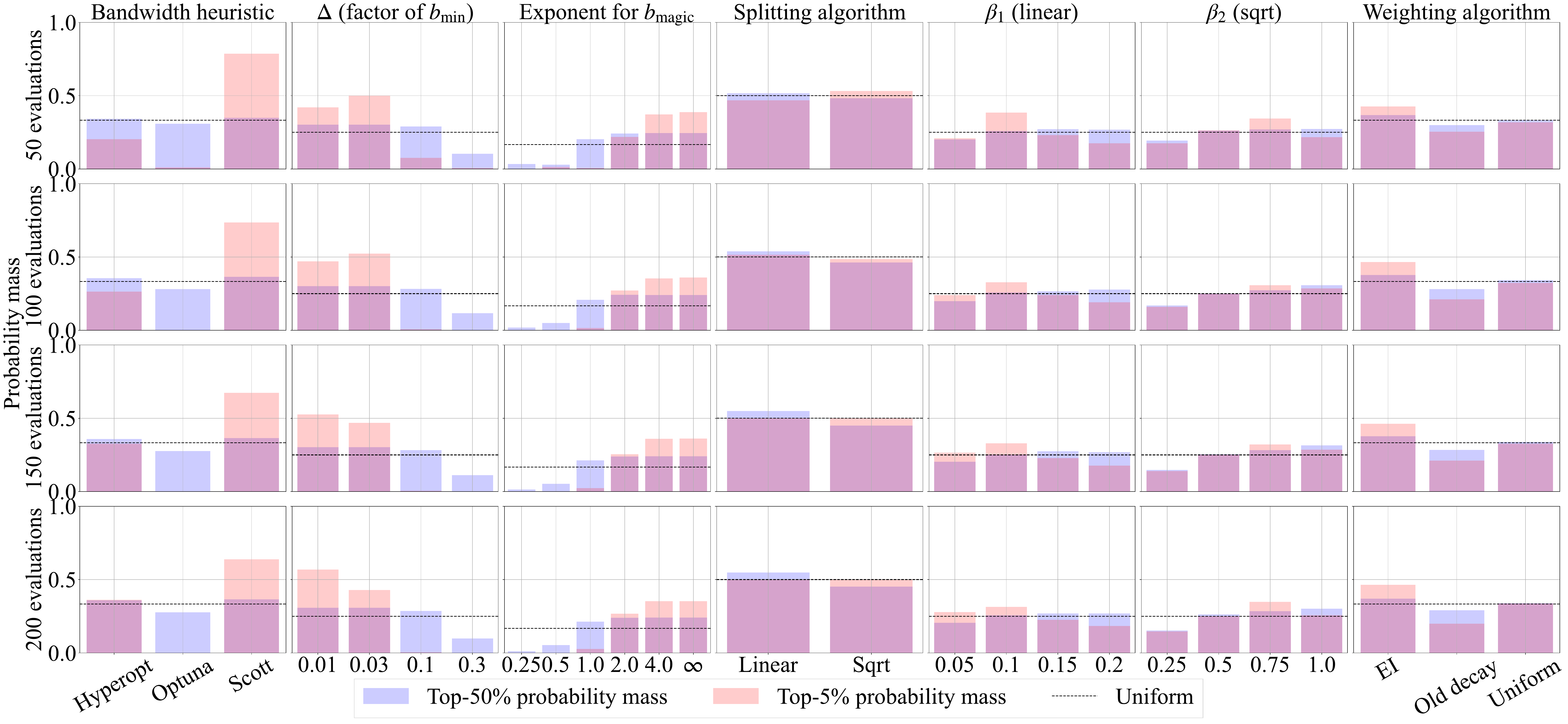}
    } \\
    \subfloat[Benchmark functions with 30D\vspace{-3mm}]{
      \includegraphics[width=0.8\textwidth]{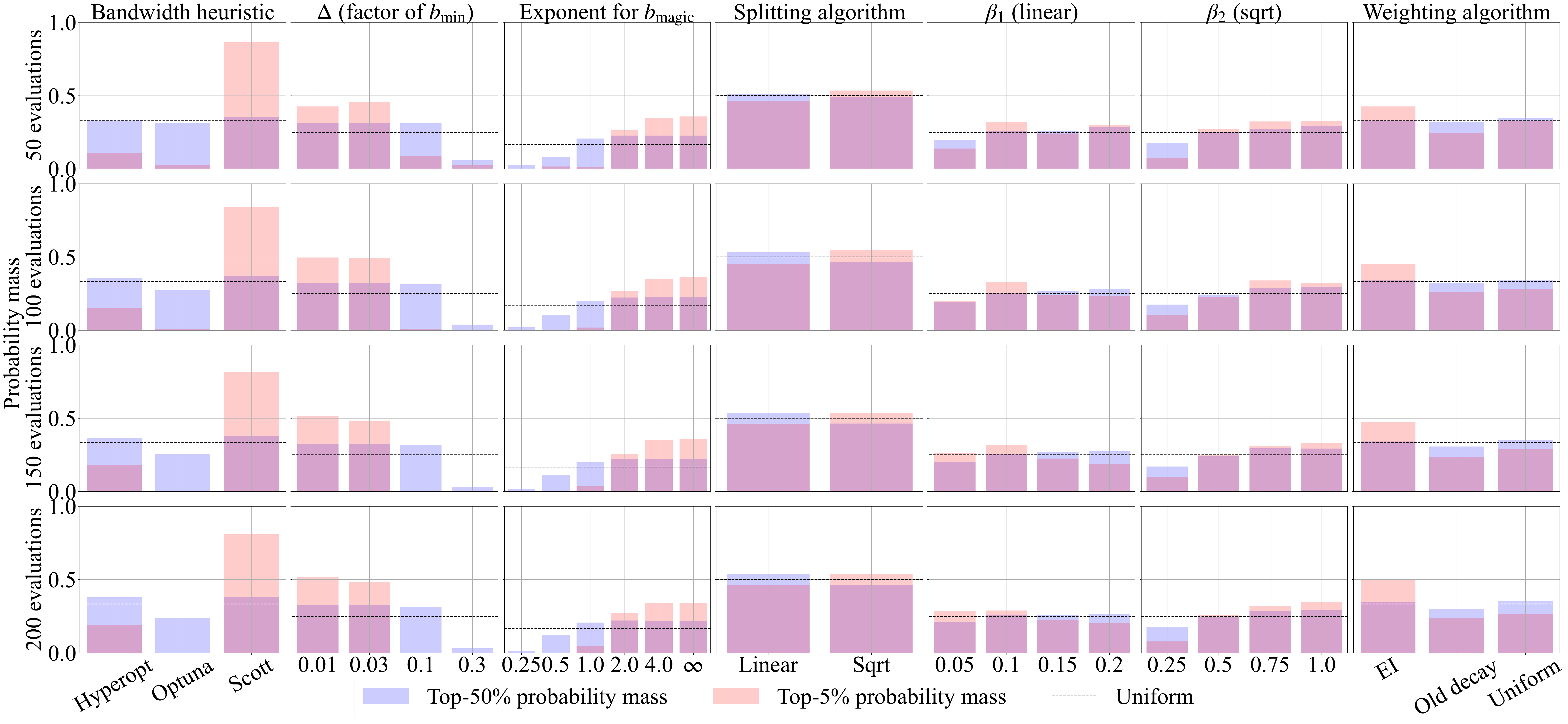}
    }
    \caption{
      The probability mass values of each control parameter for the bandwidth selection
      in the top-$5\%$ and top-$50\%$ observations
      at \{50,100,150,200\} evaluations
      on the benchmark functions.
      High probability mass in a specific value implies that
      we are likely to yield top-$5\%$ or top-$50\%$
      performance with the specific value.
    }
    \label{main:experiments:fig:bench-anova-bw}
    \vspace{-13mm}
  \end{center}
\end{figure}

\begin{table}[t]
  \begin{center}
    \caption{
      The hyperparameter importance (HPI) of each control parameter for the bandwidth selection on the benchmark functions to achieve
      top $50\%$ and to improve to top $5\%$ from top $50\%$ for the bandwidth selection.
      HPI is measured at \{50, 100, 150, 200\} evaluations.
      We bold the top-2 HPI.
      Note that while the HPI in $\beta$ quantifies whether varying $\beta$ given either \texttt{linear} or \texttt{sqrt} matters, that in ``Splitting algorithm'' quantifies whether switching between \texttt{linear} and \texttt{sqrt} matters.
    }
    \vspace{2mm}
    \label{main:experiments:tab:bench-hpi-bw}
    \makebox[1 \textwidth][c]{       
      \resizebox{1 \textwidth}{!}{   
        \begin{tabular}{c|lcc|cc|cc|cc}
          \toprule
          \multirow[]{3}{*}{Dimension} & \multirow[]{3}{*}{Control parameter} & \multicolumn{8}{c}{The number of function evaluations}                                                                                                                                                                                               \\
          \cmidrule{3-10}
                                       &                                      & \multicolumn{2}{c}{50 evaluations}                     & \multicolumn{2}{c}{100 evaluations} & \multicolumn{2}{c}{150 evaluations} & \multicolumn{2}{c}{200 evaluations}                                                                             \\
                                       &                                      & Top $50\%$                                             & Top $5\%$                           & Top $50\%$                          & Top $5\%$                           & Top $50\%$       & Top $5\%$        & Top $50\%$       & Top $5\%$        \\
          \midrule
          \multirow[]{7}{*}{5D}        & Bandwidth heuristic                  & 0.95\%                                                 & \textbf{32.15\%}                    & 1.02\%                              & \textbf{33.50\%}                    & 1.31\%           & \textbf{30.52\%} & 1.54\%           & \textbf{29.79\%} \\
                                       & $\Delta$ (factor of $b_{\min}$)      & 7.27\%                                                 & \textbf{23.56\%}                    & 11.43\%                             & \textbf{31.34\%}                    & \textbf{17.73\%} & \textbf{36.59\%} & \textbf{21.60\%} & \textbf{39.45\%} \\
                                       & Exponent for $b_{\mathrm{magic}}$    & \textbf{67.05\%}                                       & 13.83\%                             & \textbf{66.02\%}                    & 10.76\%                             & \textbf{59.48\%} & 10.37\%          & \textbf{58.83\%} & 10.99\%          \\
                                       & Splitting algorithm                  & 0.78\%                                                 & 0.56\%                              & 1.48\%                              & 0.38\%                              & 2.26\%           & 0.71\%           & 2.68\%           & 1.63\%           \\
                                       & ~~~~$\beta_1$ (\texttt{linear})      & 9.21\%                                                 & 5.83\%                              & 5.93\%                              & 5.41\%                              & 3.55\%           & 3.25\%           & 2.56\%           & 2.75\%           \\
                                       & ~~~~$\beta_2$ (\texttt{sqrt})        & \textbf{13.94\%}                                       & 11.57\%                             & \textbf{13.22\%}                    & 8.09\%                              & 15.18\%          & 7.44\%           & 12.45\%          & 6.89\%           \\
                                       & Weighting algorithm                  & 0.81\%                                                 & 12.49\%                             & 0.90\%                              & 10.50\%                             & 0.47\%           & 11.13\%          & 0.34\%           & 8.50\%           \\
          \midrule
          \multirow[]{7}{*}{10D}       & Bandwidth heuristic                  & 1.23\%                                                 & \textbf{41.90\%}                    & 2.65\%                              & \textbf{35.62\%}                    & 3.08\%           & \textbf{33.22\%} & 3.14\%           & \textbf{34.17\%} \\
                                       & $\Delta$ (factor of $b_{\min}$)      & \textbf{20.61\%}                                       & \textbf{18.65\%}                    & \textbf{18.04\%}                    & \textbf{33.10\%}                    & \textbf{18.67\%} & \textbf{32.97\%} & \textbf{22.85\%} & \textbf{33.27\%} \\
                                       & Exponent for $b_{\mathrm{magic}}$    & \textbf{67.00\%}                                       & 17.16\%                             & \textbf{63.13\%}                    & 14.19\%                             & \textbf{59.39\%} & 14.24\%          & \textbf{57.82\%} & 13.11\%          \\
                                       & Splitting algorithm                  & 0.58\%                                                 & 0.83\%                              & 1.09\%                              & 0.56\%                              & 1.77\%           & 1.40\%           & 1.68\%           & 1.20\%           \\
                                       & ~~~~$\beta_1$ (\texttt{linear})      & 3.43\%                                                 & 11.52\%                             & 3.62\%                              & 8.10\%                              & 2.71\%           & 9.35\%           & 2.39\%           & 7.75\%           \\
                                       & ~~~~$\beta_2$ (\texttt{sqrt})        & 5.75\%                                                 & 7.89\%                              & 8.34\%                              & 5.39\%                              & 11.34\%          & 6.23\%           & 10.03\%          & 6.35\%           \\
                                       & Weighting algorithm                  & 1.40\%                                                 & 2.04\%                              & 3.13\%                              & 3.04\%                              & 3.04\%           & 2.60\%           & 2.09\%           & 4.16\%           \\
          \midrule
          \multirow[]{7}{*}{30D}       & Bandwidth heuristic                  & 1.05\%                                                 & \textbf{54.42\%}                    & 3.10\%                              & \textbf{45.27\%}                    & 4.67\%           & \textbf{40.70\%} & 7.02\%           & \textbf{37.59\%} \\
                                       & $\Delta$ (factor of $b_{\min}$)      & \textbf{39.11\%}                                       & 12.94\%                             & \textbf{44.11\%}                    & \textbf{20.67\%}                    & \textbf{44.83\%} & \textbf{25.66\%} & \textbf{45.65\%} & \textbf{27.45\%} \\
                                       & Exponent for $b_{\mathrm{magic}}$    & \textbf{47.92\%}                                       & \textbf{16.94\%}                    & \textbf{41.19\%}                    & 16.66\%                             & \textbf{38.49\%} & 15.65\%          & \textbf{37.06\%} & 14.72\%          \\
                                       & Splitting algorithm                  & 0.17\%                                                 & 1.24\%                              & 0.76\%                              & 2.05\%                              & 1.04\%           & 2.31\%           & 1.21\%           & 2.13\%           \\
                                       & ~~~~$\beta_1$ (\texttt{linear})      & 4.70\%                                                 & 4.83\%                              & 3.48\%                              & 7.71\%                              & 2.62\%           & 7.83\%           & 1.82\%           & 7.33\%           \\
                                       & ~~~~$\beta_2$ (\texttt{sqrt})        & 6.76\%                                                 & 7.07\%                              & 6.88\%                              & 5.01\%                              & 7.66\%           & 4.05\%           & 6.11\%           & 5.63\%           \\
                                       & Weighting algorithm                  & 0.28\%                                                 & 2.56\%                              & 0.48\%                              & 2.63\%                              & 0.69\%           & 3.79\%           & 1.13\%           & 5.15\%           \\
          \bottomrule
        \end{tabular}
      }
    }
  \end{center}
  \vspace{-7mm}
\end{table}

\begin{figure}[t]
  \begin{center}
    \vspace{-10mm}
    \subfloat[HPOBench\vspace{-3mm}]{
      \includegraphics[width=0.84\textwidth]{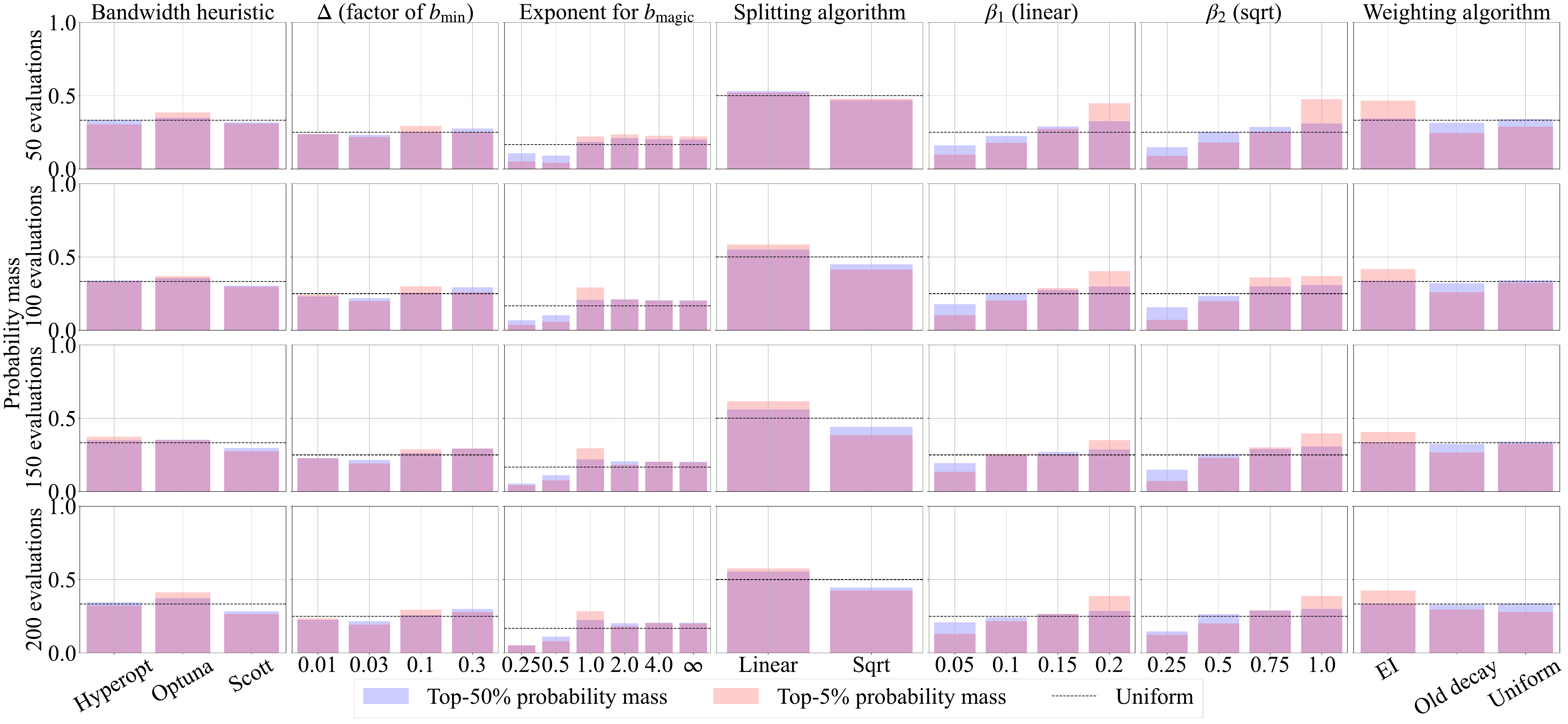}
    } \\
    \subfloat[HPOlib\vspace{-3mm}]{
      \includegraphics[width=0.84\textwidth]{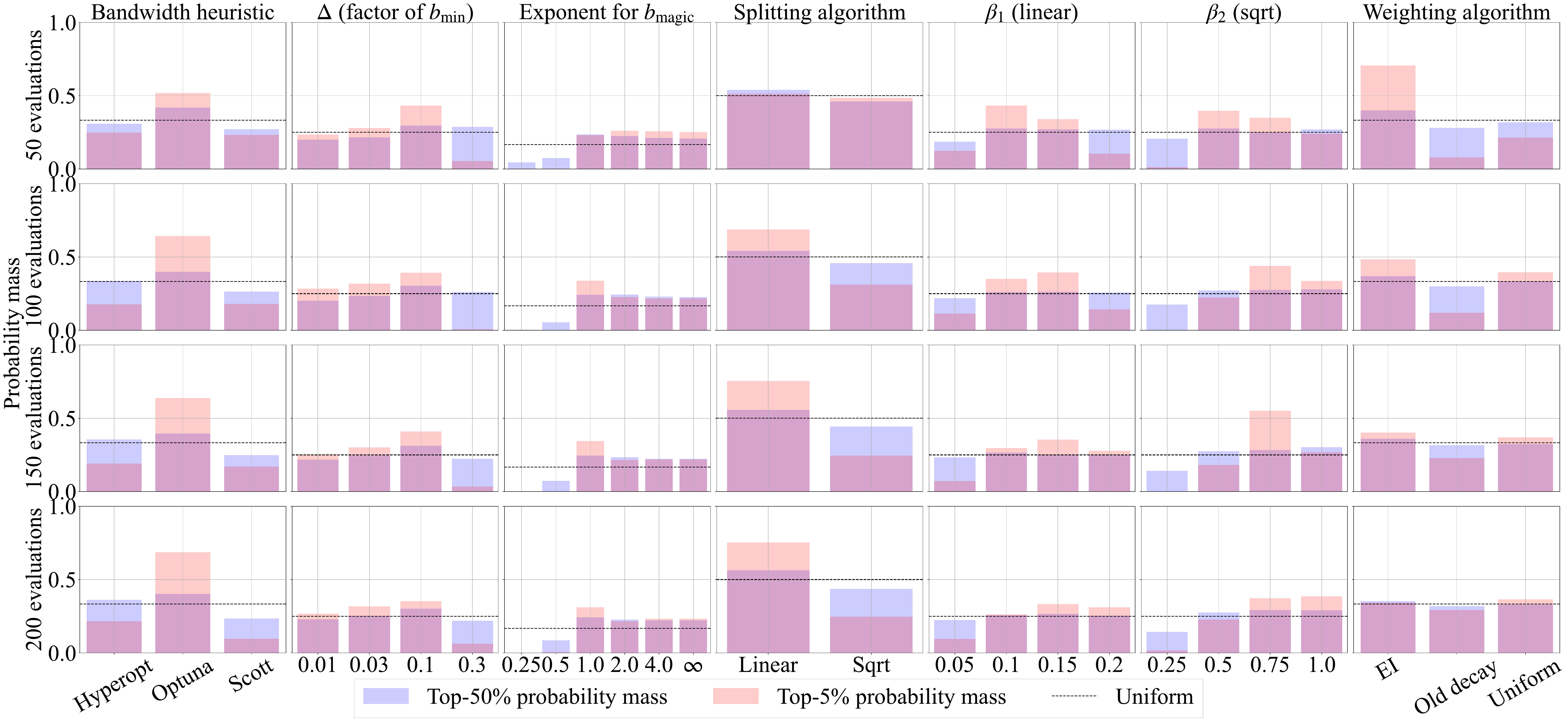}
    } \\
    \subfloat[JAHS-Bench-201\vspace{-3mm}]{
      \includegraphics[width=0.84\textwidth]{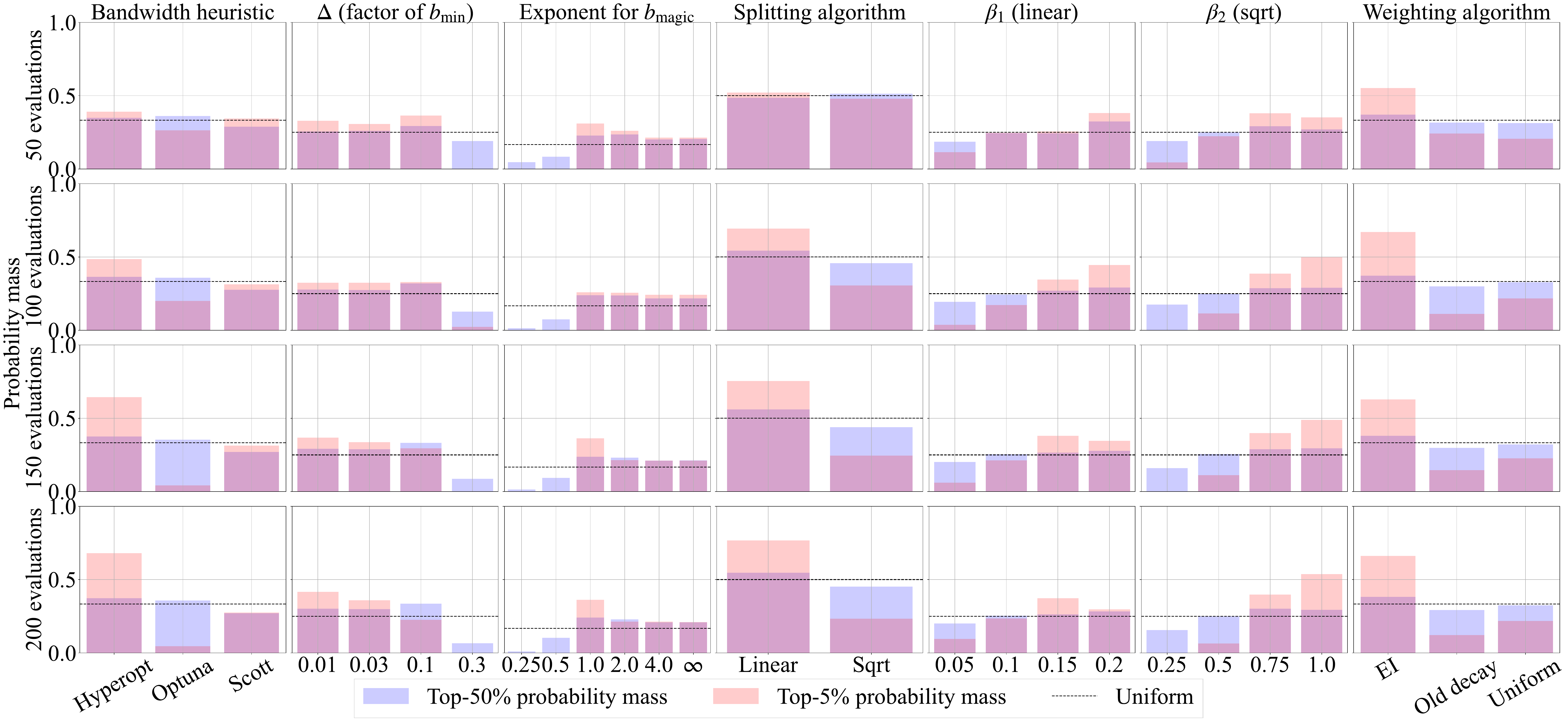}
    }
    \caption{
      The probability mass values of each control parameter for the bandwidth selection
      in the top-$5\%$ and top-$50\%$ observations
      at \{50,100,150,200\} evaluations
      on the HPO benchmarks.
      High probability mass in a specific value implies that
      we are likely to yield top-$5\%$ or top-$50\%$
      performance with the specific value.
    }
    \label{main:experiments:fig:hpobench-anova-bw}
    \vspace{-15mm}
  \end{center}
\end{figure}

\begin{table}[t]
  \begin{center}
    \caption{
      The hyperparameter importance (HPI) of each control parameter for the bandwidth selection on the HPO benchmarks to achieve top $50\%$ and to improve to top $5\%$ from top $50\%$.
      HPI is measured at \{50, 100, 150, 200\} evaluations.
      We bold the top-2 HPI.
      Note that while the HPI in $\beta$ quantifies whether varying $\beta$ given either \texttt{linear} or \texttt{sqrt} matters, that in ``Splitting algorithm'' quantifies whether switching between \texttt{linear} and \texttt{sqrt} matters.
    }
    \vspace{2mm}
    \label{main:experiments:tab:hpobench-hpi-bw}
    \makebox[1 \textwidth][c]{       
      \resizebox{1 \textwidth}{!}{   
        \begin{tabular}{c|lcc|cc|cc|cc}
          \toprule
          \multirow[]{3}{*}{Benchmark}      & \multirow[]{3}{*}{Control parameter} & \multicolumn{8}{c}{The number of function evaluations}                                                                                                                                                                                               \\
          \cmidrule{3-10}
                                            &                                      & \multicolumn{2}{c}{50 evaluations}                     & \multicolumn{2}{c}{100 evaluations} & \multicolumn{2}{c}{150 evaluations} & \multicolumn{2}{c}{200 evaluations}                                                                             \\
                                            &                                      & Top $50\%$                                             & Top $5\%$                           & Top $50\%$                          & Top $5\%$                           & Top $50\%$       & Top $5\%$        & Top $50\%$       & Top $5\%$        \\
          \midrule
          \multirow[]{7}{*}{HPOBench}       & Bandwidth heuristic                  & 1.35\%                                                 & 7.00\%                              & 2.05\%                              & 7.67\%                              & 2.94\%           & 5.98\%           & 5.05\%           & 6.42\%           \\
                                            & $\Delta$ (factor of $b_{\min}$)      & 4.16\%                                                 & 8.54\%                              & 7.25\%                              & 8.82\%                              & 7.29\%           & 7.57\%           & 8.28\%           & 8.07\%           \\
                                            & Exponent for $b_{\mathrm{magic}}$    & \textbf{36.61\%}                                       & \textbf{21.51\%}                    & \textbf{45.66\%}                    & \textbf{22.88\%}                    & \textbf{48.82\%} & \textbf{23.75\%} & \textbf{50.08\%} & \textbf{25.10\%} \\
                                            & Splitting algorithm                  & 1.98\%                                                 & 4.60\%                              & 4.10\%                              & 7.42\%                              & 6.01\%           & 3.99\%           & 5.14\%           & 1.39\%           \\
                                            & ~~~~$\beta_1$ (\texttt{linear})      & \textbf{28.42\%}                                       & 14.47\%                             & 14.35\%                             & 19.82\%                             & 9.49\%           & 17.06\%          & 7.49\%           & 18.82\%          \\
                                            & ~~~~$\beta_2$ (\texttt{sqrt})        & 26.51\%                                                & \textbf{33.87\%}                    & \textbf{25.72\%}                    & \textbf{25.85\%}                    & \textbf{25.00\%} & \textbf{33.13\%} & \textbf{23.51\%} & \textbf{26.16\%} \\
                                            & Weighting algorithm                  & 0.97\%                                                 & 10.01\%                             & 0.87\%                              & 7.55\%                              & 0.45\%           & 8.53\%           & 0.45\%           & 14.04\%          \\
          \midrule
          \multirow[]{7}{*}{HPOlib}         & Bandwidth heuristic                  & \textbf{9.53\%}                                        & 3.97\%                              & 5.83\%                              & 13.15\%                             & 7.09\%           & \textbf{16.80\%} & 9.83\%           & \textbf{27.77\%} \\
                                            & $\Delta$ (factor of $b_{\min}$)      & 8.56\%                                                 & 17.98\%                             & 5.52\%                              & \textbf{16.57\%}                    & 5.51\%           & 13.77\%          & 6.27\%           & 14.28\%          \\
                                            & Exponent for $b_{\mathrm{magic}}$    & \textbf{58.61\%}                                       & 8.98\%                              & \textbf{73.31\%}                    & 5.68\%                              & \textbf{65.36\%} & 7.53\%           & \textbf{62.96\%} & 8.18\%           \\
                                            & Splitting algorithm                  & 2.24\%                                                 & 0.29\%                              & 1.73\%                              & 4.78\%                              & 3.44\%           & 10.45\%          & 3.95\%           & 12.01\%          \\
                                            & ~~~~$\beta_1$ (\texttt{linear})      & 7.80\%                                                 & 20.76\%                             & 3.10\%                              & 13.24\%                             & 1.02\%           & 13.25\%          & 1.17\%           & 12.28\%          \\
                                            & ~~~~$\beta_2$ (\texttt{sqrt})        & 6.86\%                                                 & \textbf{22.93\%}                    & \textbf{8.68\%}                     & \textbf{37.13\%}                    & \textbf{16.79\%} & \textbf{28.33\%} & \textbf{15.37\%} & \textbf{18.11\%} \\
                                            & Weighting algorithm                  & 6.39\%                                                 & \textbf{25.10\%}                    & 1.83\%                              & 9.46\%                              & 0.78\%           & 9.86\%           & 0.45\%           & 7.37\%           \\
          \midrule
          \multirow[]{7}{*}{JAHS-Bench-201} & Bandwidth heuristic                  & 5.82\%                                                 & 7.53\%                              & 5.43\%                              & 8.66\%                              & 5.68\%           & \textbf{26.17\%} & 4.58\%           & \textbf{23.71\%} \\
                                            & $\Delta$ (factor of $b_{\min}$)      & 7.40\%                                                 & \textbf{21.60\%}                    & \textbf{17.52\%}                    & 7.27\%                              & \textbf{28.02\%} & 6.48\%           & \textbf{33.17\%} & 7.55\%           \\
                                            & Exponent for $b_{\mathrm{magic}}$    & \textbf{61.77\%}                                       & \textbf{22.37\%}                    & \textbf{58.50\%}                    & 6.91\%                              & \textbf{47.60\%} & 9.09\%           & \textbf{44.72\%} & 7.32\%           \\
                                            & Splitting algorithm                  & 0.32\%                                                 & 2.36\%                              & 1.64\%                              & 4.82\%                              & 2.93\%           & 8.21\%           & 1.63\%           & 8.05\%           \\
                                            & ~~~~$\beta_1$ (\texttt{linear})      & \textbf{14.95\%}                                       & 13.60\%                             & 6.21\%                              & 17.50\%                             & 3.31\%           & 10.23\%          & 3.01\%           & 7.99\%           \\
                                            & ~~~~$\beta_2$ (\texttt{sqrt})        & 7.59\%                                                 & 17.91\%                             & 8.77\%                              & \textbf{33.75\%}                    & 10.17\%          & \textbf{25.37\%} & 10.64\%          & \textbf{29.95\%} \\
                                            & Weighting algorithm                  & 2.15\%                                                 & 14.63\%                             & 1.93\%                              & \textbf{21.10\%}                    & 2.30\%           & 14.46\%          & 2.25\%           & 15.42\%          \\
          \bottomrule
        \end{tabular}
      }
    }
  \end{center}
\end{table}

\subsubsection{Results \& Discussion}
\label{main:experiments:section:bandwidth-results}

Figures~\ref{main:experiments:fig:bench-anova-bw},\ref{main:experiments:fig:hpobench-anova-bw}
present the probability mass of each choice in the top-$50\%$ and -$5\%$ observations
and Tables~\ref{main:experiments:tab:bench-hpi-bw},\ref{main:experiments:tab:hpobench-hpi-bw}
present the HPI of each control parameter for the bandwidth selection.
Appendix~\ref{appx:experiments:section:additional-results} provides the individual results to supplement the information loss by the summarization in the figures.

\paragraph{Minimum Bandwidth Factor $\Delta$.}
This factor is the second most important parameter for the benchmark functions and small factors $\Delta = 0.01, 0.03$ are effective due to the intrinsic cardinality as discussed in Section~\ref{main:algorithm-detail:section:bandwidth}.
Although the factor $\Delta$ is not a primarily important parameter for the HPO benchmarks, discrete search spaces may require a large $\Delta$.
Note, however, that $\Delta = 0.3$ is too large in our setups.
Although the varying noise level depending on tasks makes it impossible to generalize the optimal $\Delta$, we recommend using $\Delta = 0.03$.

\paragraph{Bandwidth Selection Heuristic.}
According to the top-$5\%$ probability mass, an appropriate bandwidth selection heuristic varies across tasks.
The best choice is \texttt{scott} for the benchmark functions, \texttt{optuna} for HPOBench and HPOlib, and \texttt{hyperopt} for JAHS-Bench-201, respectively.
Interestingly, \texttt{optuna} rarely achieves the top-$5\%$ performance on search spaces with continuous parameters.
Indeed, this finding relates to the discussion about the minimum bandwidth.
For example, $b$ calculated by \texttt{optuna} takes about $(R - L) / 10 \sim (R - L) / 5$ based on Eq.~(\ref{appx:algorithm-detail:eq:optuna-numerical-bandwidth}), matching the poor performance with the minimum bandwidth factor $\Delta \geq 0.1$.
On the other hand, \texttt{optuna}, which enforces $b \geq (R - L)/10$, outperforms the other heuristics on discrete search spaces in HPOBench and HPOlib, coinciding with the observations at the top-$5\%$ probability mass for the factor $\Delta$ peaking at $\Delta = 0.1$.
For \texttt{scott}, zero bandwidths are often observed for discrete parameters, suggesting tuning the factor $\Delta$ carefully for noisy problems when using \texttt{scott}.
While both \texttt{scott} and \texttt{optuna} exhibit clear drawbacks, \texttt{hyperopt} shows the most stable performance thanks to a flexible handling of the intrinsic cardinality, making our recommendation \texttt{hyperopt} by default.

\paragraph{Exponent $\alpha$ for $b_{\mathrm{magic}}$.}
As discussed in Section~\ref{main:experiments:section:ablation-results}, a large $\alpha$, which leads to a small $\bmagic$, and a small $\alpha$, which leads to a large $\bmagic$, are effective for the benchmark functions and the HPO benchmarks, respectively.
We recommend $\alpha = 2.0$, which is the compromise for both the benchmark functions and the HPO benchmarks based on the results.
However, the exponent for $\bmagic$ is the most important parameter, so careful tuning is still necessary.

\paragraph{Splitting and Weighting Algorithms (\texttt{gamma}, \texttt{weights}).}
The conclusion for these parameters does not change largely from Section~\ref{main:experiments:section:ablation-results} except \texttt{linear} with $\beta_1 = 0.15$ generalizes the most.

\paragraph{Ablation Study Summary.} To sum up the results of the ablation study for the bandwidth selection, our recommendation is to use:
\begin{itemize}
  \vspace{-1mm}
  \item \texttt{hyperopt} by default, \texttt{scott} for noiseless objectives, and \texttt{optuna} for discrete spaces,
  \vspace{-1.5mm}
  \item $\Delta = 0.03$ by default, small $\Delta$ ($0.01$ or $0.03$) for noiseless objectives, and large $\Delta$ ($0.03$ or $0.1$) for noisy objectives,
  \vspace{-1.5mm}
  \item $\alpha = 2$ by default, $\alpha = \infty$ and $\alpha = 1$ for noiseless and noisy objectives, respectively, and
  \vspace{-6mm}
  \item \texttt{linear} with $\beta_1 = 0.15$ (or \texttt{sqrt} with $\beta_2 = 0.75$) and \texttt{weights=EI}.
  \vspace{-1mm}
\end{itemize}
The next section validates the performance of our recommended setting.

\subsection{Comparison with Baseline Methods}

This section discusses the performance improvement by our recommended setting against various BO methods.
To generalize the assessment, our recommended setting is evaluated not only on the problem set used in the earlier sections but also on an unused problem set.

\subsubsection{Setup}
\label{main:experiments:section:comparison-setup}
The following baseline methods are used:
\begin{itemize}
  \vspace{-1mm}
  \item \texttt{BORE}~\shortcite{tiao2021bore}:
  a classifier-based BO method inspired by TPE,
  \vspace{-1mm}
  \item \texttt{HEBO}~\footnote{
    We used v0.3.2 in \url{https://github.com/huawei-noah/HEBO}.
  }~\shortcite{cowen2022hebo}:
  the winner solution of the black-box optimization challenge 2020~\shortcite{turner2021bayesian},
  \vspace{-1mm}
  \item \texttt{Random search}~\shortcite{bergstra2012random}:
  the most basic baseline method in HPO,
  \vspace{-1mm}
  \item \texttt{SMAC}~\footnote{
    We used v1.4.0 in \url{https://github.com/automl/SMAC3}.
  }~\shortcite{hutter2011sequential,lindauer2022smac3}:
  a random forest-based BO method widely used in practice, and
  \vspace{-1mm}
  \item \texttt{TurBO}~\footnote{
    We used the implementation by \url{https://github.com/uber-research/TuRBO} (Accessed on Jan 2023).
  }~\shortcite{eriksson2019scalable}:
  a recent strong baseline method used in the black-box optimization challenge 2020.
  \vspace{-1mm}
\end{itemize}
Each package is used by its default setting.
Note that since the default setting of \texttt{BORE} is not specified in the original paper~\shortcite{tiao2021bore}, we use the default setting of \texttt{BORE} in Syne Tune~\shortcite{salinas2022syne}.
More specifically, XGBoost is used as a classifier model in \texttt{BORE} and the best point is picked from $500$ random points.
\texttt{TurBO} follows the default setting in SMAC3, which uses \texttt{TurBO-1} where \texttt{TurBO-M} refers to \texttt{TurBO} with $M$ trust regions.
One-hot encoding is applied to the categorical parameters in each benchmark for \texttt{TurBO} and \texttt{HEBO}, which use the Gaussian process.
TPE uses the following control parameters:
\begin{itemize}
  \vspace{-1mm}
  \item \texttt{multivariate=True},
  \vspace{-1mm}
  \item \texttt{consider\_prior=True},
  \vspace{-1mm}
  \item \texttt{consider\_magic\_clip=True} with the exponent $\alpha = 2.0$ for $\bmagic$,
  \vspace{-1mm}
  \item \texttt{gamma=linear} with $\beta = 0.15$,
  \vspace{-1mm}
  \item \texttt{weights=EI},
  \vspace{-1mm}
  \item \texttt{optuna} for the categorical bandwidth selection,
  \vspace{-1mm}
  \item \texttt{hyperopt} for the bandwidth selection heuristic, and
  \vspace{-1mm}
  \item the minimum bandwidth factor $\Delta = 0.03$.
  \vspace{-1mm}
\end{itemize}
Furthermore, we run Optuna v4.0.0 and Hyperopt v0.2.7, both of which use TPE internally, to see the improvements.
The visualization uses the average rank of the median performance over $10$ random seeds.
To avoid the overfitting to the previously used benchmark problems, we conduct experiments using some additional benchmark functions detailed in Appendix~\ref{appx:experiments:section:detail-of-funcs}, LCBench in YAHPO Gym~\shortcite{pfisterer2022yahpo}, and the surrogate version~\footnote{\url{https://github.com/nabenabe0928/olympus-surrogate-bench/tree/main}} of Olympus~\shortcite{hickman_olympus_2023}.
Note that we call the benchmarks used in the previous sections \emph{benchmarked task set} and the additional benchmarks \emph{validation task set} hereafter.

\begin{figure}[t]
  \centering
  \includegraphics[width=0.98\textwidth]{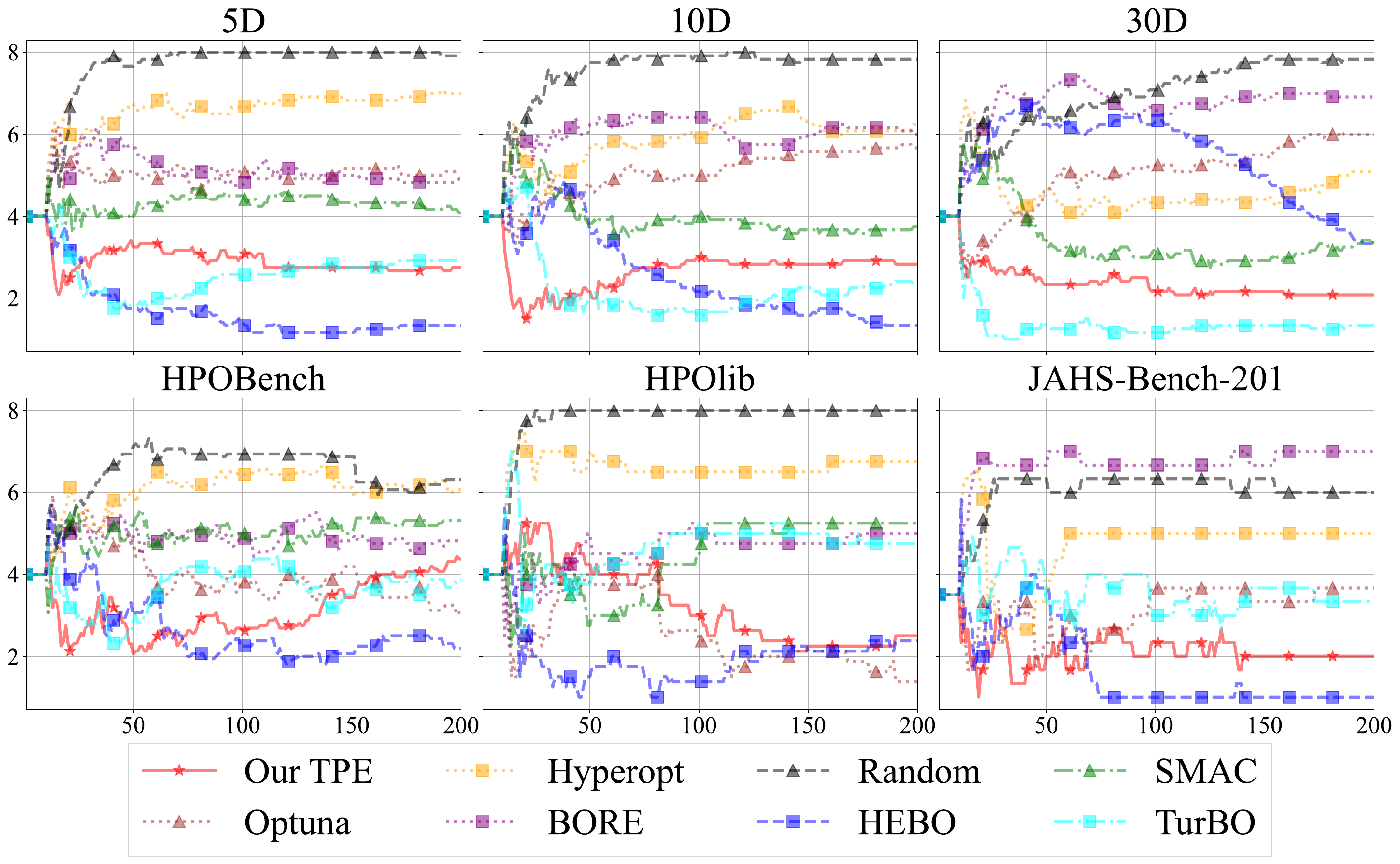}
  \vspace{-3mm}
  \caption{
    The average rank of each optimization method on the benchmarked task set.
    The initial average ranks are constant over all samplers due to the shared initial configurations.
    Note that \texttt{SMAC} is omitted for JAHS-Bench-201 due to the package dependency issue.
    The individual results are provided in Appendix~\ref{appx:experiments:section:additional-results}.
  }
  \vspace{-3mm}
  \label{main:experiments:fig:avgrank}
\end{figure}

\begin{figure}[t]
  \begin{center}
    \vspace{-10mm}
    \subfloat[\sloppy Benchmarking results on the additional benchmark functions]{
      \includegraphics[width=0.98\textwidth]{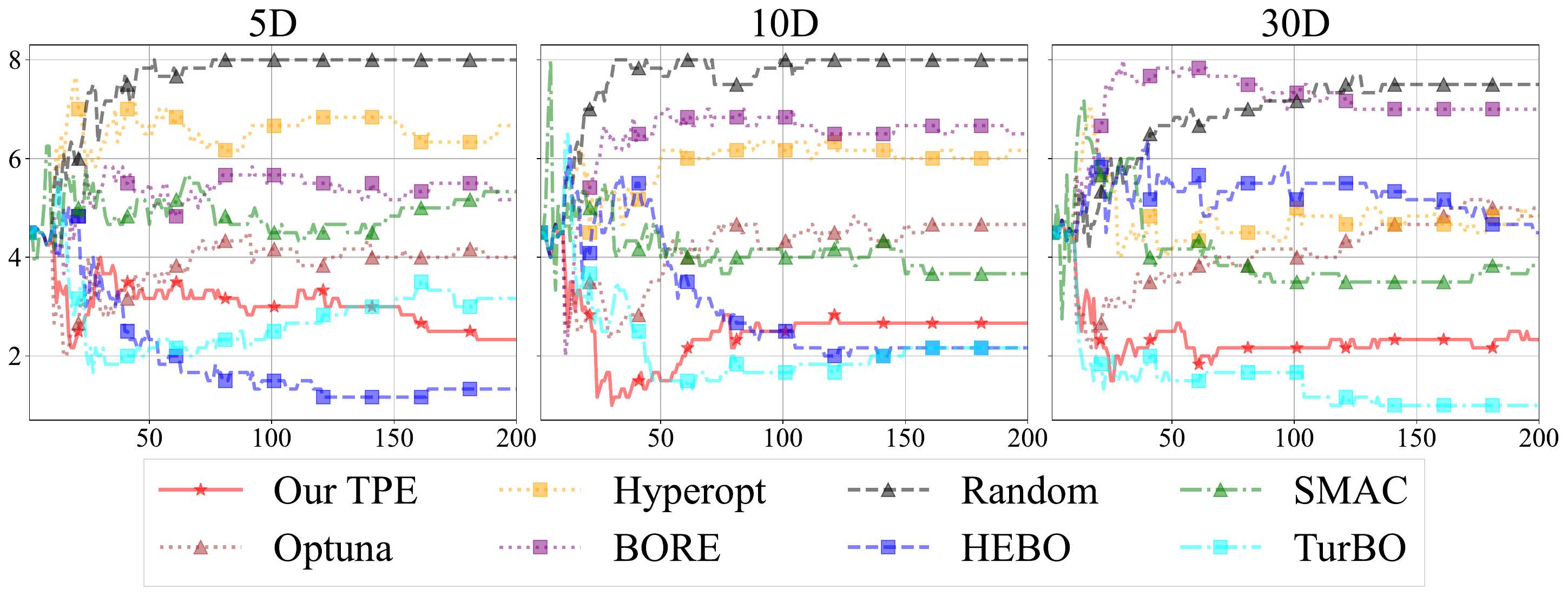}
    }\\
    \vspace{-2mm}
    \subfloat[\sloppy Benchmarking results on the additional HPO benchmarks]{
      \includegraphics[width=0.84\textwidth]{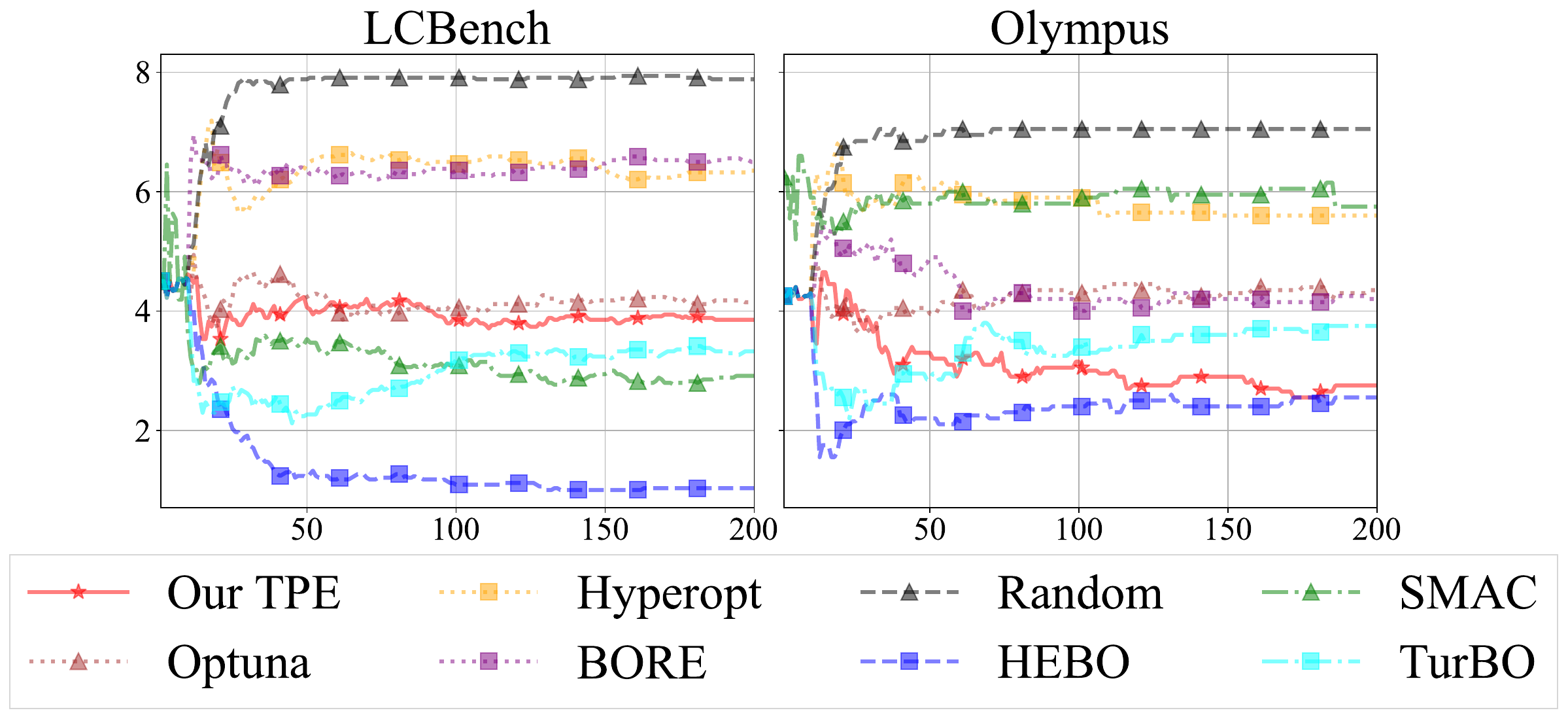}
    }
    \caption{
      The average rank of each optimization method on the validation task set.
      The initial average ranks are constant over all samplers due to the shared initial configurations.
      \label{main:experiments:fig:avgrank-extra}
    }
    \vspace{-7mm}
  \end{center}
\end{figure}

\begin{figure}[t]
  \centering
  \includegraphics[width=0.98\textwidth]{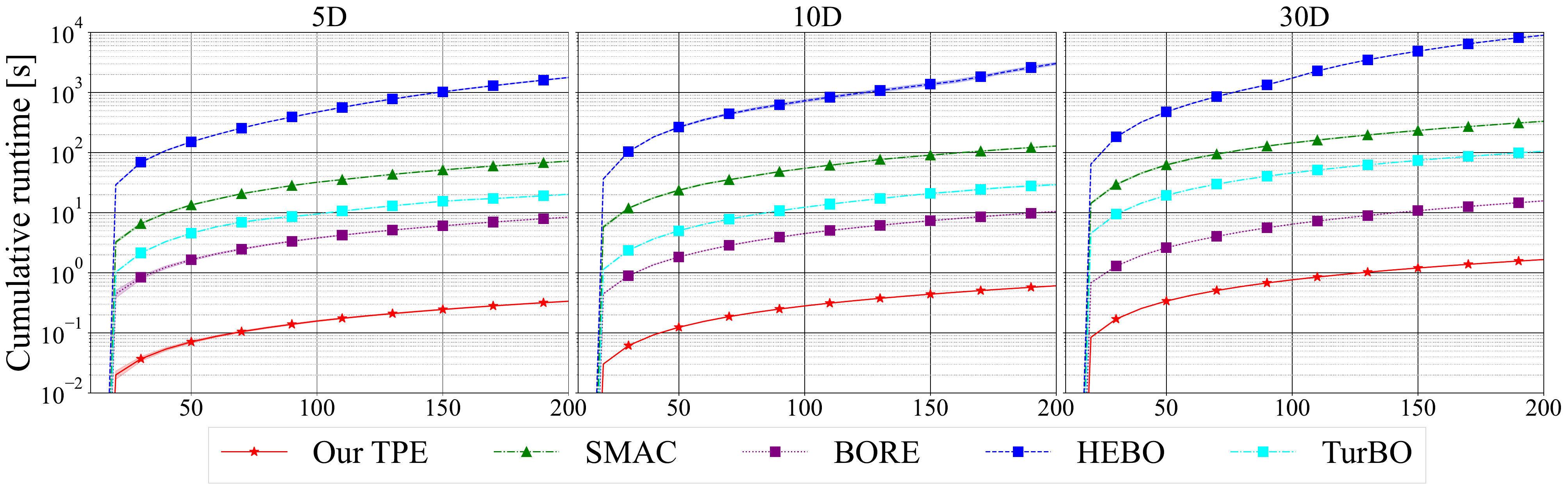}
  \vspace{-3mm}
  \caption{
    The cumulative runtime of each optimization method
    on the sphere function with different dimensionalities
    (5D for \textbf{Left}, 10D for \textbf{Center}, and 30D for \textbf{Right}).
    The horizontal axis is the number of configuration evaluations
    and the weak-color bands show the standard error over $10$ independent runs.
    Note that we used 8 cores of Intel Core i7-10700.
  }
  \vspace{-5mm}
  \label{main:experiments:fig:runtime}
\end{figure}

\begin{table}[t]
  \begin{center}
    \caption{
      The qualitative summary of the results obtained from the comparison.
      Only \texttt{HEBO}, \texttt{TurBO}, and our TPE are listed as they exhibit the best performance on at least one of the benchmark sets.
      Each method is qualitatively rated for each column by \ok~(relatively good), \soso~(medium), and \bad~(relatively bad).
    }
    \vspace{2mm}
    \label{main:experiments:tab:pros-cons-each-method}
    \begin{tabular}{l|c|cc|cc}
      \toprule
                     & \multirow[]{2}{*}{Runtime} & \multicolumn{2}{c|}{Continuous (Low/High dimensional)} & \multicolumn{2}{|c}{Discrete (W/O categorical)}                  \\
                     &                            & Low ($\sim 15\mathrm{D}$)                              & High ($15\mathrm{D} \sim$)                      & With & Without \\
      \midrule
      Our TPE        & \ok                        & \soso                                                  & \soso                                           & \ok  & \soso   \\
      \texttt{HEBO}  & \bad                       & \ok                                                    & \bad                                            & \ok  & \ok     \\
      \texttt{TurBO} & \soso                      & \soso                                                  & \ok                                             & \bad & \soso   \\
      \bottomrule
    \end{tabular}
  \end{center}
  \vspace{-5mm}
\end{table}

\subsubsection{Results \& Discussion}
Figure~\ref{main:experiments:fig:avgrank}, \ref{main:experiments:fig:avgrank-extra} present the average rank of each optimization method on the benchmarked and validation task sets, respectively.
Most importantly, the benchmarking results show very similar performance on both benchmarked and validation task sets, implying that our recommendation setting is robust to diverse tasks.
Furthermore, our TPE consistently outperforms the Hyperopt TPE, which is used by many research papers, and the Optuna TPE both on the benchmarked and validation task sets.
Notice, however, that the Optuna TPE outperforms our TPE on discrete search spaces.
The high performance of the Optuna TPE on discrete search spaces is already discussed in ``Bandwidth Selection Heuristic'' of Section~\ref{main:experiments:section:bandwidth-results}.
Although the Optuna TPE is better than our TPE on discrete search spaces, the performance of our TPE still remains comparable, suggesting that our recommendation setting is preferred.
With that being said, our TPE is not a silver bullet.
Indeed, \texttt{HEBO} consistently outperforms our TPE except for the 30D problems and \texttt{TurBO} exhibits very strong performance on the benchmark functions.
Meanwhile, our TPE is much quicker compared to \texttt{HEBO}, which takes $2.5$ hours to sample $200$ configurations on the 30D problems, as shown in Figure~\ref{main:experiments:fig:runtime}.
As the computational overhead of an objective varies, higher performance does not necessarily mean one method is better than the others.
Hence, practitioners must consider which optimization method to use based on their specific use cases.
For example, if the evaluation of an objective takes only a few seconds, \texttt{HEBO} is probably not an appropriate choice as the sampling overhead dominates the evaluation overhead.
The qualitative evaluations of \texttt{HEBO}, \texttt{TurBO}, and our TPE are summarized in Table~\ref{main:experiments:tab:pros-cons-each-method}.

\section{Conclusion}
This paper provided detailed explanations of each component in the TPE algorithm and showed how we should determine each control parameter.
Roughly speaking, the control parameter settings should be changed depending on how much noise objective functions have.
Accounting for the noise level is especially important for the bandwidth selection algorithm owing to the intrinsic cardinality.
Despite that, we provided a recommended default setting as a conclusion and compared our recommended version of TPE with various baseline methods on diverse problems.
The results demonstrated that our TPE performs much better than the existing TPE-related packages such as Hyperopt and Optuna, and potentially outperforms the state-of-the-art BO methods with much shorter computational overhead.
As this paper focused on the single-objective optimization setting, we did not discuss settings such as multi-objective, constrained, multi-fidelity, and batch optimization, and the investigation of these settings would be our future work.

\ifappendix
\clearpage
\appendix

\section{The Derivation of the Acquisition Function}
\label{appx:theoretical-details-of-tpe:section}
\shortciteA{watanabe2022ctpe,watanabe2023ctpe,song2022general} independently prove that \texttt{EI} and \texttt{PI} are equivalent under the assumption in Eq.~(\ref{main:background:eq:tpe-transform-assumption}).
For this reason, we only discuss \texttt{PI} for simplicity.
We show the detail of the transformations to obtain Eq.~(\ref{main:background:eq:density-ratio}).
We first plug in Eq.~(\ref{main:background:eq:tpe-transform-assumption}) to Eq.~(\ref{main:background:eq:probability-of-improvement})~(the formulation of \texttt{PI}) as follows:
\begin{equation}
\begin{aligned}
  \int_{-\infty}^{y^\gamma} p(y | \xv, \D)dy 
  &= \int_{-\infty}^{y^\gamma} \frac{p(\xv | y, \D)p(y|\D)}{p(\xv|\D)}dy 
  ~(\because \mathrm{Bayes' ~theorem})\\
  &= \frac{p(\xv | \Dl)}{p(\xv|\D)}
  \underbrace{\int_{-\infty}^{y^\gamma} p(y|\D) dy}_{= \gamma}
  ~(\because y \in (-\infty, y^\gamma] \Rightarrow p(\xv | y, \D) = p(\xv|\Dl))\\
  &= \frac{\gamma p(\xv | \Dl)}{\gamma p(\xv | \Dl) + (1 - \gamma)p(\xv|\Dg)}
\end{aligned}
\end{equation}
where the last transformation used the following marginalization:
\begin{equation}
\begin{aligned}
  p(\xv|\D) &= \int_{-\infty}^\infty p(\xv|y, \D) p(y|\D) dy \\
  &= p(\xv|\Dl) \int_{-\infty}^{y^\gamma} p(y|\D) dy +  
  p(\xv|\Dg) \int_{y^\gamma}^{\infty} p(y|\D) dy \\
  &= \gamma p(\xv | \Dl) + (1 - \gamma) p(\xv|\Dg).
\end{aligned}
\end{equation}

\section{Related Work of Tree-Structured Parzen Estimator}
\label{appx:related-work:section}
While this paper focuses solely on the single-objective setting, strict generalizations with multi-objective (MOTPE)~\shortcite{ozaki2020multiobjective,ozaki2022multiobjective} and constrained optimization (c-TPE)~\shortcite{watanabe2023ctpe} settings are available.
Namely, MOTPE and c-TPE are identical to the original TPE when the number of objectives is 1 and the violation probability is almost everywhere zero, i.e., $\prob[\bigcap_{i=1}^K c_i \leq c_i^\star | \xv] = 0$ for almost all~\footnote{
  More formally, almost everywhere zero is defined by $\mu(\{\xv \in \X | \prob[\bigcap_{i=1}^K c_i \leq c_i^\star | \xv] = 0\}) = 0$
  where $\mu: \mathcal{X} \rightarrow \mathbb{R}_{\geq 0}$ is the Lebesgue measure.
} $\xv \in \X$, respectively.
Note that Optuna uses the joint density generalization of c-TPE~\shortcite{watanabe2026optuna} for constrained optimization.
TPE has been adapted to the meta-learning, multi-fidelity, and combinatorial optimization settings as well.
\shortciteA{falkner2018bohb} extend TPE to the multi-fidelity setting by combining TPE and Hyperband~\shortcite{li2017hyperband}.
\shortciteA{watanabe2023speeding} introduce meta-learning by considering task similarity via the overlap of promising domains.
\shortciteA{abe2025tree} introduce a categorical kernel for TPE to handle categorical parameters more efficiently.

Furthermore, BORE~\shortcite{tiao2021bore} is inspired by TPE.
BORE replaces the density ratio with a classifier model based on the fact that TPE evaluates the promise of configurations by binary classification.
\shortciteA{song2022general} and \shortciteA{oliveira2022batch} build some theories on top of BORE.
\shortciteA{song2022general} formally derive the expected improvement for a classifier-based surrogate.
While TPE and BORE train a binary classifier with uniform weights for each sample, the expected improvement requires a binary classifier with weights proportional to the improvement from a given threshold.\footnote{This statement is not exact. See Erratum in Appendix~\ref{appx:erratum:section}.}
\shortciteA{oliveira2022batch} provide a theoretical analysis of regret using the Gaussian process classifier.
In contrast to BORE, TPE has almost no theories due to the multimodality nature of KDE and the explicit handling of density ratio.
\shortciteA{watanabe2023speeding} show that $\Dl$ asymptotically converges to a set of the $\gamma^\prime$-quantile ($\gamma^\prime < \gamma$) configurations if the objective function $f$ does not have noise, we use the $\epsilon$-greedy algorithm in Line~\ref{main:background:line:tpe-algo-greedy}, and we use a fixed $\gamma$.
To the best of our knowledge, this is the only theory available for TPE.
This analysis suggests picking a random configuration once in a while, which is, in principle, the $\epsilon$-greedy algorithm.
However, it is not realistic to use the $\epsilon$-greedy algorithm in practice due to the severely limited computational budget.

\section{More Details of Kernel Density Estimator}
This section describes the implementation details of kernel density estimator in each package.

\subsection{Uniform Weighting Algorithm in BOHB}
Suppose we have $\{x_n\}_{n=1}^N$ where $x_n \in [L, R]$, the KDE in BOHB is computed as follows:
\begin{equation}
\begin{aligned}
  \sum_{n=1}^N \frac{z_n}{\sum_{i=1}^{N} z_i} g(x, x_n | b).
\end{aligned}
\label{appx:algorithm-detail:eq:bohb-weighting}
\end{equation}
Recall that $g$ is defined in Eq.~(\ref{main:algorithm-detail:eq:gauss-kernel}) and $z_n = \int_{L}^{R} g(x, x_n)dx$.
In principle, this formulation allows to flatten PDFs even when the observations concentrate near a boundary.
Note that this formulation is not implemented in our experiments.

\subsection{\texttt{group} Parameter in Optuna}
\label{appx:algorithm-detail:section:group-param}
In this section, we use $\xnull$ as a placeholder for an undefined value and $\Rnull \coloneqq \mathbb{R} \cup \{\xnull\}$ for convenience.
We first formally define the tree-structured search space.
Suppose $\X \coloneqq \X_1 \times \X_2 \times \dots \X_D$ is a $D$-dimensional search space
with $\X_d \subseteq \Rnull$ for each $d \in [D] \coloneqq \{1,\dots, D\}$ and
the $d_i$-th ($d_i \in [D]$) dimension is conditional for each $i \in [D_c]$ where
$D_c (< D)$ is the number of conditional parameters.
Furthermore, assume binary functions $c_{d_i}: \prod_{d \neq d_i} \X_d \rightarrow \{0, 1\}$ for each $i \in [D_c]$ are given.
For example, consider the search space with (1) the number of layers $x_1 \coloneqq L \in [3] = \X_1$, and (2) the dropout rates at the $l$-th layer $x_{l+1} \coloneqq p_l \in [0, 1] = \X_{l+1}$ for $l \in [3]$,
then $x_3$ and $x_4$ are the conditional parameters in this search space.
The binary functions for this search space are $c_{3}(x_1, x_2, x_4) = \mathbb{I}[x_1 \geq 2] = \mathbb{I}[L \geq 2]$
and $c_4(x_1, x_2, x_3) = \mathbb{I}[x_1 \geq 3] = \mathbb{I}[L \geq 3]$.
Note that the dropout rate at the $l$-th layer will not be defined if there are no more than $l$ layers.
Another example is the following search space:
\begin{itemize}
  \vspace{-1mm}
  \item The number of layers $x_1 \coloneqq L \in [2] = \X_1$,
  \vspace{-2mm}
  \item The optimizer at the $1$st layer $x_{2} \in \{\mathtt{adam}, \mathtt{sgd}\} = \X_2$~\footnote{
    See https://pytorch.org/docs/stable/generated/torch.optim.Adam.html for \texttt{adam}
    and https://pytorch.org/docs/stable/generated/torch.optim.SGD.html for \texttt{sgd}.
  },
  \vspace{-2mm}
  \item The coefficient $\beta_1 \in (0, 1) = \X_3$ for \texttt{adam} in the 1st layer,
  \vspace{-2mm}
  \item The coefficient $\beta_2 \in (0, 1) = \X_4$ for \texttt{adam} in the 1st layer,
  \vspace{-2mm}
  \item The momentum $m \in (0, 1) = \X_5$ for \texttt{sgd} in the 1st layer,
  \vspace{-2mm}
  \item The optimizer at the $2$nd layer $x_{6} \in \{\mathtt{adam}, \mathtt{sgd}\} = \X_6$,
  \vspace{-2mm}
  \item The coefficient $\beta_1 \in (0, 1) = \X_7$ for \texttt{adam} in the 2nd layer,
  \vspace{-2mm}
  \item The coefficient $\beta_2 \in (0, 1) = \X_8$ for \texttt{adam} in the 2nd layer, and
  \vspace{-2mm}
  \item The momentum $m \in (0, 1) = \X_9$ for \texttt{sgd} in the 2nd layer.
  \vspace{-1mm}
\end{itemize}
All the parameters except $x_1, x_2$ are conditional in this example, making the binary functions for each dimension, $c_3 = c_4 = \mathbb{I}[x_2 = \mathtt{adam}]$, $c_5 = \mathbb{I}[x_2 = \mathtt{sgd}]$, $c_6= \mathbb{I}[x_1 \geq 2]$, $c_7 = c_8 = \mathbb{I}[x_1 \geq 2]\mathbb{I}[x_6 = \mathtt{adam}]$, and $c_9 = \mathbb{I}[x_1 \geq 2]\mathbb{I}[x_6 = \mathtt{sgd}]$.
As can be seen, $x_7, x_8, x_9$ require $x_6$ to be defined and $x_6$ requires $x_1$ to be no less than $2$.
This hierarchical structure, i.e., $x_7, x_8, x_9$ require $x_6$ and, in turn, $x_1$ as well to be defined, is the exact reason why we call search spaces with conditional parameters tree-structured.
Strictly speaking, $x_7, x_8, x_9$ cannot be provided to $c_6$, but they are unnecessary for $c_6$ thanks to the tree structure.
Hence, we simply pad $x_7, x_8, x_9$ with a placeholder $\xnull$.

Using the second example above, we will explain the \texttt{group} parameter in Optuna.
First, we define a set of dimensions as $s \subseteq [D]$ and a subspace of $\X$ that takes the dimensions specified in $s$ as $\X_s \coloneqq \prod_{d \in s}\X_d$.
Then \texttt{group} enumerates all possible $\X_s$ based on a set of observations $\D$ and optimizes the acquisition function separately in each subspace.
For example, the second example above could take
$\X_{\{1,2,3,4\}}$,
$\X_{\{1,2,5\}}$,
$\X_{\{1,2,3,4,6,7,8\}}$,
$\X_{\{1,2,3,4,6,9\}}$,
$\X_{\{1,2,5,6,7,8\}}$, and
$\X_{\{1,2,5,6,9\}}$, as subspaces when we ignore dimensions filled with a placeholder $\xnull$.
The enumeration of the subspaces can be easily reproduced by checking missing values in each observation from $\D$ with the time complexity of $O(DN)$.
Then we perform a sampling by the TPE algorithm in each subspace using the observations that belong exactly to $\X_s$.
In other words, when we have an observation $\{2, \mathtt{sgd}, \xnull, \xnull, 0.5, \mathtt{sgd}, \xnull, \xnull, 0.5\}$, we use this observation only for the sampling in $\X_{\{1,2,5,6,9\}}$, but not for that in $\X_{\{1,2,5\}}$.
Recently, \shortciteA{baba2026conditional} discussed how to quantify HP importances in a search space that includes conditional parameters using PED-ANOVA, which measures importances based on the TPE's logic.
The discussion in this paper may help improve the TPE's performance in such search spaces in the future.

\subsection{Bandwidth Selection}
\label{appx:algorithm-detail:section:bandwidth-selection}

Throughout this section, the bandwidth of KDE is assumed to be computed based on a set of observations $\{x_n\}_{n=1}^N \in [L, R]^N$ where $x_n$ is sorted such that $x_1 \leq x_2 \leq \dots \leq x_N$ holds.
If the prior is included, we simply include $x = (L + R) / 2$ in the observations.
Note that all methods select the bandwidth for each dimension independently.

\subsubsection{Hyperopt Implementation}
\label{appx:algorithm-detail:section:hyperopt-bandwidth-selection}
When \texttt{consider\_endpoints=True}, we first
augment the observations as $\{x_n\}_{n=0}^{N+1}$ where $x_0 = L, x_{N + 1} = R$;
otherwise, we just use $\{x_n\}_{n=1}^N$.
Then the bandwidth $b_n$ for the $n$-th kernel function $k(\cdot, x_n | b_n)$ ($n = 1, 2, \dots, N$)
is computed as follows:
\begin{eqnarray}
  b_n \coloneqq \left\{
  \begin{array}{ll}
    x_{n} - x_{n - 1} & (\mathrm{if}~x_{n+1}~\mathrm{not~exist}) \\
    x_{n+1} - x_{n}  & (\mathrm{if}~x_{n-1}~\mathrm{not~exist})\\
    \max(\{x_{n+1} - x_{n}, x_{n} - x_{n - 1}\}) & (\mathrm{otherwise}) \\
  \end{array}
  \right..
  \label{appx:algorithm-detail:eq:hyperopt-bandwidth-selection}
\end{eqnarray}
This heuristic adapts the bandwidth depending on the concentration of observations.
Namely, the bandwidth will be wider at sparse regions and narrower at dense regions, respectively.

\subsubsection{BOHB Implementation (Scott's Rule)}
\label{appx:algorithm-detail:section:scott-bandwidth-selection}
Scott's rule calculates bandwidth for the univariate Gaussian kernel as follows:
\begin{equation}
\begin{aligned}
  b = \biggl(
    \frac{4}{3N}
  \biggr)^{1/5} \min\biggl(
    \sigma, \frac{\mathrm{IQR}}{\Phi^{-1}(0.75) - \Phi^{-1}(0.25)}
  \biggr)
  \simeq 1.059 N^{-1/5}
  \min\biggl(
    \sigma, \frac{\mathrm{IQR}}{1.34}
  \biggr)
\end{aligned}
\label{appx:algorithm-detail:eq:scott-bandwidth}
\end{equation}
where $(4/3N)^{1/5}$ comes from Eq.~(3.28) by \shortciteA{silverman2018density},
$\mathrm{IQR}$ is the interquartile range of the observations, $\sigma$ is the standard deviation of the observations, and $\Phi: \mathbb{R} \rightarrow [0, 1]$ is the cumulative distribution of $\mathcal{N}(0, 1)$.
BOHB calculates the bandwidth for each dimension separately
and calculates the bandwidth for categorical parameters as if they are numerical parameters.

\subsubsection{Optuna v4.0.0 Implementation}
\label{appx:algorithm-detail:section:optuna-bandwidth-selection}
The bandwidth is computed in Optuna v4.0.0 as follows:
\begin{equation}
\begin{aligned}
  b = \frac{R - L}{5}N^{-1/(D + 4)}
\end{aligned}
\label{appx:algorithm-detail:eq:optuna-numerical-bandwidth}
\end{equation}
where $D$ is the dimension of search space, $N$ is the number of observations, and the target parameter is defined on $[L, R]$.
In contrast to the other methods, the bandwidth depends only on the number of observations and the dimension of search space.
The bandwidth of a categorical parameter with $C \in \mathbb{Z}_+$ categories is determined in Optuna v4.0.0 as follows:
\begin{equation}
\begin{aligned}
  b = \frac{C - 1}{N + C}
\end{aligned}
\label{appx:algorithm-detail:eq:optuna-categorical-bandwidth}
\end{equation}
where $N$ is the number of observations, and the Aitchison-Aitken kernel is used.

\begin{table}[t]
  \caption{
    The list of the benchmark functions.
    Each function can take an arbitrary dimension $D \in \mathbb{Z}_+$.
    The right column shows the domain of each function.
    One of the dimensions in $D \in \{5, 10, 30\}$ is used in the experiments.
  }
  \vspace{-2mm}
  \makebox[1 \textwidth][c]{       
    \resizebox{1 \textwidth}{!}{   
      \label{appx:experiments:tab:benchmark-funcs}
      \begin{tabular}{lll}
        \toprule
        Name            & $f(\xv)$ ($\xv \coloneqq [x_1, x_2, \dots, x_D]$) & $|x_d|\leq R$ \\
        \midrule
        Ackley          & $\ackley$                                         & $32.768$      \\
        Griewank        & $\griewank$                                       & $600$         \\
        K-Tablet        & $\ktablet$ where $K = \ceil{D/4}$                 & $5.12$        \\
        Levy            & $\levy$                                           & $10$          \\
                        & where $w_d = 1 + \frac{x_d - 1}{4}$               &               \\
        Perm            & $\perm$                                           & $1$           \\
        Rastrigin       & $\rastrigin$                                      & $5.12$        \\
        Rosenbrock      & $\rosenbrock$                                     & $5$           \\
        Schwefel        & $\schwefel$                                       & $500$         \\
        Sphere          & $\sphere$                                         & $5$           \\
        Styblinski      & $\styblinski$                                     & $5$           \\
        Weighted sphere & $\weightedsphere$                                 & $5$           \\
        Xin-She-Yang    & $\xinsheyang$                                     & $2\pi$        \\
        \bottomrule
      \end{tabular}
    }
  }
  \vspace{-3mm}
\end{table}

\begin{table}[t]
  \caption{
    The characteristics of the benchmark functions.
  }
  \vspace{2mm}
  \makebox[1 \textwidth][c]{       
    \resizebox{0.6\textwidth}{!}{   
      \label{appx:experiments:tab:benchmark-funcs-characteristics}
      \begin{tabular}{ll}
        \toprule
        Name            & Characteristics            \\
        \midrule
        Ackley          & Multimodal                 \\
        Griewank        & Multimodal                 \\
        K-Tablet        & Monomodal, ill-conditioned \\
        Levy            & Multimodal                 \\
        Perm            & Monomodal                  \\
        Rastrigin       & Multimodal                 \\
        Rosenbrock      & Monomodal                  \\
        Schwefel        & Multimodal                 \\
        Sphere          & Convex, monomodal          \\
        Styblinski      & Multimodal                 \\
        Weighted sphere & Convex, monomodal          \\
        Xin-She-Yang    & Multimodal                 \\
        \bottomrule
      \end{tabular}
    }
  }
  \vspace{-5mm}
\end{table}

\begin{table}[t]
  \begin{center}
    \caption{
      The search space of HPOBench (5 discrete parameters).
      HPOBench is a tabular benchmark
      and we can query the performance of
      a specified configuration from the tabular.
      HPOBench stores all possible configurations
      of an MLP in this table for 8 different
      OpenML datasets (\texttt{Vehicle},
      \texttt{Segmentation},
      \texttt{Car evaluation},
      \texttt{Australian credit approval},
      \texttt{German credit},
      \texttt{Blood transfusion service center},
      \texttt{KC1 software detect prediction},
      \texttt{Phoneme}).
      Each parameter except ``Depth'' has 10 grids
      and the grids are taken so that
      each grid is equally distributed in the log-scaled
      range.
    }
    \vspace{2mm}
    \label{appx:experiments:tab:search-space-hpobench}
    \begin{tabular}{lll}
      \toprule
      Hyperparameter        & Parameter type & Range          \\
      \midrule
      L2 regularization     & Discrete       & $[10^{-8}, 1]$ \\
      Batch size            & Discrete       & $[4, 256]$     \\
      Depth                 & Discrete       & $[1, 3]$       \\
      Initial learning rate & Discrete       & $[10^{-5}, 1]$ \\
      Width                 & Discrete       & $[16, 1024]$   \\
      \bottomrule
    \end{tabular}
  \end{center}
  \vspace{-2mm}
\end{table}

\begin{table}[t]
  \begin{center}
    \caption{
      The search space of HPOlib (6 discrete + 3 categorical parameters).
      HPOlib is a tabular benchmark
      and we can query the performance of
      a specified configuration from the tabular.
      HPOlib stores all possible configurations
      of an MLP in this table for 4 different
      datasets (\texttt{Parkinsons telemonitoring},
      \texttt{Protein structure},
      \texttt{Naval propulsion},
      \texttt{Slice localization}).
    }
    \vspace{2mm}
    \label{appx:experiments:tab:search-space-hpolib}
    \makebox[1 \textwidth][c]{       
      \resizebox{1 \textwidth}{!}{   
        \begin{tabular}{lll}
          \toprule
          Hyperparameter               & Parameter type & Range                                                                                 \\
          \midrule
          Initial learning rate        & Discrete       & $\{5 \times 10^{-4}, 10^{-3}, 5 \times 10^{-3}, 10^{-2}, 5 \times 10^{-2}, 10^{-1}\}$ \\
          Dropout rate \{1, 2\}        & Discrete       & $\{0.0, 0.3, 0.6\}$                                                                   \\
          Number of units \{1, 2\}     & Discrete       & $\{2^4, 2^5, 2^6, 2^7, 2^8, 2^9\}$                                                    \\
          Batch size                   & Discrete       & $\{2^3, 2^4, 2^5, 2^6\}$                                                              \\
          Learning rate scheduling     & Categorical    & \{\texttt{cosine}, \texttt{constant}\}                                                \\
          Activation function \{1, 2\} & Categorical    & \{\texttt{ReLU}, \texttt{tanh}\}                                                      \\
          \bottomrule
        \end{tabular}
      }}
  \end{center}
  \vspace{-5mm}
\end{table}

\begin{table}[t]
  \begin{center}
    \caption{
      The search space of JAHS-Bench-201 (2 continuous + 2 discrete + 8 categorical parameters).
      JAHS-Bench-201 is a surrogate benchmark
      and it uses an XGBoost surrogate
      model trained on pre-evaluated configurations
      for 3 different datasets (\texttt{CIFAR10},
      \texttt{Fashion MNIST},
      \texttt{Colorectal histology}).
      We use the output of the XGBoost surrogate
      given a specified configuration as a query.
    }
    \vspace{2mm}
    \label{appx:experiments:tab:search-space-jahs}
    \makebox[1 \textwidth][c]{       
      \resizebox{1 \textwidth}{!}{   
        \begin{tabular}{lll}
          \toprule
          Hyperparameter                                      & Parameter type & Range                                                \\
          \midrule
          Learning rate                                       & Continuous     & $[10^{-3}, 1]$                                       \\
          L2 regularization                                   & Continuous     & $[10^{-5}, 10^{-2}]$                                 \\
          Depth multiplier                                    & Discrete       & $\{1, 2, 3\}$                                        \\
          Width multiplier                                    & Discrete       & $\{2^2, 2^3, 2^4\}$                                  \\
          Cell search space                                   & Categorical    & \{\texttt{none}, \texttt{avg-pool-3x3}, \texttt{bn-conv-1x1}, \\
          (NAS-Bench-201~\shortcite{dong2020bench}, Edge 1 -- 6) &                & \: \texttt{bn-conv-3x3}, \texttt{skip-connection}\}            \\
          Activation function                                 & Categorical    & \{\texttt{ReLU}, \texttt{Hardswish}, \texttt{Mish}\}                            \\
          Trivial augment~\shortcite{muller2021trivialaugment}   & Categorical    & \{\texttt{True}, \texttt{False}\}                                      \\
          \bottomrule
        \end{tabular}
      }}
  \end{center}
\end{table}

\section{The Details of Benchmarks}
\label{appx:experiments:section:detail-of-funcs}
Table~\ref{appx:experiments:tab:benchmark-funcs} lists the $12$ different benchmark functions used in the experiments with $3$ different dimensionalities.
Their characteristics are available in Table~\ref{appx:experiments:tab:benchmark-funcs-characteristics}.
For the HPO benchmarks, HPOBench~\shortcite{eggensperger2021hpobench} in Table~\ref{appx:experiments:tab:search-space-hpobench}, HPOlib~\shortcite{klein2019tabular} in Table~\ref{appx:experiments:tab:search-space-hpolib}, and JAHS-Bench-201~\shortcite{bansal2022jahs} in Table~\ref{appx:experiments:tab:search-space-jahs} are used.
The validation task set includes LCBench in YAHPO Gym, the surrogate version of Olympus, and 6 benchmark functions.
LCBench provides the seven-dimensional continuous search space for $33$ different datasets and the Olympus surrogate benchmark provides $10$ chemistry continuous BBO problems each with a different dimensionality.
The benchmark functions used in the validation task set are different power, Dixon-Price, Langermann, Michalewicz, Powell, and Trid.
HPO benchmarks have two types:
\begin{enumerate}
  \item \textbf{tabular benchmark}, which queries pre-recorded performance metric values from a static table which is why it cannot handle continuous parameters, and
  \item \textbf{surrogate benchmark}, which returns the predicted performance metric values by a surrogate model for the corresponding HP configurations.
\end{enumerate}
HPOlib and HPOBench are tabular benchmarks and JAHS-Bench-201 is a surrogate benchmark.
The visualization for Figures~\ref{main:experiments:fig:bench-anova} -- \ref{main:experiments:fig:hpobench-anova-bw} uses the methodologies invented by \shortciteA{watanabe2023pedanova}.
Assume there are $K$ possible combinations of control parameters
and a set of observations $\{(\xv_{k,n}^{(m)}, y_{k,n}^{(m)})\}_{n=1}^{N}$ are obtained on the $m$-th benchmark ($m = 1, \dots, M$) with the $k$-th possible set of the control parameters $\thetav_k$.
$\thetav_k$ ($k = 1, \dots, K$) is one of the possible sets of the control parameters specified in Table~\ref{main:experiments:tab:search-space-ablation} and $N = 200$ is used in the experiments.
Then we collect a set of results $\mathcal{R}_{n}^{(m)} \coloneqq \{(\thetav_k, \zeta^{(m)}_{k,n})\}_{k=1}^K$ with the budget of $n$ where $\zeta_{k,n}^{(m)} \coloneqq \min_{n^\prime \leq n} y_{k,n^\prime}^{(m)}$ and $n \in \{50, 100, 150, 200\}$ are used in the experiments.
Furthermore, we define $i_{m,k}$ ($k = 1, \dots, K$)
such that $\zeta^{(m)}_{i_{m,1},n} \leq \zeta^{(m)}_{i_{m,2},n} \leq \dots \leq \zeta^{(m)}_{i_{m,K},n}$.
The visualization of the probability mass functions performs the following:
\begin{enumerate}
  \vspace{-1mm}
  \item Pick a top-performance quantile $\alpha$ (in our case, $\alpha = 0.05, 0.5$),
  \vspace{-1mm}
  \item Extract the top-$\alpha$ quantile observations $\{(\thetav_{i_{m,k}}, \zeta^{(m)}_{i_{m,k}, n})\}_{k=1}^{\ceil{\alpha K}}$,
  \vspace{-1mm}
  \item Build 1D KDEs $p_d^{(m)}(\theta_d) \coloneqq \sum_{k=1}^{\ceil{\alpha N}} k(\theta_d, \theta_{i_{m,k},d})$ for each control parameter,
  \vspace{-1mm}
  \item Compute the mean of the KDEs from all the tasks $\bar{p}_d \coloneqq 1/M \sum_{m=1}^{M} p_d^{(m)}(\theta_d)$,
  \vspace{-1mm}
  \item Plot the probability mass function of the mean $\bar{p}_d$ of the KDEs.
  \vspace{-1mm}
\end{enumerate}
The hyperparameter importance (HPI) is computed by PED-ANOVA~\shortcite{watanabe2023pedanova}.
The top-$50\%$ HPI captures the importance of each control parameter to achieve the top-$50\%$ performance (global HPI) while the top-$5\%$ HPI captures the importance of each control parameter to achieve the top-$5\%$ performance from the top-$50\%$ performance (local HPI).

\begin{figure}[t]
  \vspace{-10mm}
  \begin{center}
    \subfloat[Ackley with 5D (\textbf{Top row}), 10D (\textbf{Middle row}), and 30D (\textbf{Bottom row})]{
      \includegraphics[width=0.95\textwidth]{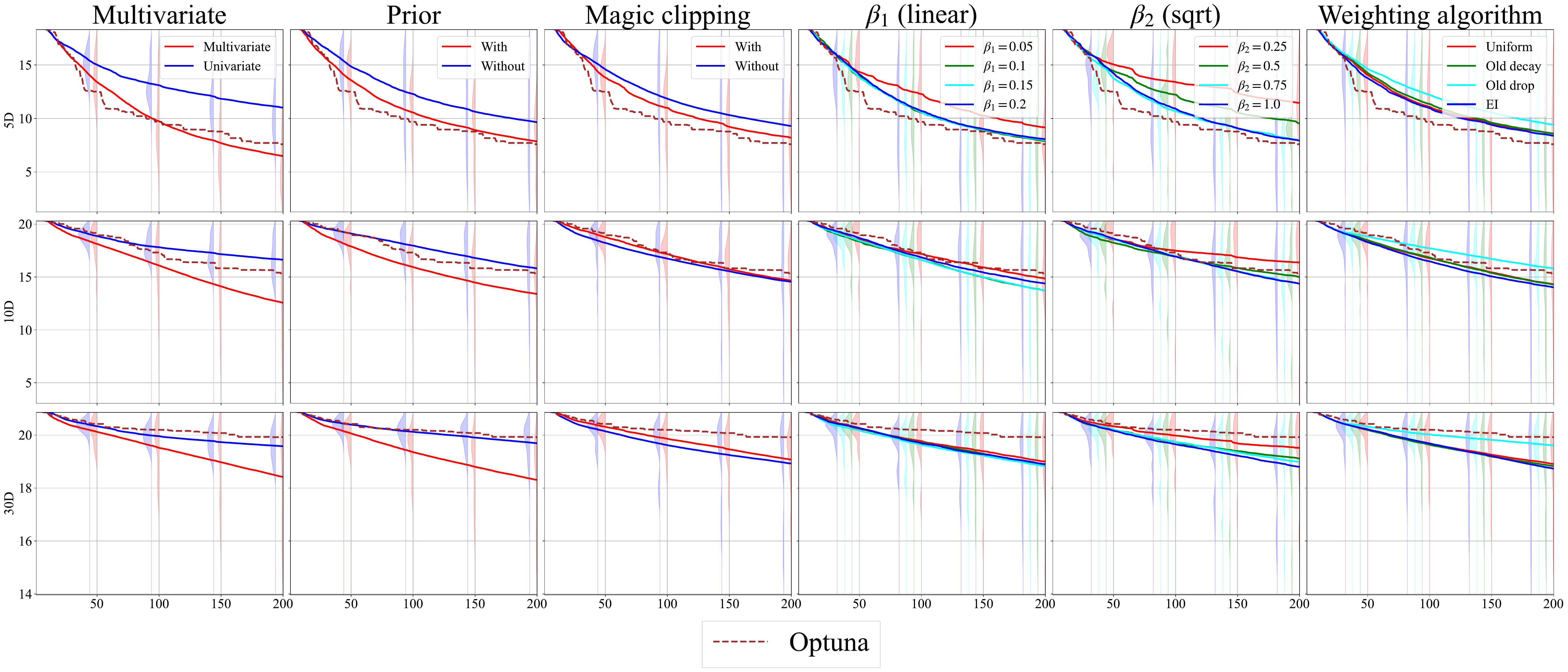}
    }\\
    \vspace{-3mm}
    \subfloat[Griewank with 5D (\textbf{Top row}), 10D (\textbf{Middle row}), and 30D (\textbf{Bottom row})]{
      \includegraphics[width=0.95\textwidth]{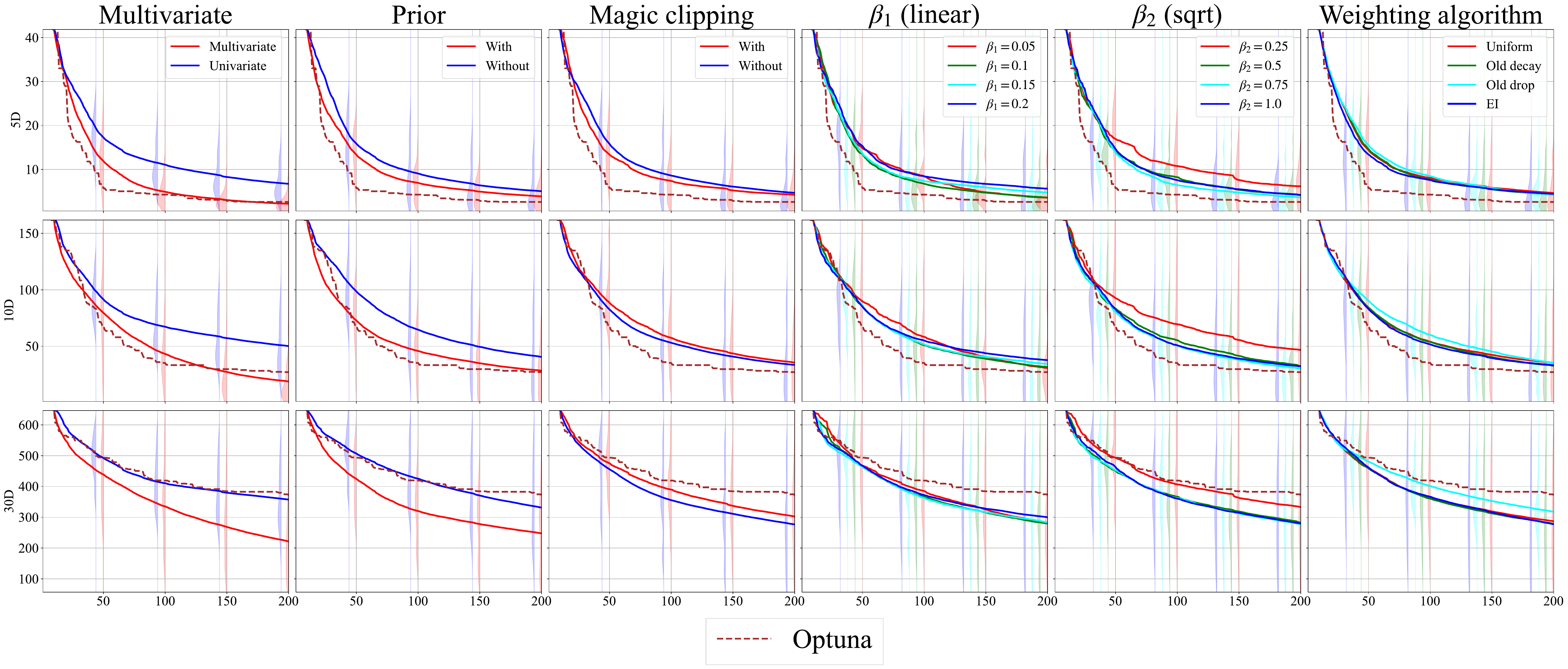}
    }\\
    \vspace{-3mm}
    \subfloat[K-Tablet with 5D (\textbf{Top row}), 10D (\textbf{Middle row}), and 30D (\textbf{Bottom row})]{
      \includegraphics[width=0.95\textwidth]{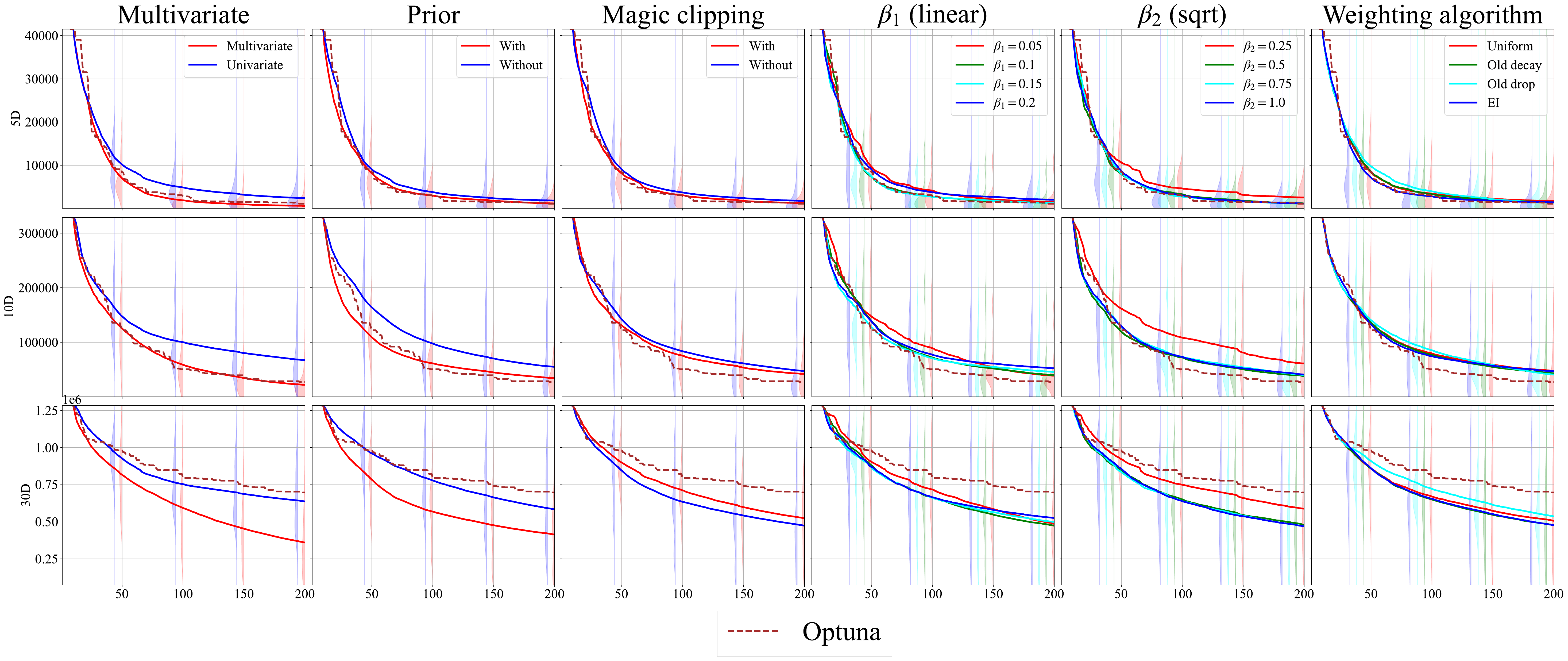}
    }
    \vspace{-2mm}
    \caption{
      The ablation study on benchmark functions.
      The $x$-axis is the number of evaluations and the $y$-axis is the cumulative minimum objective.
      The solid lines in each figure show the mean of the cumulative minimum objective over all control parameter configurations.
      The transparent shades represent the distributions of the cumulative minimum objective at $\{50,100,150,200\}$ evaluations.
      The performance of Optuna v4.0.0 (brown dotted lines) is provided as a baseline.
    }
    \label{appx:experiments:fig:ablation-bench00}
  \end{center}
\end{figure}

\begin{figure}[p]
  \vspace{-10mm}
  \begin{center}
    \subfloat[Levy with 5D (\textbf{Top row}), 10D (\textbf{Middle row}), and 30D (\textbf{Bottom row})]{
      \includegraphics[width=0.95\textwidth]{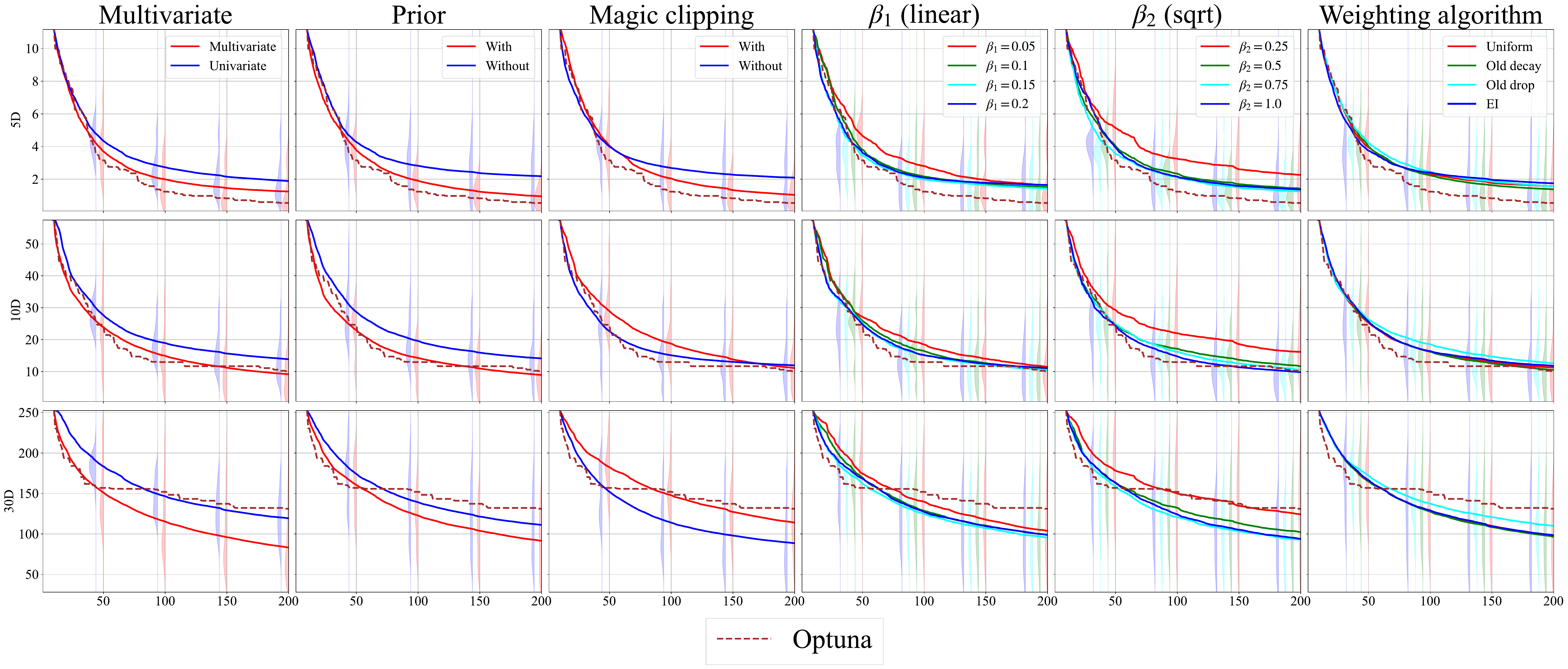}
    }\\
    \vspace{-3mm}
    \subfloat[Perm with 5D (\textbf{Top row}), 10D (\textbf{Middle row}), and 30D (\textbf{Bottom row})]{
      \includegraphics[width=0.95\textwidth]{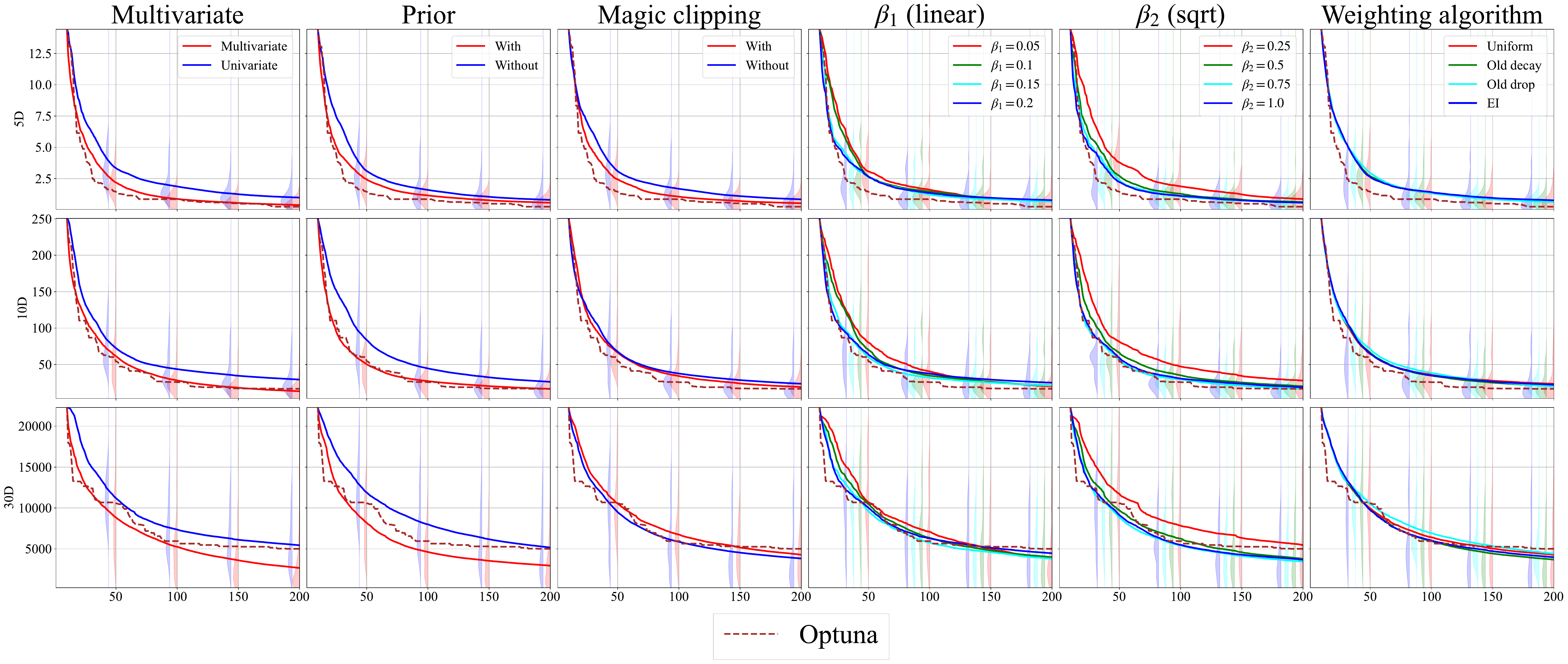}
    }\\
    \vspace{-3mm}
    \subfloat[Rastrigin with 5D (\textbf{Top row}), 10D (\textbf{Middle row}), and 30D (\textbf{Bottom row})]{
      \includegraphics[width=0.95\textwidth]{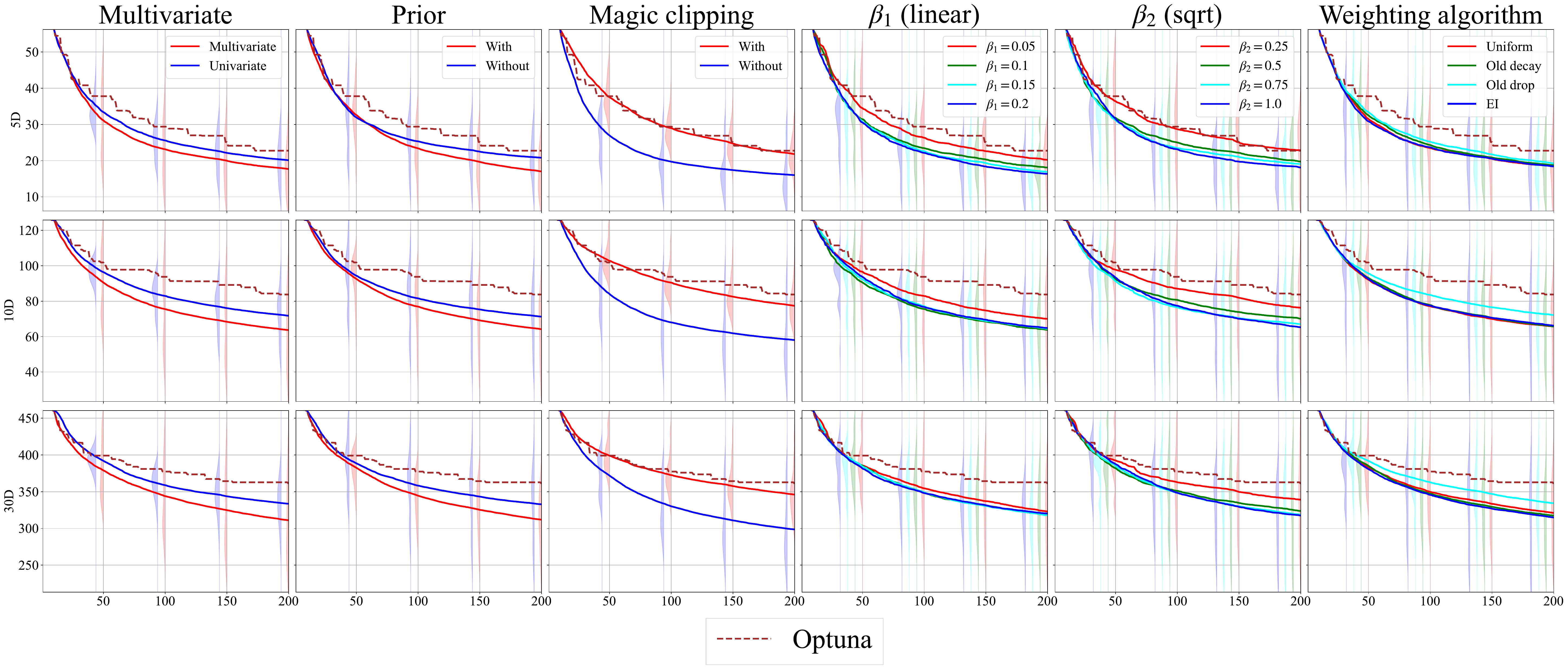}
    }
    \vspace{-2mm}
    \caption{
      The ablation study on benchmark functions.
      The $x$-axis is the number of evaluations and the $y$-axis is the cumulative minimum objective.
      The solid lines in each figure show the mean of the cumulative minimum objective over all control parameter configurations.
      The transparent shades represent the distributions of the cumulative minimum objective at $\{50,100,150,200\}$ evaluations.
      The performance of Optuna v4.0.0 (brown dotted lines) is provided as a baseline.
    }
    \label{appx:experiments:fig:ablation-bench01}
  \end{center}
\end{figure}

\begin{figure}[p]
  \vspace{-10mm}
  \begin{center}
    \subfloat[Rosenbrock with 5D (\textbf{Top row}), 10D (\textbf{Middle row}), and 30D (\textbf{Bottom row})]{
      \includegraphics[width=0.95\textwidth]{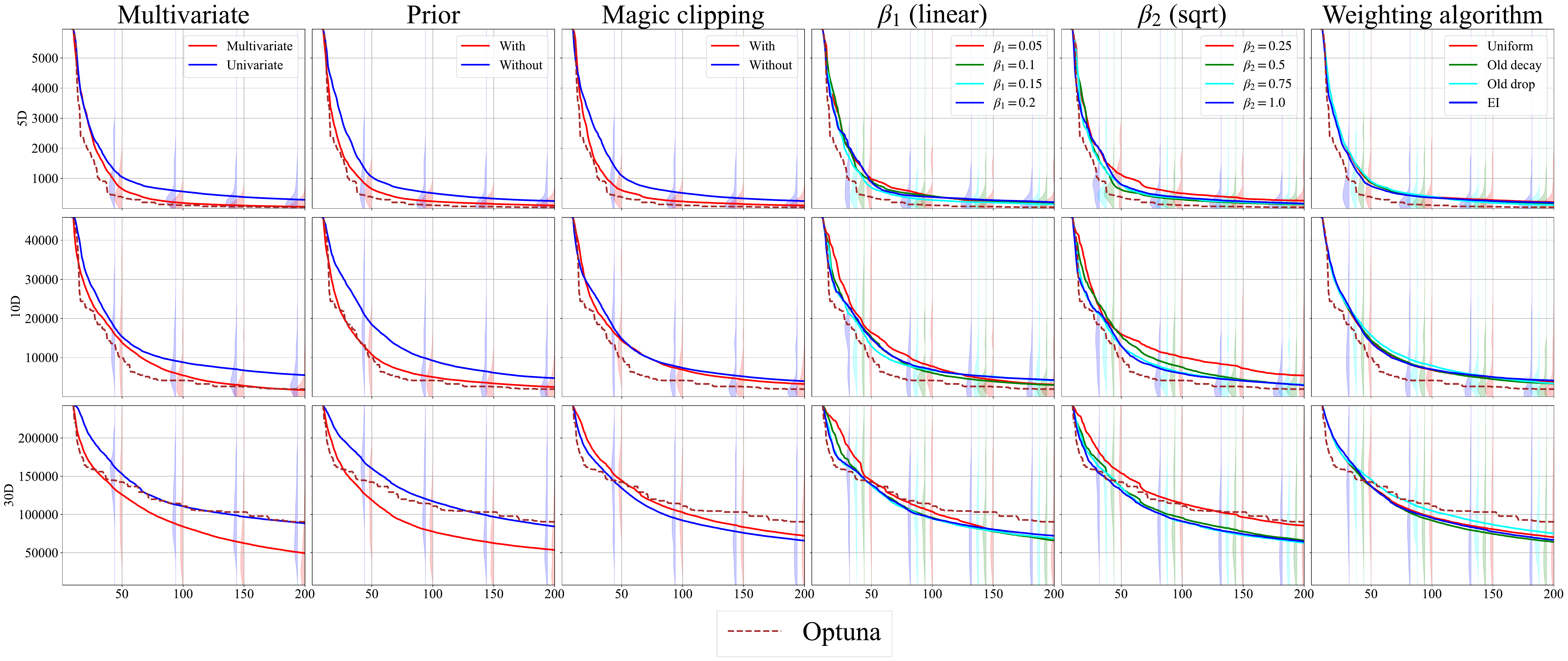}
    }\\
    \vspace{-3mm}
    \subfloat[Schwefel with 5D (\textbf{Top row}), 10D (\textbf{Middle row}), and 30D (\textbf{Bottom row})]{
      \includegraphics[width=0.95\textwidth]{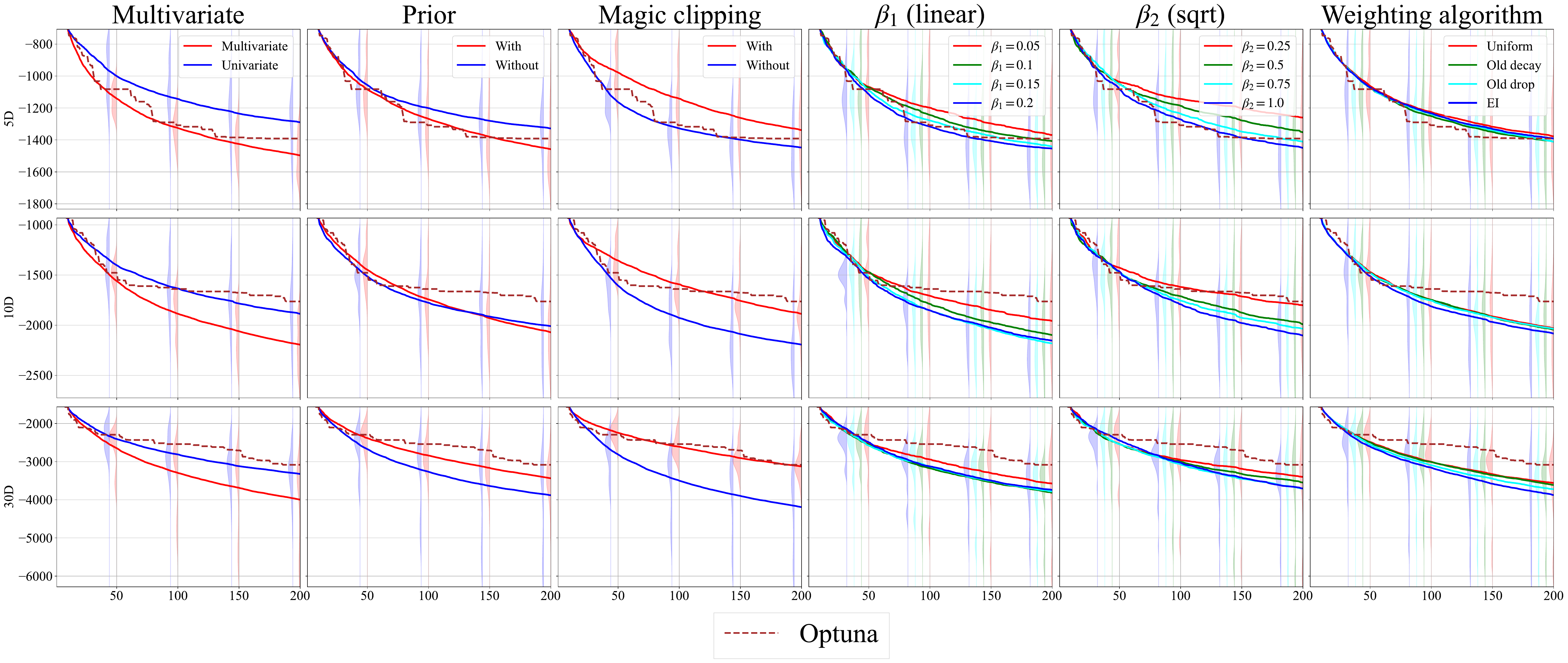}
    }\\
    \vspace{-3mm}
    \subfloat[Sphere with 5D (\textbf{Top row}), 10D (\textbf{Middle row}), and 30D (\textbf{Bottom row})]{
      \includegraphics[width=0.95\textwidth]{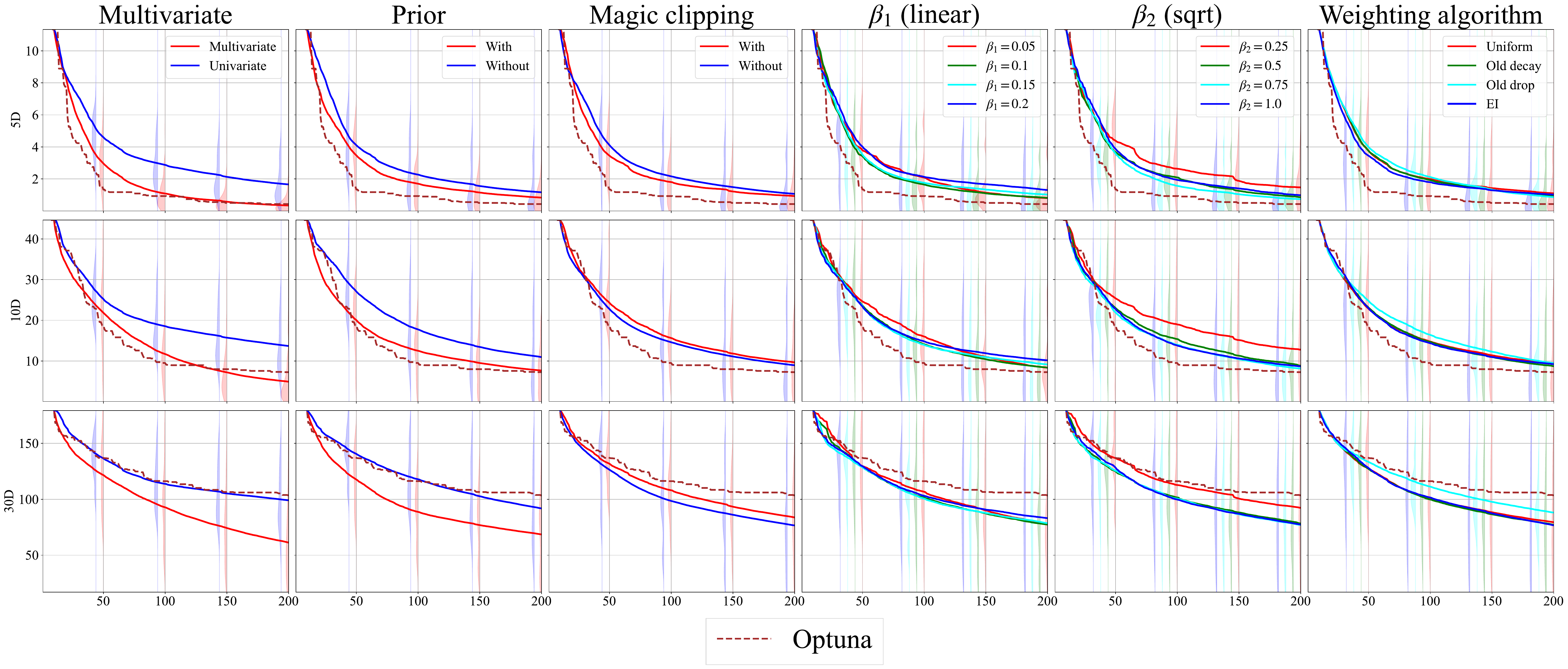}
    }
    \vspace{-2mm}
    \caption{
      The ablation study on benchmark functions.
      The $x$-axis is the number of evaluations and the $y$-axis is the cumulative minimum objective.
      The solid lines in each figure show the mean of the cumulative minimum objective over all control parameter configurations.
      The transparent shades represent the distributions of the cumulative minimum objective at $\{50,100,150,200\}$ evaluations.
      The performance of Optuna v4.0.0 (brown dotted lines) is provided as a baseline.
    }
    \label{appx:experiments:fig:ablation-bench02}
  \end{center}
\end{figure}

\begin{figure}[p]
  \vspace{-10mm}
  \begin{center}
    \subfloat[Styblinski with 5D (\textbf{Top row}), 10D (\textbf{Middle row}), and 30D (\textbf{Bottom row})]{
      \includegraphics[width=0.95\textwidth]{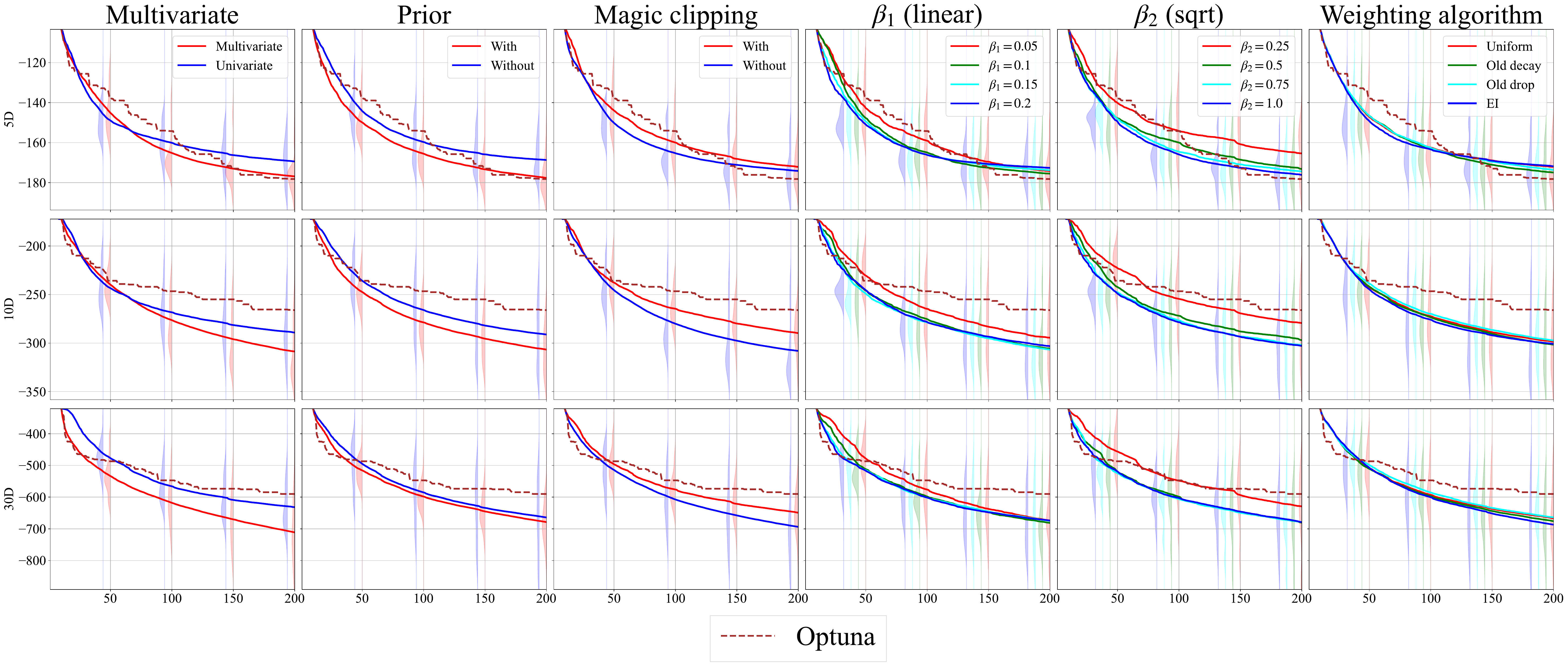}
    }\\
    \vspace{-3mm}
    \subfloat[Weighted sphere with 5D (\textbf{Top row}), 10D (\textbf{Middle row}), and 30D (\textbf{Bottom row})]{
      \includegraphics[width=0.95\textwidth]{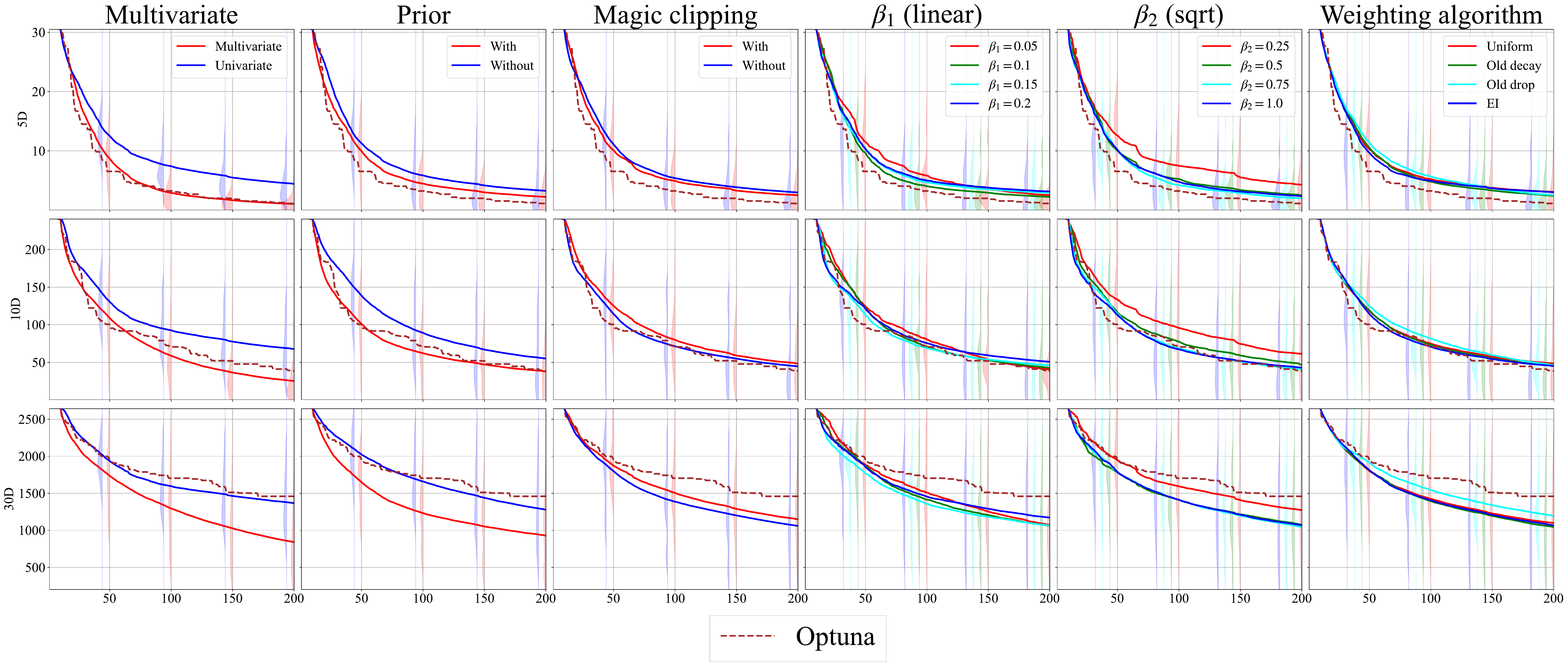}
    }\\
    \vspace{-3mm}
    \subfloat[Xin-She-Yang with 5D (\textbf{Top row}), 10D (\textbf{Middle row}), and 30D (\textbf{Bottom row})]{
      \includegraphics[width=0.95\textwidth]{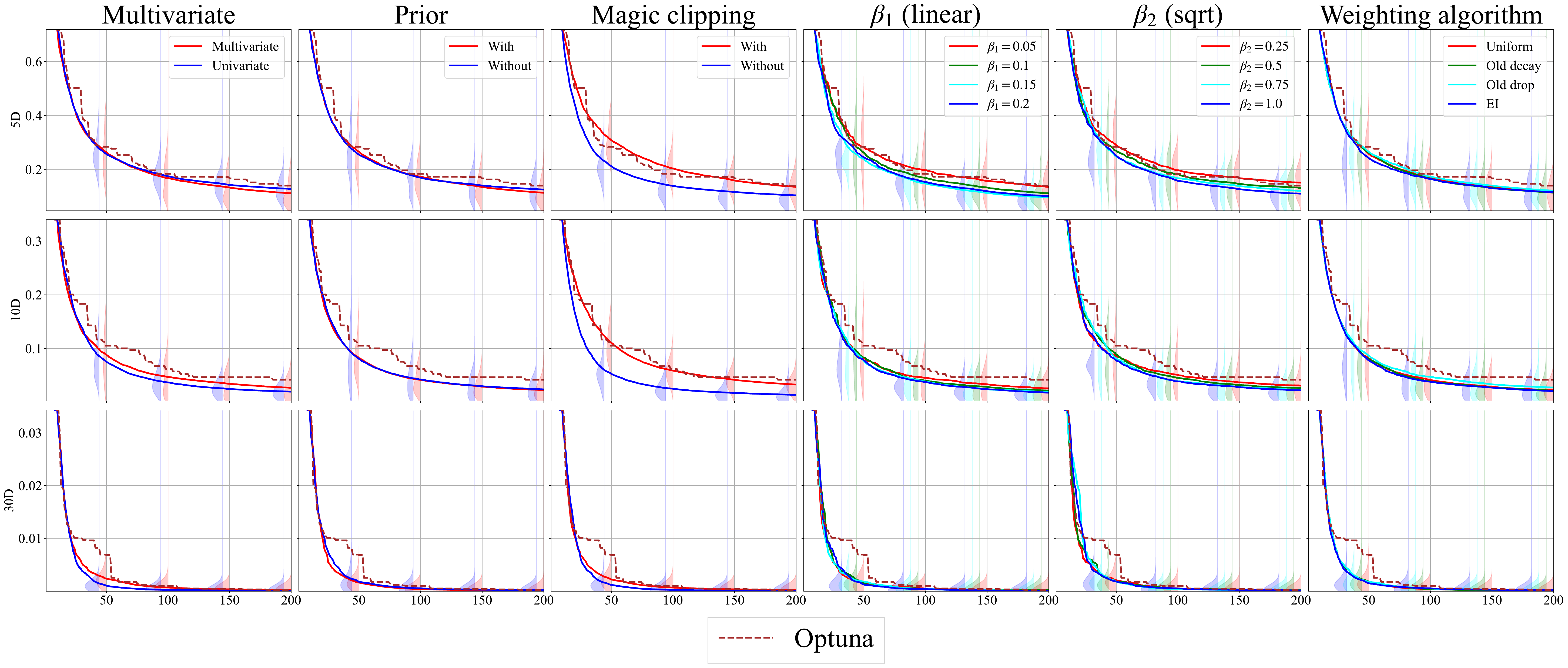}
    }
    \vspace{-2mm}
    \caption{
      The ablation study on benchmark functions.
      The $x$-axis is the number of evaluations and the $y$-axis is the cumulative minimum objective.
      The solid lines in each figure show the mean of the cumulative minimum
      objective over all control parameter configurations.
      The transparent shades represent the distributions of the cumulative minimum objective at $\{50,100,150,200\}$ evaluations.
      The performance of Optuna v4.0.0 (brown dotted lines) is provided as a baseline.
    }
    \label{appx:experiments:fig:ablation-bench03}
  \end{center}
\end{figure}

\begin{figure}[p]
  \centering
  \includegraphics[width=0.98\textwidth]{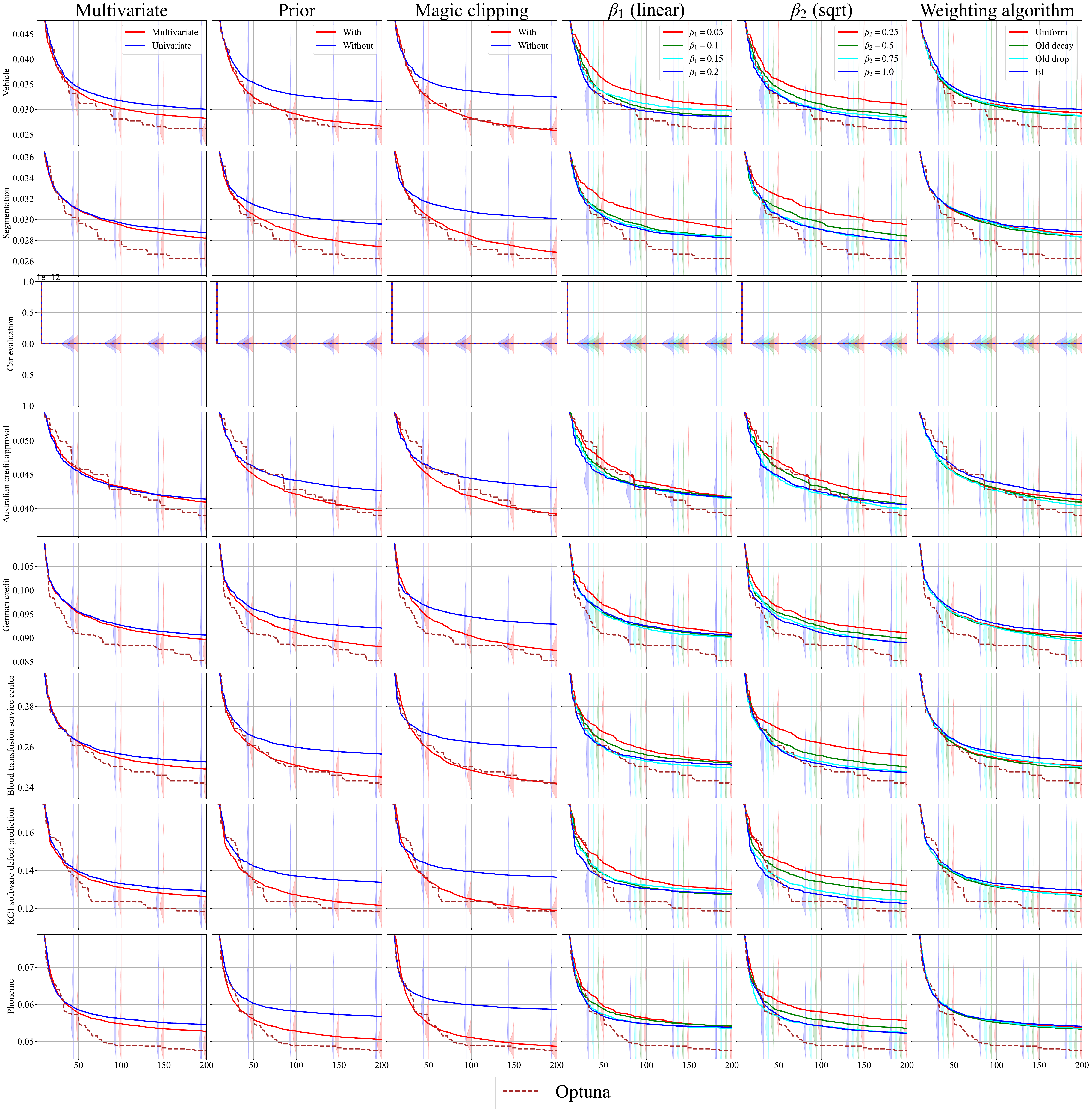}
  \caption{
    The ablation study on HPOBench ($8$ tasks).
    The $x$-axis is the number of evaluations and the $y$-axis is the cumulative minimum objective.
    The solid lines in each figure show the mean of the cumulative minimum objective over all control parameter configurations.
    The transparent shades represent the distributions of the cumulative minimum objective at $\{50,100,150,200\}$ evaluations.
    The performance of Optuna v4.0.0 (brown dotted lines) is provided as a baseline.
  }
  \label{appx:experiments:fig:ablation-hpobench}
\end{figure}

\begin{figure}[p]
  \begin{center}
    \subfloat[HPOlib]{
      \includegraphics[width=0.98\textwidth]{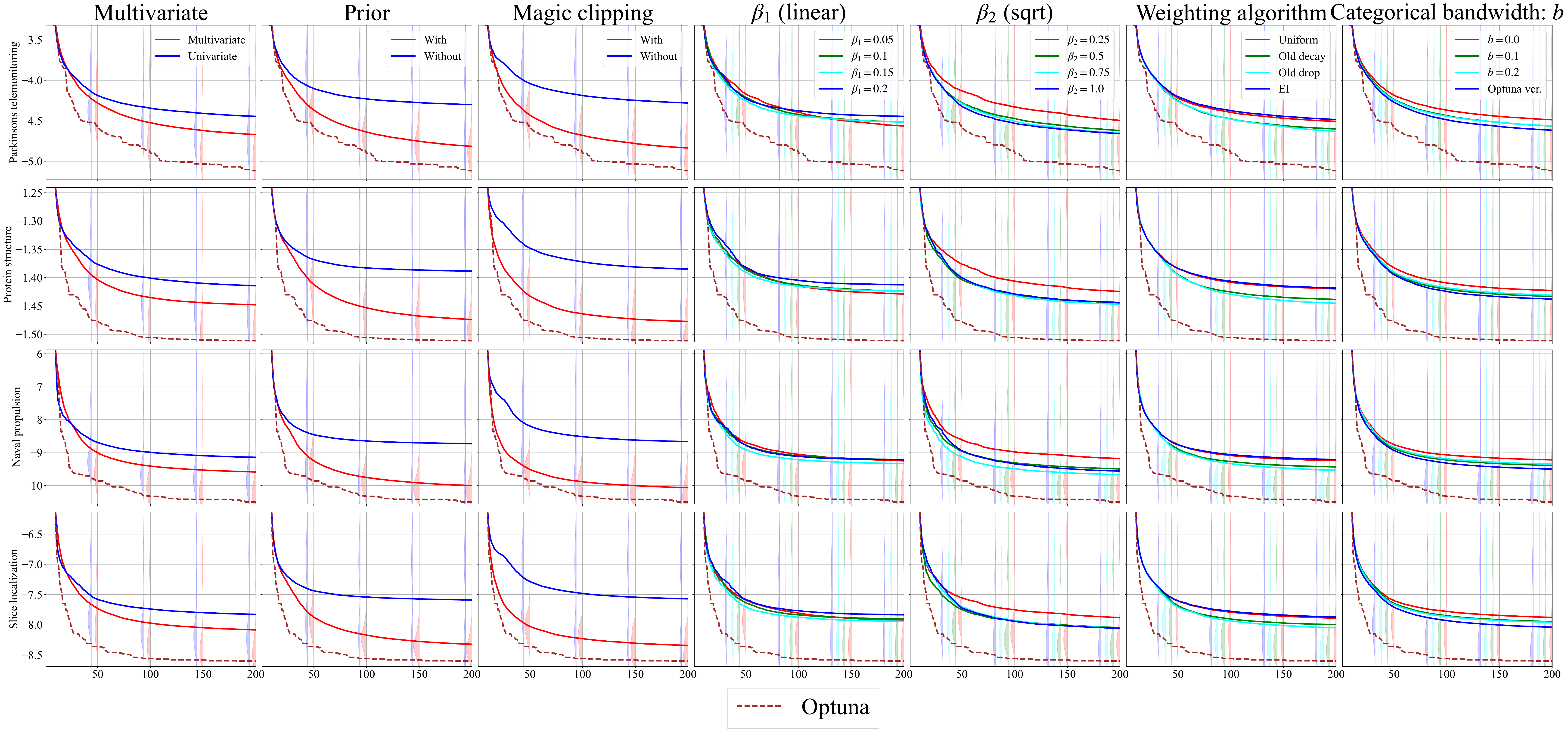}
    }\\
    \vspace{-3mm}
    \subfloat[JAHS-Bench-201]{
      \includegraphics[width=0.98\textwidth]{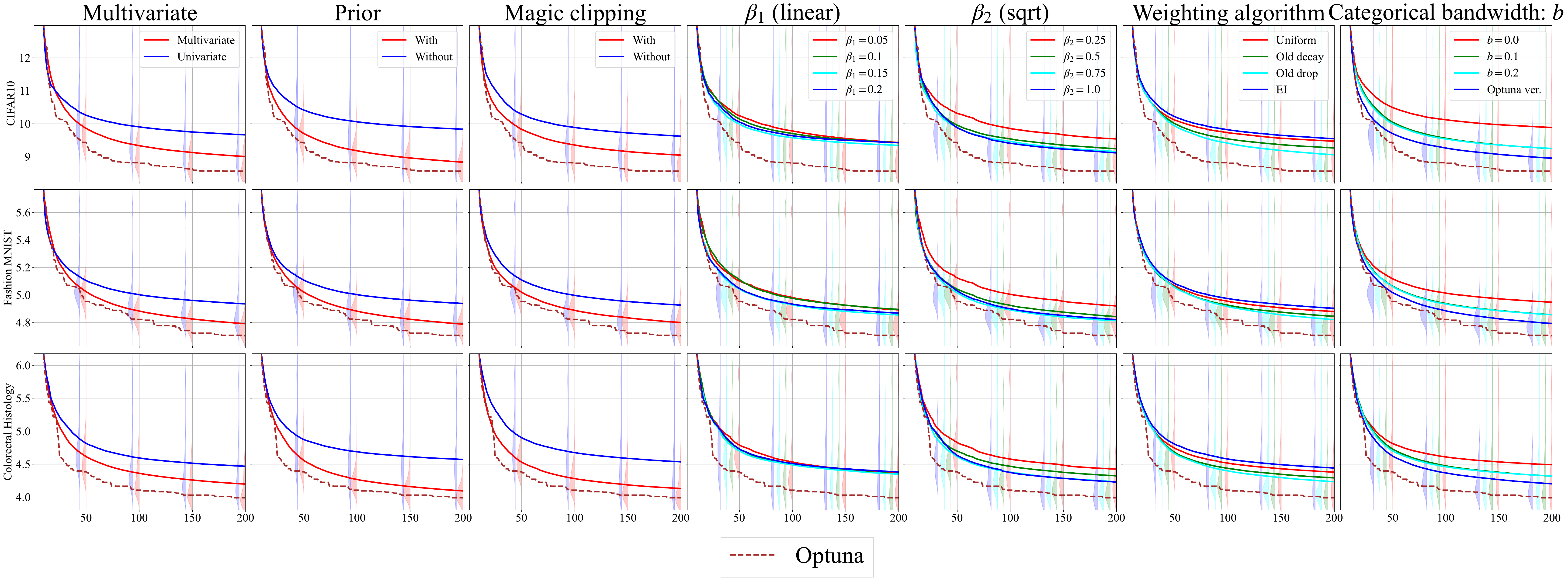}
    }\\
    \vspace{-2mm}
    \caption{
      The ablation study on HPOlib ($4$ tasks) and JAHS-Bench-201 ($3$ tasks), which have categorical parameters.
      The $x$-axis is the number of evaluations and the $y$-axis is the cumulative minimum objective.
      The solid lines in each figure show the mean of the cumulative minimum objective over all control parameter configurations.
      Note that the objective of HPOlib is the $\log$ of validation mean squared error.
      The transparent shades represent the distributions of the cumulative minimum objective at $\{50,100,150,200\}$ evaluations.
      The performance of Optuna v4.0.0 (brown dotted lines) is provided as a baseline.
    }
    \label{appx:experiments:fig:ablation-with-categorical}
  \end{center}
\end{figure}

\section{Additional Results}
\label{appx:experiments:section:additional-results}
This section provides additional results.
The visualization of figures performs the following operations (see the notations invented in Section~\ref{main:experiments:section:ablation-setup}):
\begin{enumerate}
  \vspace{-1mm}
  \item Fix a control parameter (e.g., \texttt{multivariate=True}),
  \vspace{-1mm}
  \item Gather all the cumulative minimum performance curves $\{\{\zeta_{k,n}^{(m)}\}_{k=1}^K\}_{n=1}^{N}$ where $N = 200$ was used in the experiments,
  \vspace{-1mm}
  \item Plot the mean value $1/K \sum_{k=1}^{K}\zeta_{k,n}^{(m)}$ over the gathered results at each ``$n$ evaluations'' ($n = 1, \dots, N$) as a solid line,
  \vspace{-1mm}
  \item Plot vertically the (violin-plot-like) distribution of the gathered results $\{\zeta_{k,n}^{(m)}\}_{k=1}^K$ at $n \in \{50,100,150,200\}$ evaluations as a transparent shadow, and
  \vspace{-1mm}
  \item Repeat Operations 1.--4. for all settings.
  \vspace{-1mm}
\end{enumerate}
The violin-plot-like distributions are used in the visualization instead of the standard error to conpensate for the lack of distributional information.
Each result includes optimizations by Optuna v4.0.0 as a baseline.
The following results are presented:
\begin{enumerate}
  \vspace{-1mm}
  \item \textbf{Ablation Study}: Figures~\ref{appx:experiments:fig:ablation-bench00}--\ref{appx:experiments:fig:ablation-bench03} present the results on the benchmark functions and Figures~\ref{appx:experiments:fig:ablation-hpobench},\ref{appx:experiments:fig:ablation-with-categorical} present the results on the HPO benchmarks.
  \vspace{-1mm}
  \item \textbf{Analysis of Bandwidth Selection}: Figures~\ref{appx:experiments:fig:bandwidth-ackley}--\ref{appx:experiments:fig:bandwidth-xin-she-yang} present the results on the benchmark functions and Figures~\ref{appx:experiments:fig:bandwidth-hpobench0}--\ref{appx:experiments:fig:bandwidth-jahs} present the results on the HPO benchmarks.
  \vspace{-1mm}
  \item \textbf{Comparison with Baseline Methods}: Figures~\ref{appx:experiments:fig:comparison-bench0}--\ref{appx:experiments:fig:comparison-bench3} present the results on the benchmark functions and Figure~\ref{appx:experiments:fig:comparison-hpobench} presents the results on the HPO benchmarks.
  \vspace{-1mm}
\end{enumerate}

\begin{figure}[p]
  \vspace{0mm}
  \centering
  \includegraphics[width=0.98\textwidth]{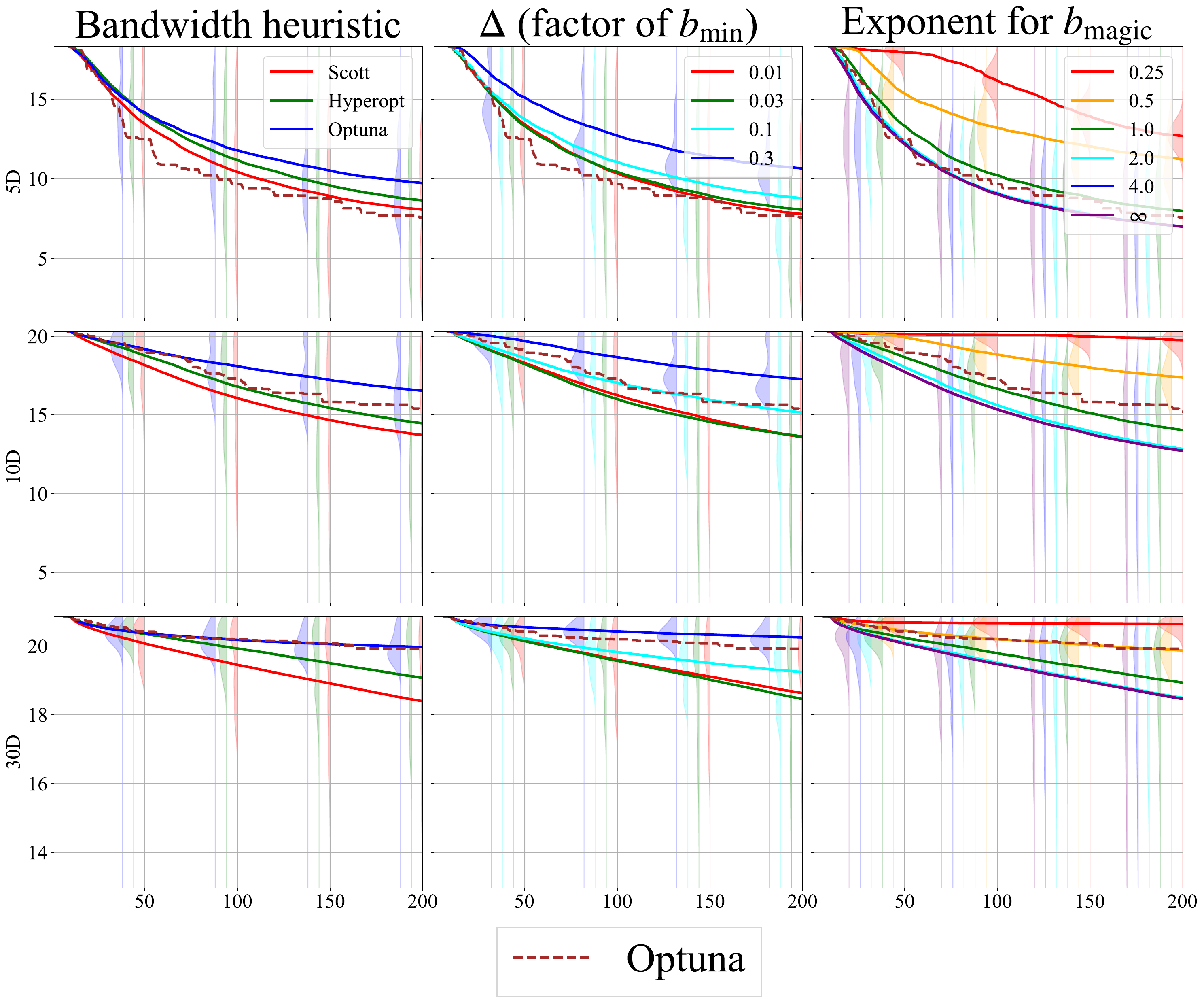}
  \vspace{-3mm}
  \caption{
    The ablation study of bandwidth related algorithms on the Ackley function.
    The $x$-axis is the number of evaluations and the $y$-axis is the cumulative minimum objective.
    The solid lines in each figure show the mean of the cumulative minimum objective over all control parameter configurations.
    The transparent shades represent the distributions of the cumulative minimum objective at $\{50,100,150,200\}$ evaluations.
    The performance of Optuna v4.0.0 (brown dotted lines) is provided as a baseline.
  }
  \label{appx:experiments:fig:bandwidth-ackley}
\end{figure}

\begin{figure}[p]
  \vspace{0mm}
  \centering
  \includegraphics[width=0.98\textwidth]{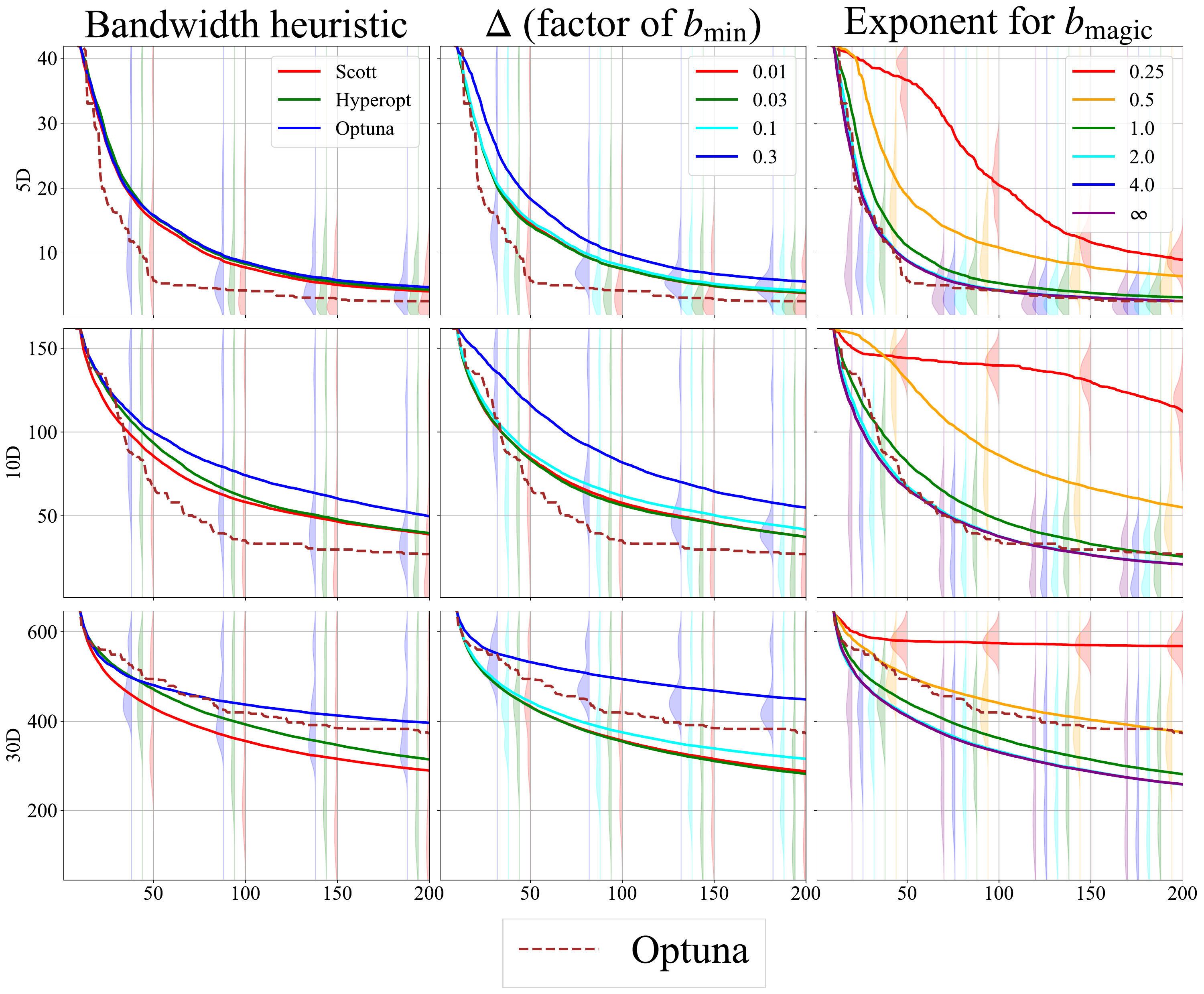}
  \vspace{-3mm}
  \caption{
    The ablation study of bandwidth related algorithms on the Griewank function.
    The $x$-axis is the number of evaluations and the $y$-axis is the cumulative minimum objective.
    The solid lines in each figure show the mean of the cumulative minimum objective over all control parameter configurations.
    The transparent shades represent the distributions of the cumulative minimum objective at $\{50,100,150,200\}$ evaluations.
    The performance of Optuna v4.0.0 (brown dotted lines) is provided as a baseline.
  }
  \label{appx:experiments:fig:bandwidth-griewank}
\end{figure}

\begin{figure}[p]
  \vspace{0mm}
  \centering
  \includegraphics[width=0.98\textwidth]{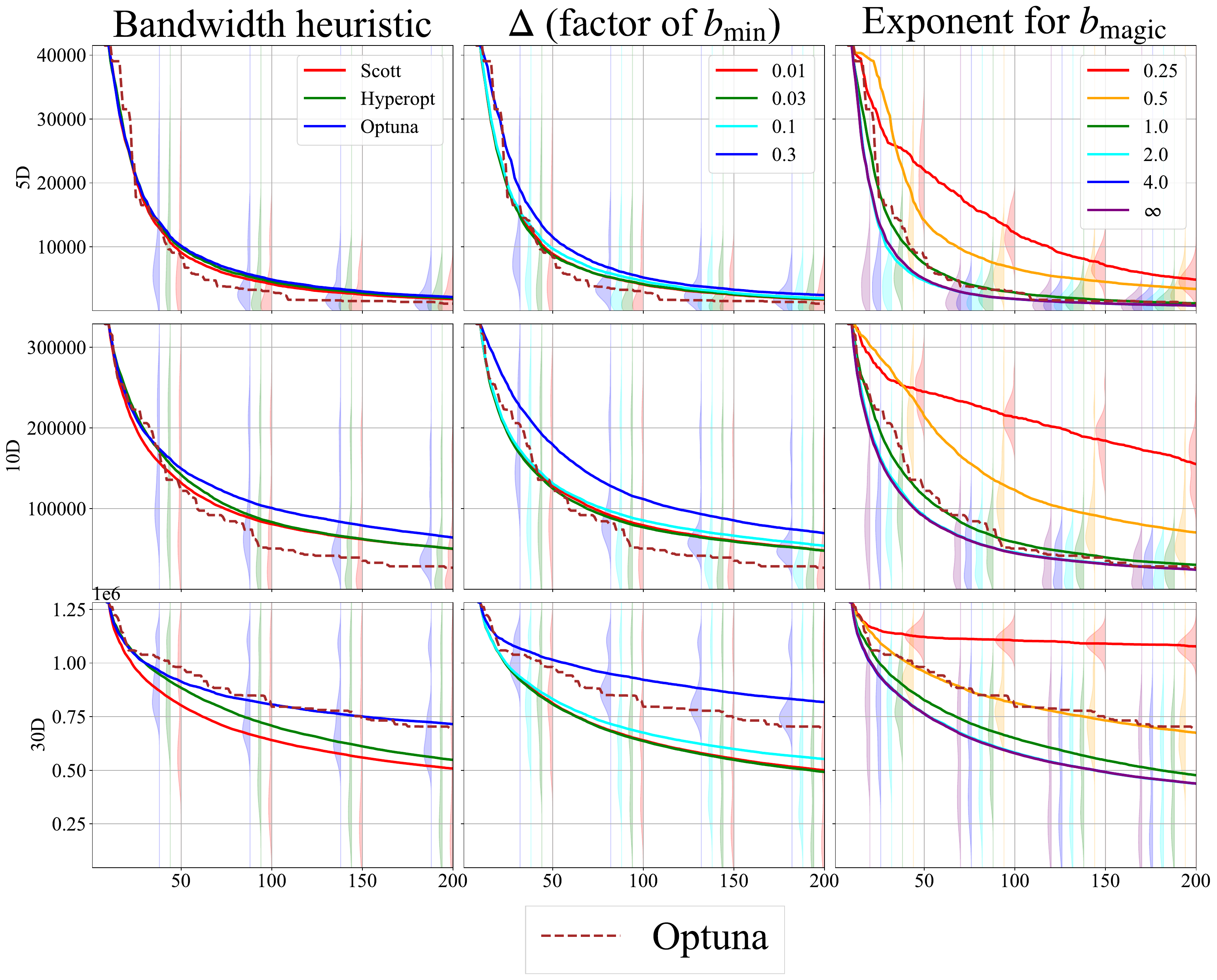}
  \vspace{-3mm}
  \caption{
    The ablation study of bandwidth related algorithms on the K-Tablet function.
    The $x$-axis is the number of evaluations and the $y$-axis is the cumulative minimum objective.
    The solid lines in each figure show the mean of the cumulative minimum objective over all control parameter configurations.
    The transparent shades represent the distributions of the cumulative minimum objective at $\{50,100,150,200\}$ evaluations.
    The performance of Optuna v4.0.0 (brown dotted lines) is provided as a baseline.
  }
  \label{appx:experiments:fig:bandwidth-ktablet}
\end{figure}

\begin{figure}[p]
  \vspace{0mm}
  \centering
  \includegraphics[width=0.98\textwidth]{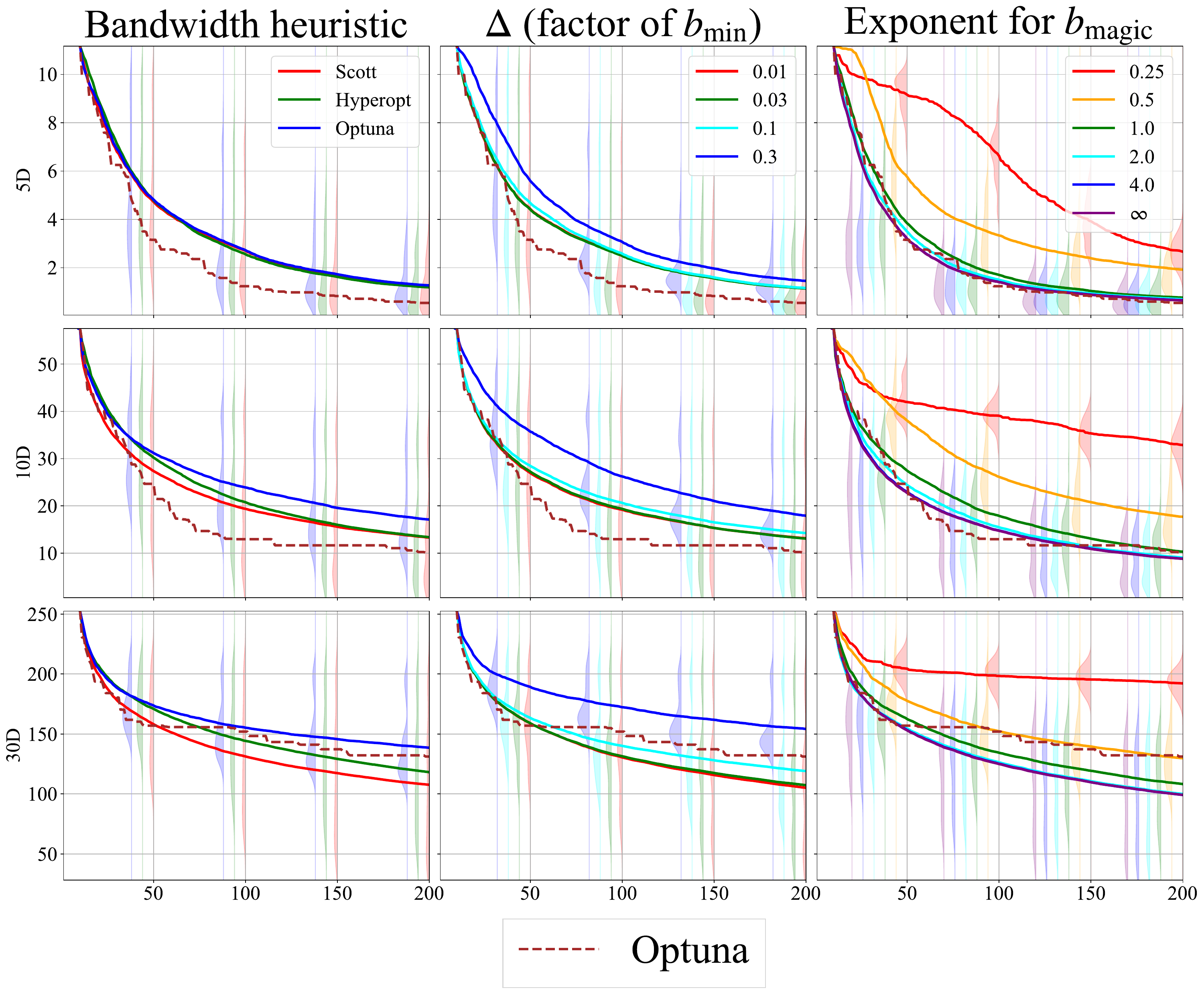}
  \vspace{-3mm}
  \caption{
    The ablation study of bandwidth related algorithms on the Levy function.
    The $x$-axis is the number of evaluations and the $y$-axis is the cumulative minimum objective.
    The solid lines in each figure show the mean of the cumulative minimum objective over all control parameter configurations.
    The transparent shades represent the distributions of the cumulative minimum objective at $\{50,100,150,200\}$ evaluations.
    The performance of Optuna v4.0.0 (brown dotted lines) is provided as a baseline.
  }
  \label{appx:experiments:fig:bandwidth-levy}
\end{figure}

\begin{figure}[p]
  \vspace{0mm}
  \centering
  \includegraphics[width=0.98\textwidth]{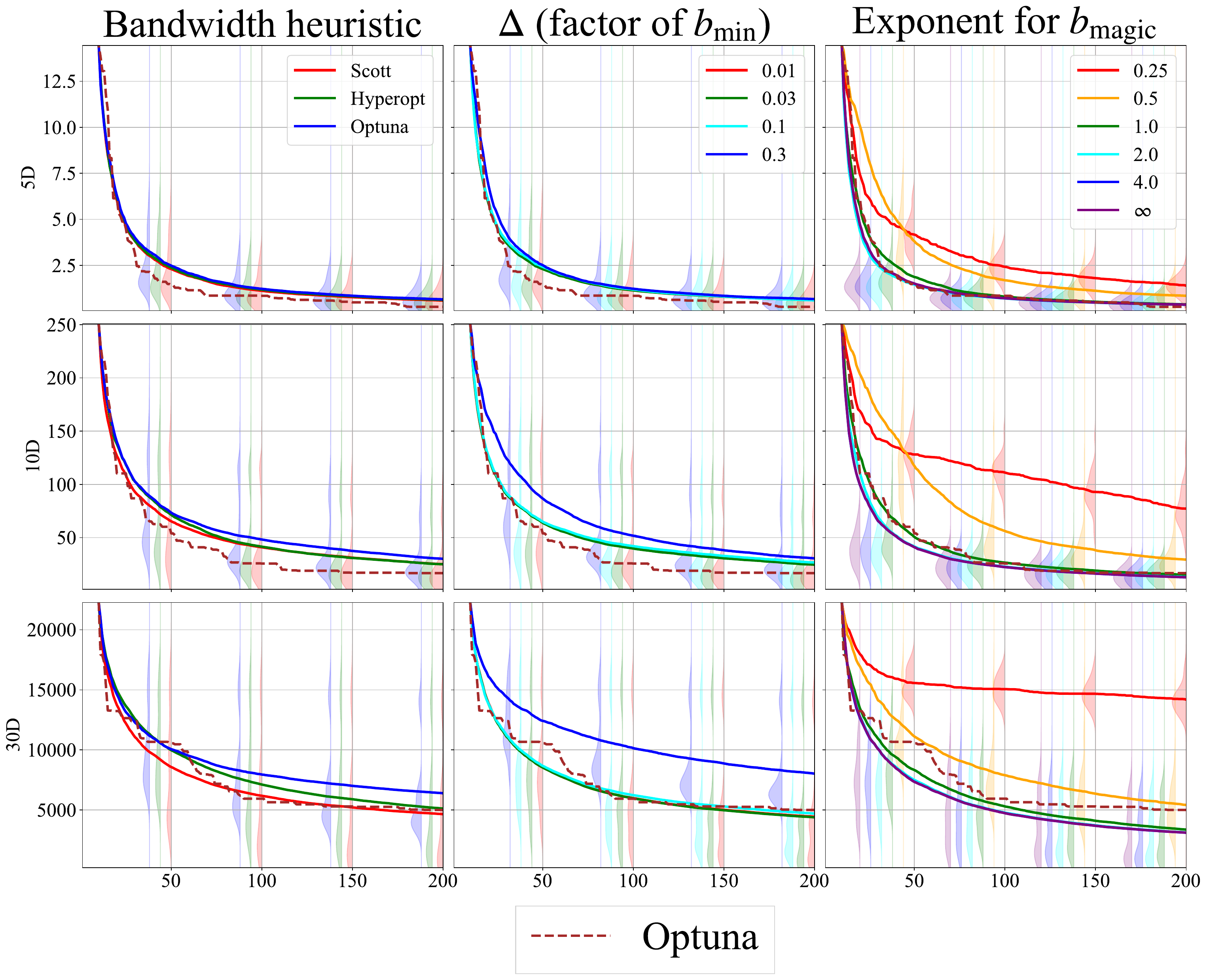}
  \vspace{-3mm}
  \caption{
    The ablation study of bandwidth related algorithms on the Perm function.
    The $x$-axis is the number of evaluations and the $y$-axis is the cumulative minimum objective.
    The solid lines in each figure show the mean of the cumulative minimum objective over all control parameter configurations.
    The transparent shades represent the distributions of the cumulative minimum objective at $\{50,100,150,200\}$ evaluations.
    The performance of Optuna v4.0.0 (brown dotted lines) is provided as a baseline.
  }
  \label{appx:experiments:fig:bandwidth-perm}
\end{figure}

\begin{figure}[p]
  \vspace{0mm}
  \centering
  \includegraphics[width=0.98\textwidth]{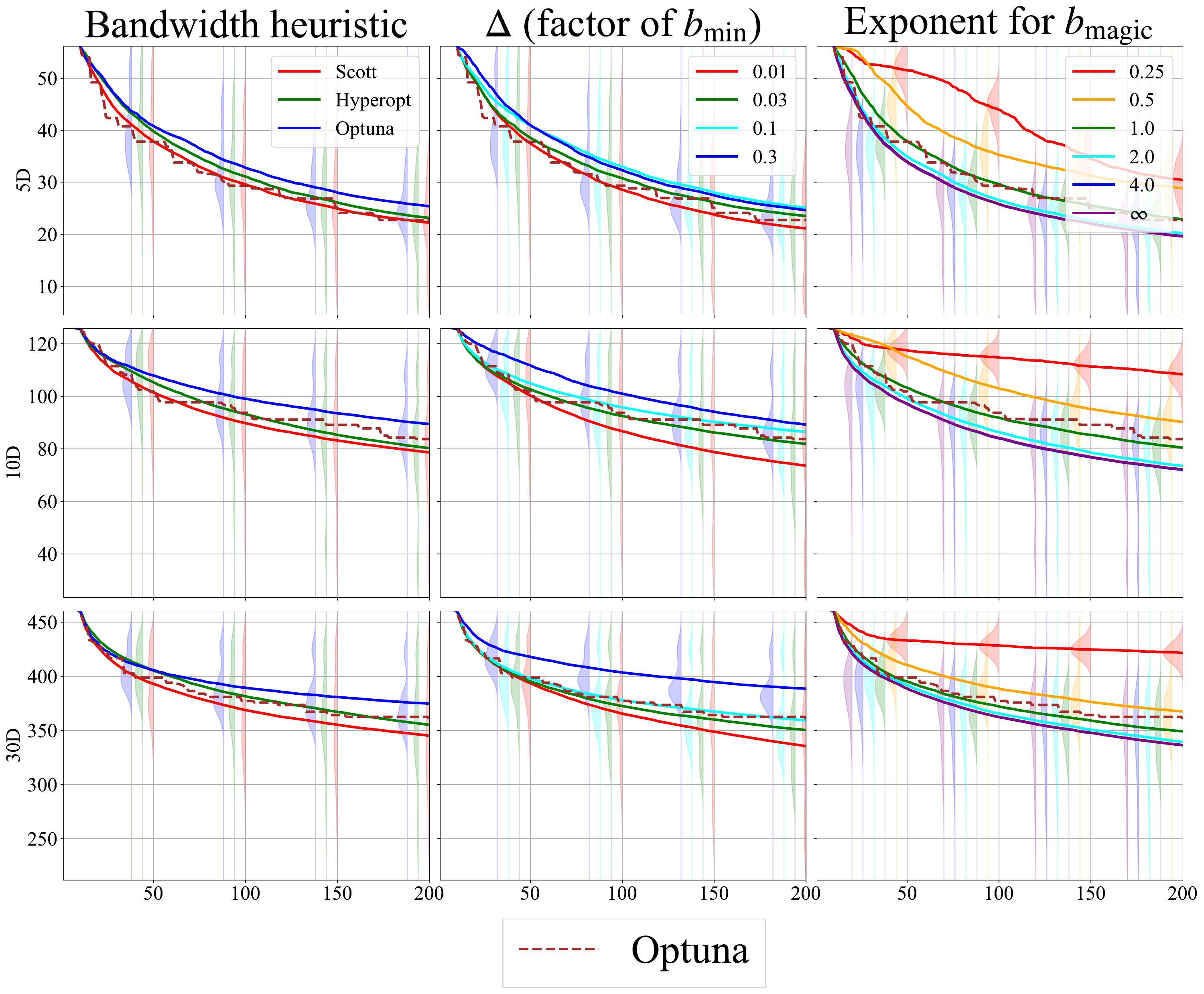}
  \vspace{-3mm}
  \caption{
    The ablation study of bandwidth related algorithms on the Rastrigin function.
    The $x$-axis is the number of evaluations and the $y$-axis is the cumulative minimum objective.
    The solid lines in each figure show the mean of the cumulative minimum objective over all control parameter configurations.
    The transparent shades represent the distributions of the cumulative minimum objective at $\{50,100,150,200\}$ evaluations.
    The performance of Optuna v4.0.0 (brown dotted lines) is provided as a baseline.
  }
  \label{appx:experiments:fig:bandwidth-rastrigin}
\end{figure}

\begin{figure}[p]
  \vspace{-10mm}
  \centering
  \includegraphics[width=0.98\textwidth]{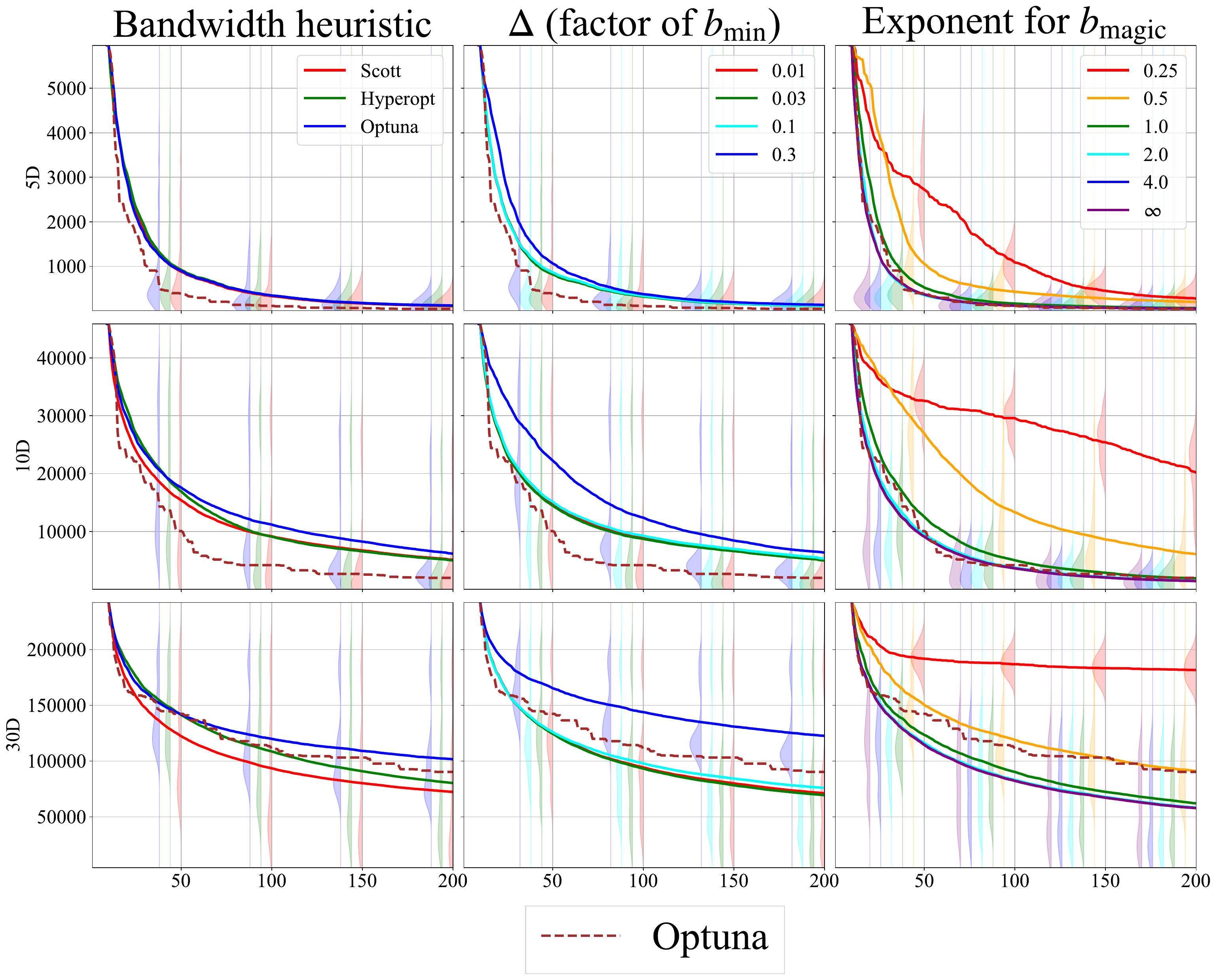}
  \vspace{-3mm}
  \caption{
    The ablation study of bandwidth related algorithms on the Rosenbrock function.
    The $x$-axis is the number of evaluations and the $y$-axis is the cumulative minimum objective.
    The solid lines in each figure show the mean of the cumulative minimum objective over all control parameter configurations.
    The transparent shades represent the distributions of the cumulative minimum objective at $\{50,100,150,200\}$ evaluations.
    The performance of Optuna v4.0.0 (brown dotted lines) is provided as a baseline.
  }
  \label{appx:experiments:fig:bandwidth-rosenbrock}
\end{figure}

\begin{figure}[p]
  \vspace{-10mm}
  \centering
  \includegraphics[width=0.98\textwidth]{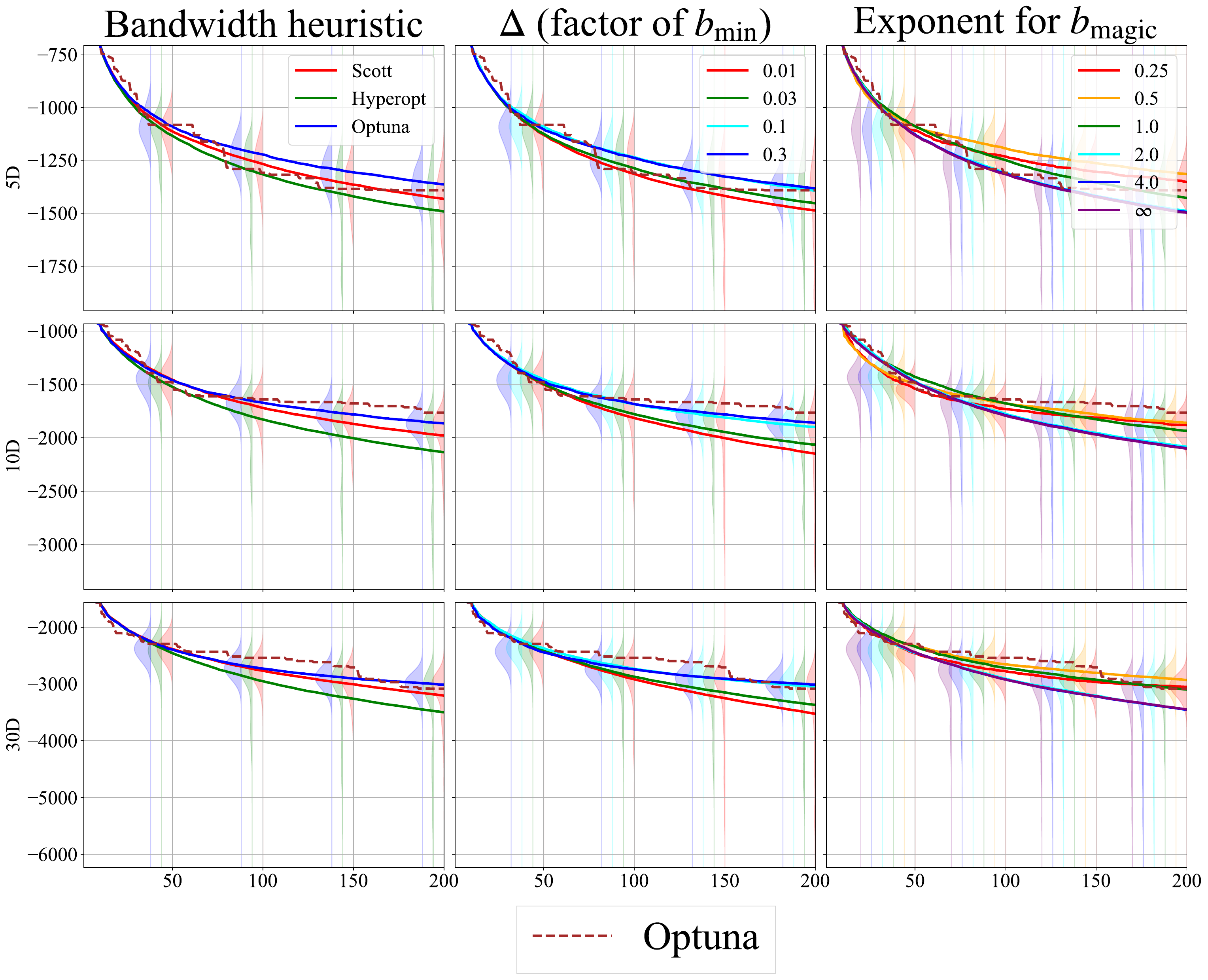}
  \vspace{-3mm}
  \caption{
    The ablation study of bandwidth related algorithms on the Schwefel function.
    The $x$-axis is the number of evaluations and the $y$-axis is the cumulative minimum objective.
    The solid lines in each figure show the mean of the cumulative minimum objective over all control parameter configurations.
    The transparent shades represent the distributions of the cumulative minimum objective at $\{50,100,150,200\}$ evaluations.
    The performance of Optuna v4.0.0 (brown dotted lines) is provided as a baseline.
  }
  \label{appx:experiments:fig:bandwidth-schwefel}
\end{figure}

\begin{figure}[p]
  \vspace{-10mm}
  \centering
  \includegraphics[width=0.98\textwidth]{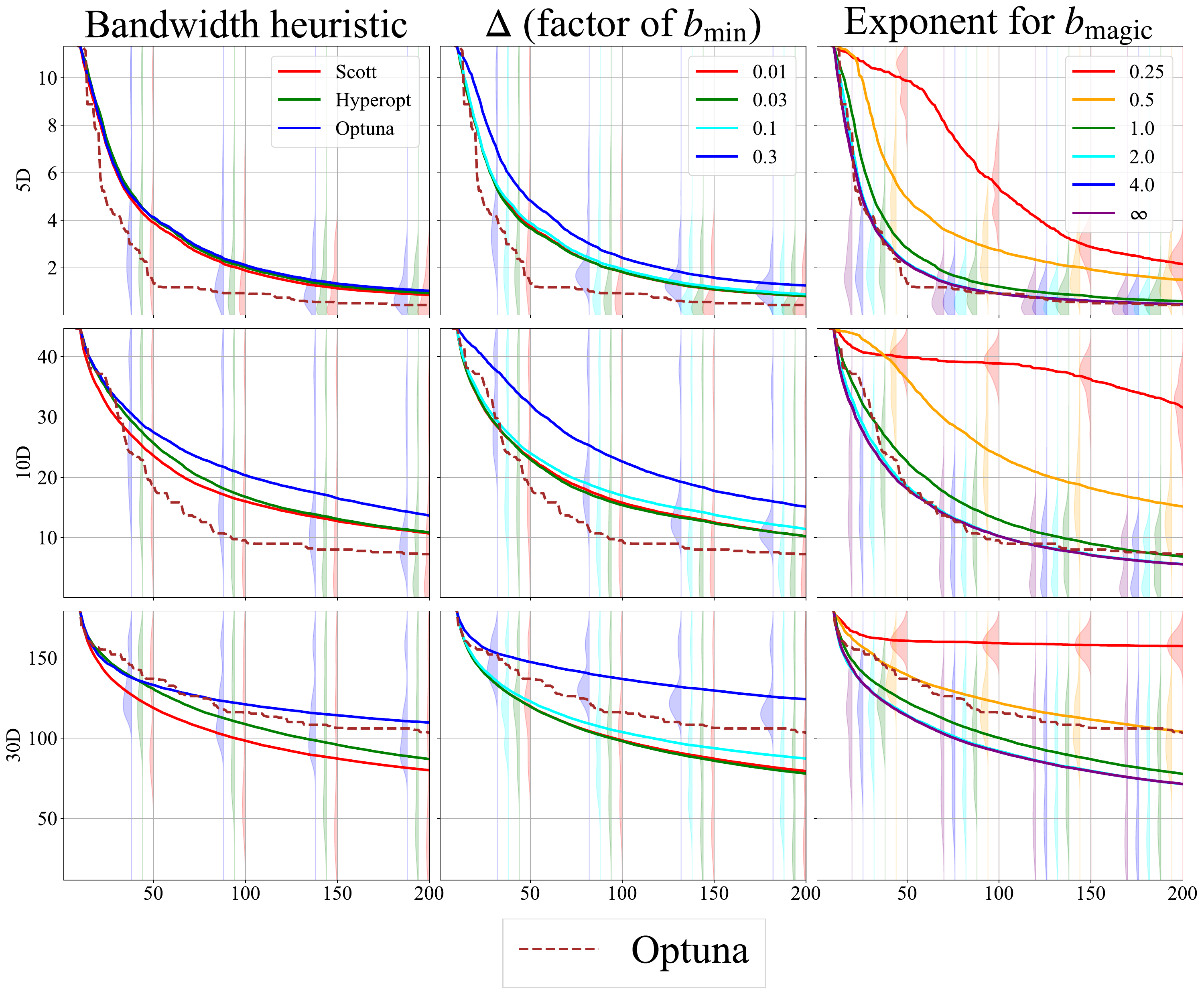}
  \vspace{-3mm}
  \caption{
    The ablation study of bandwidth related algorithms on the Sphere function.
    The $x$-axis is the number of evaluations and the $y$-axis is the cumulative minimum objective.
    The solid lines in each figure show the mean of the cumulative minimum objective over all control parameter configurations.
    The transparent shades represent the distributions of the cumulative minimum objective at $\{50,100,150,200\}$ evaluations.
    The performance of Optuna v4.0.0 (brown dotted lines) is provided as a baseline.
  }
  \label{appx:experiments:fig:bandwidth-sphere}
\end{figure}

\begin{figure}[p]
  \vspace{-10mm}
  \centering
  \includegraphics[width=0.98\textwidth]{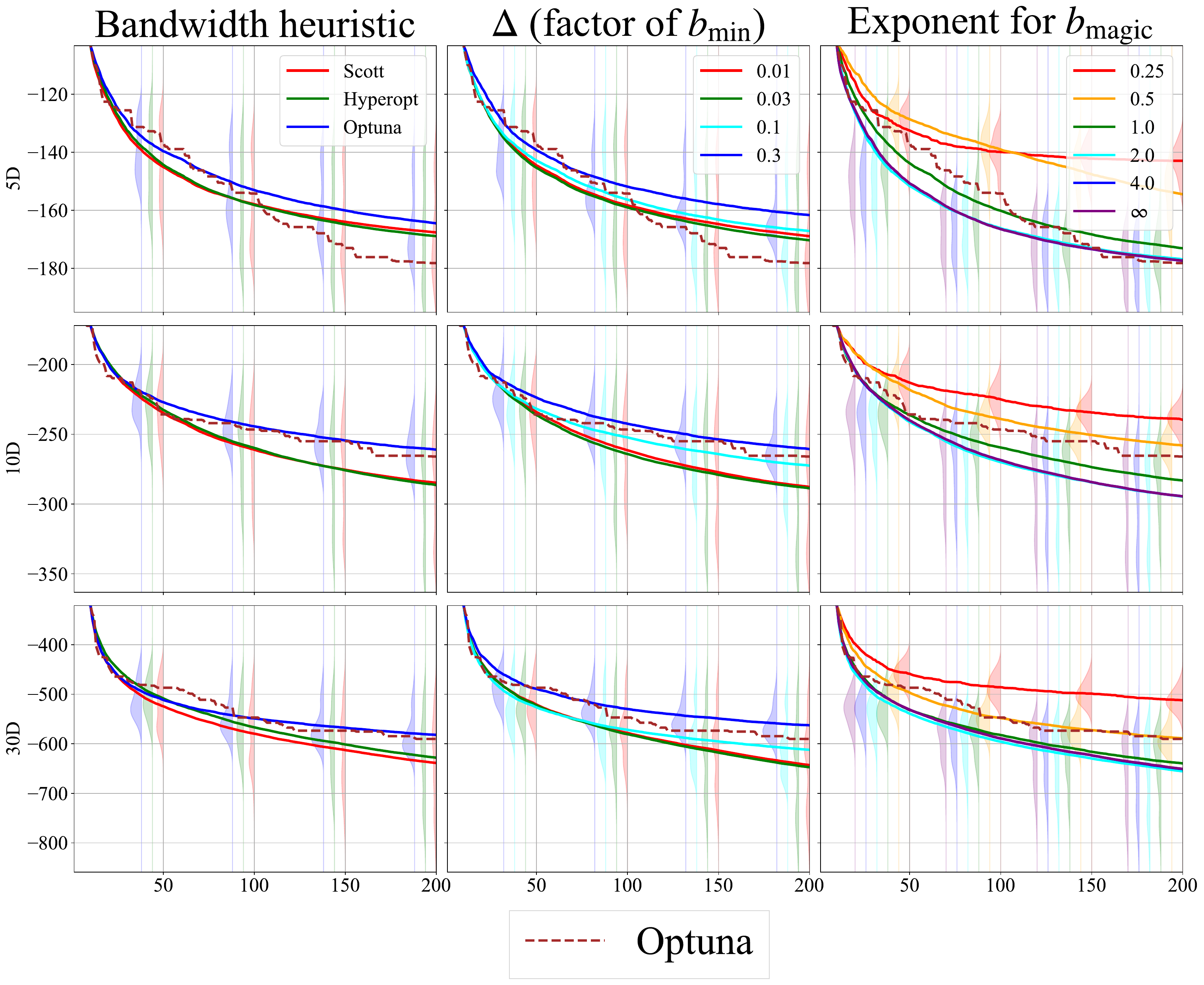}
  \vspace{-3mm}
  \caption{
    The ablation study of bandwidth related algorithms on the Styblinski function.
    The $x$-axis is the number of evaluations and the $y$-axis is the cumulative minimum objective.
    The solid lines in each figure show the mean of the cumulative minimum objective over all control parameter configurations.
    The transparent shades represent the distributions of the cumulative minimum objective at $\{50,100,150,200\}$ evaluations.
    The performance of Optuna v4.0.0 (brown dotted lines) is provided as a baseline.
  }
  \label{appx:experiments:fig:bandwidth-styblinski}
\end{figure}

\begin{figure}[p]
  \vspace{-10mm}
  \centering
  \includegraphics[width=0.98\textwidth]{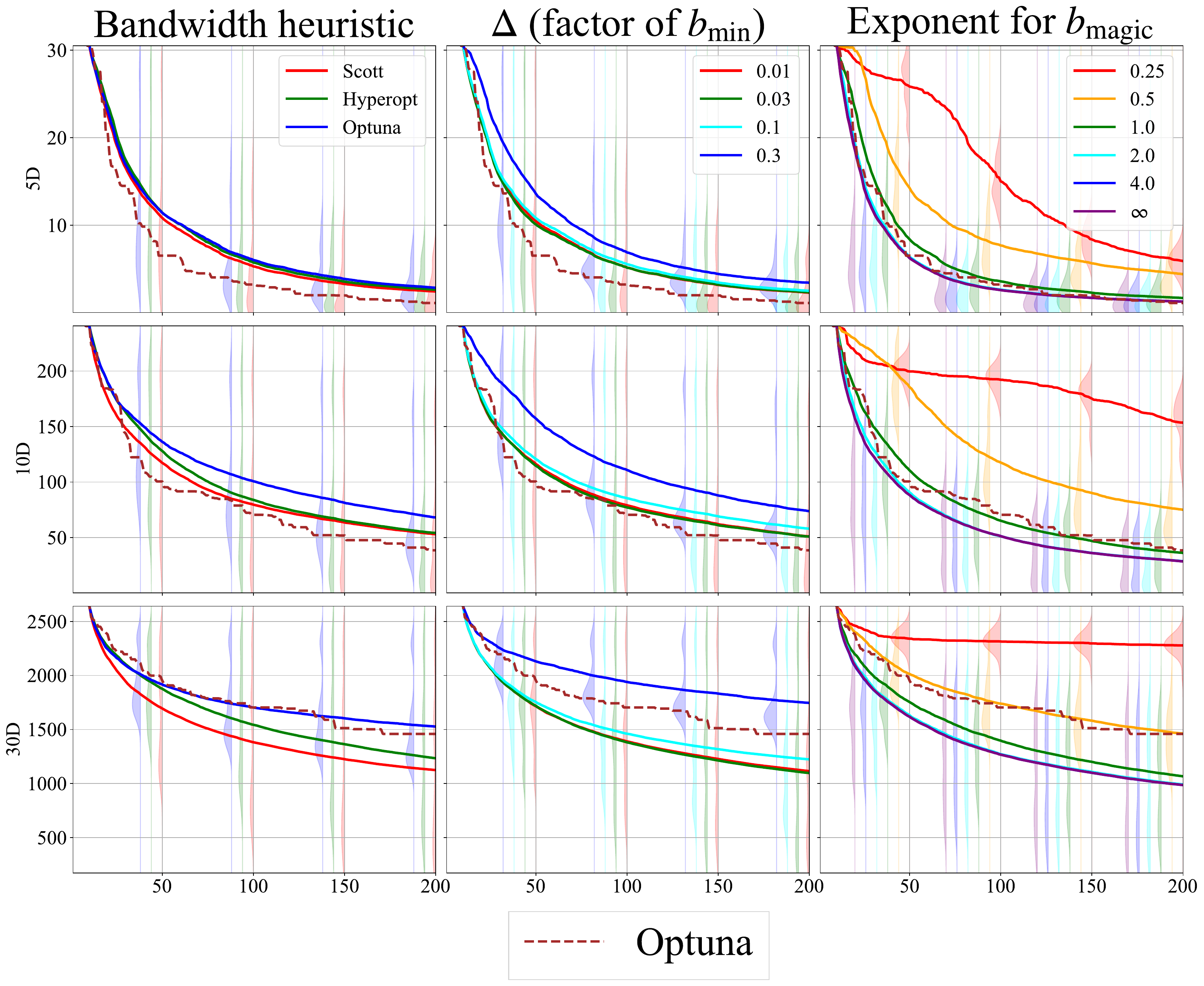}
  \vspace{-3mm}
  \caption{
    The ablation study of bandwidth related algorithms on the weighted sphere function.
    The $x$-axis is the number of evaluations and the $y$-axis is the cumulative minimum objective.
    The solid lines in each figure show the mean of the cumulative minimum objective over all control parameter configurations.
    The transparent shades represent the distributions of the cumulative minimum objective at $\{50,100,150,200\}$ evaluations.
    The performance of Optuna v4.0.0 (brown dotted lines) is provided as a baseline.
  }
  \label{appx:experiments:fig:bandwidth-weighted-sphere}
\end{figure}

\begin{figure}[p]
  \vspace{-10mm}
  \centering
  \includegraphics[width=0.98\textwidth]{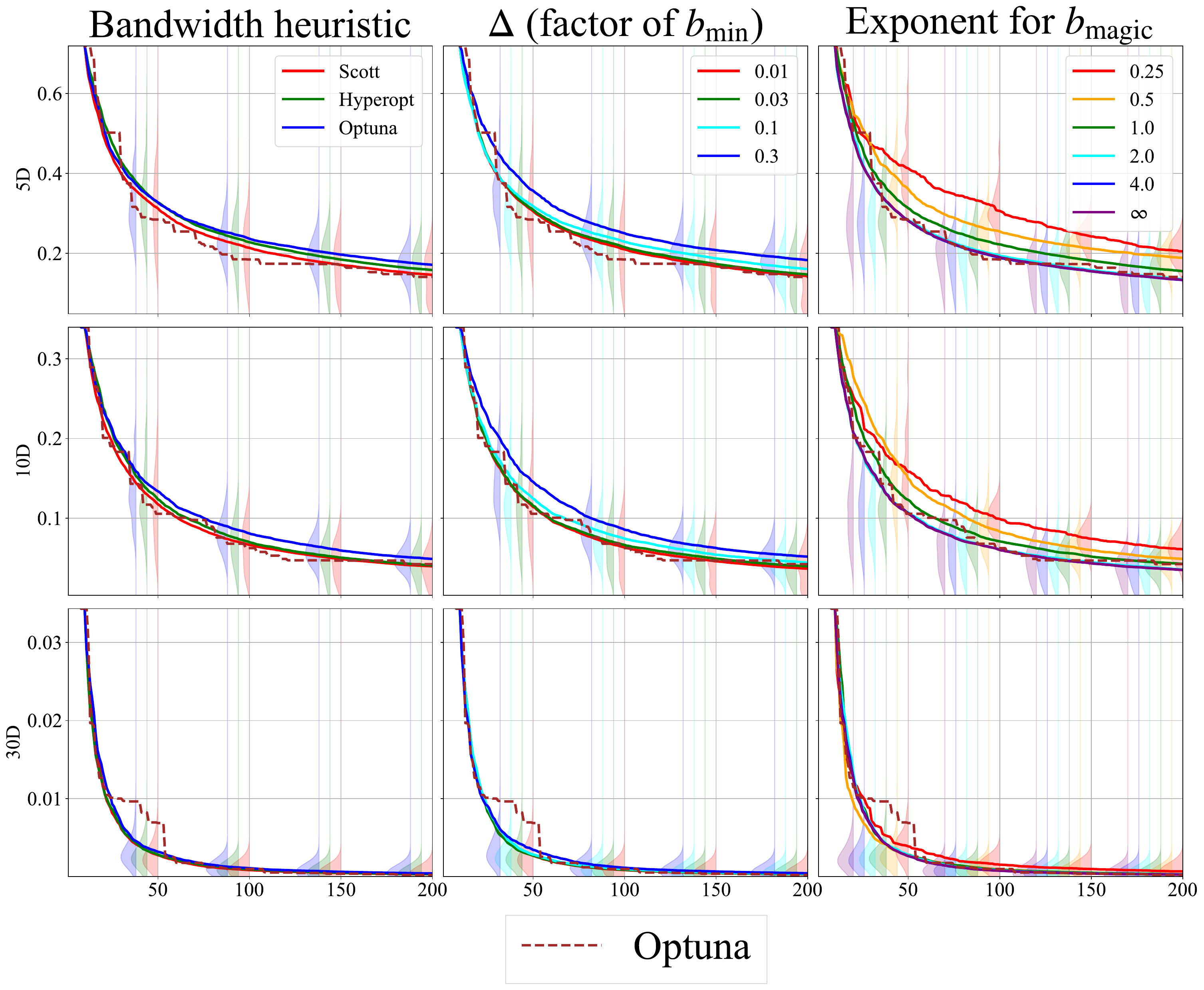}
  \vspace{-3mm}
  \caption{
    The ablation study of bandwidth related algorithms on the Xin-She-Yang function.
    The $x$-axis is the number of evaluations and the $y$-axis is the cumulative minimum objective.
    The solid lines in each figure show the mean of the cumulative minimum objective over all control parameter configurations.
    The transparent shades represent the distributions of the cumulative minimum objective at $\{50,100,150,200\}$ evaluations.
    The performance of Optuna v4.0.0 (brown dotted lines) is provided as a baseline.
  }
  \label{appx:experiments:fig:bandwidth-xin-she-yang}
\end{figure}

\begin{figure}[p]
  \vspace{-10mm}
  \centering
  \includegraphics[width=0.98\textwidth]{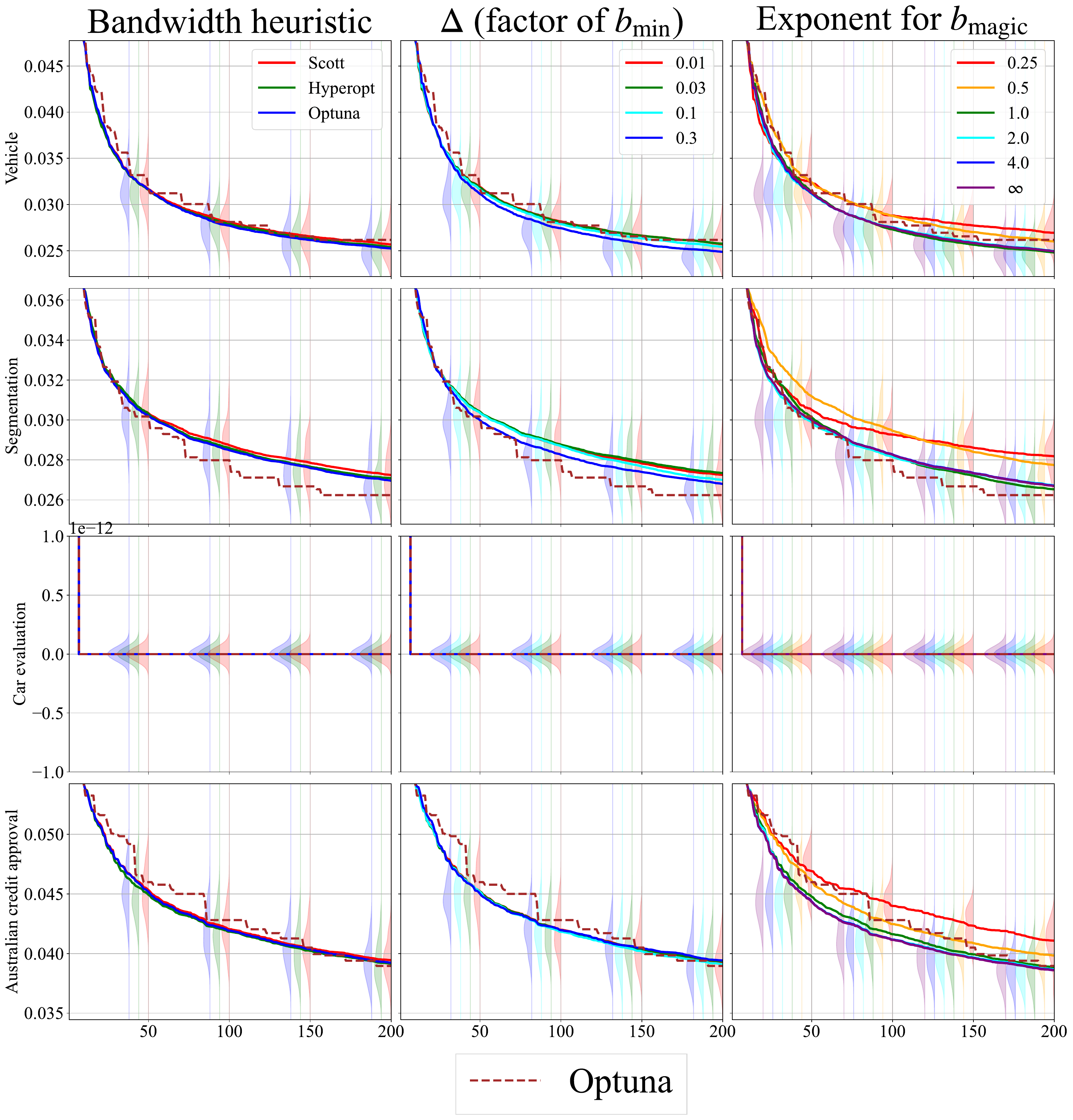}
  \vspace{-3mm}
  \caption{
    The ablation study of bandwidth algorithms on HPOBench
    (\texttt{Vehicle}, \texttt{Segmentation}, \texttt{Car evaluation}, \texttt{Australian credit approval}).
    The $x$-axis is the number of evaluations and the $y$-axis is the cumulative minimum objective.
    The solid lines in each figure show the mean of the cumulative minimum objective over all control parameter configurations.
    The transparent shades represent the distributions of the cumulative minimum objective at $\{50,100,150,200\}$ evaluations.
    The performance of Optuna v4.0.0 (brown dotted lines) is provided as a baseline.
  }
  \label{appx:experiments:fig:bandwidth-hpobench0}
\end{figure}

\begin{figure}[p]
  \vspace{-10mm}
  \centering
  \includegraphics[width=0.98\textwidth]{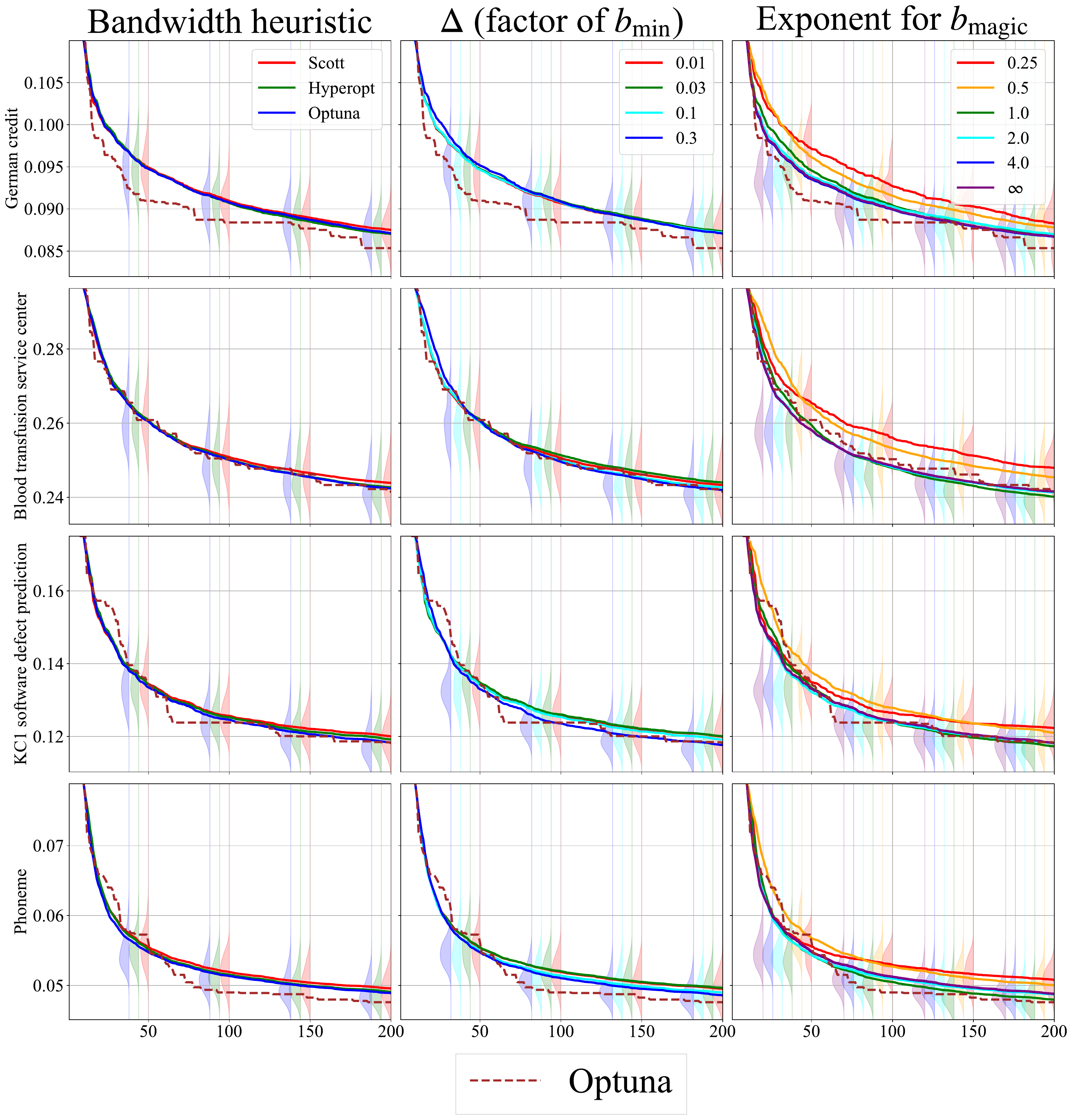}
  \vspace{-3mm}
  \caption{
    The ablation study of bandwidth algorithms on HPOBench
    (\texttt{German credit}, \texttt{Blood transfusion service center}, \texttt{KC1 software defect prediction}, \texttt{Phoneme}).
    The $x$-axis is the number of evaluations and the $y$-axis is the cumulative minimum objective.
    The solid lines in each figure show the mean of the cumulative minimum objective over all control parameter configurations.
    The transparent shades represent the distributions of the cumulative minimum objective at $\{50,100,150,200\}$ evaluations.
    The performance of Optuna v4.0.0 (brown dotted lines) is provided as a baseline.
  }
  \label{appx:experiments:fig:bandwidth-hpobench1}
\end{figure}

\begin{figure}[p]
  \vspace{-10mm}
  \centering
  \includegraphics[width=0.98\textwidth]{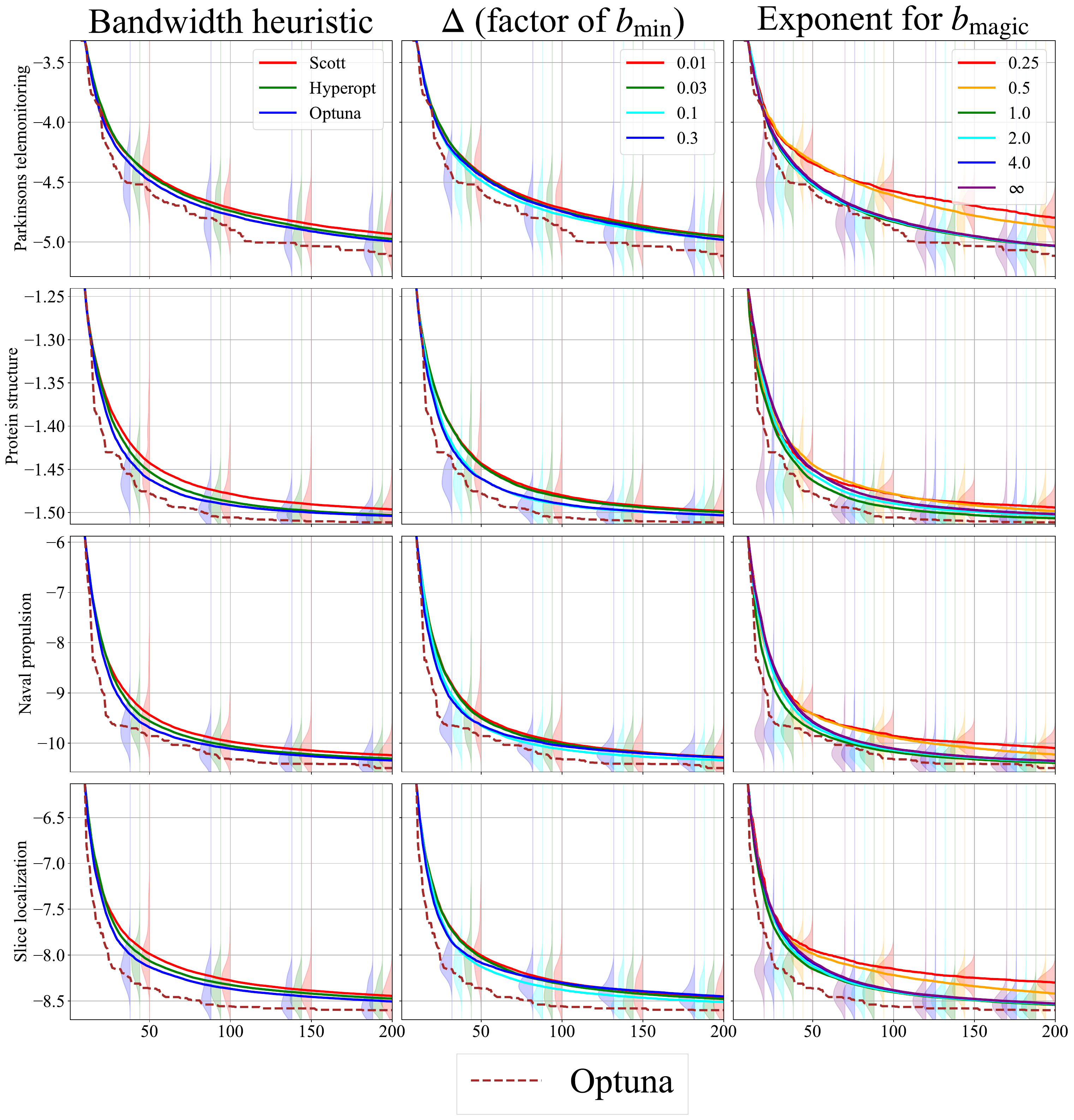}
  \vspace{-3mm}
  \caption{
    The ablation study of bandwidth algorithms on HPOlib.
    The $x$-axis is the number of evaluations and the $y$-axis is the cumulative minimum objective.
    The solid lines in each figure show the mean of the cumulative minimum objective over all control parameter configurations.
    Note that the objective of HPOlib is the $\log$ of validation mean squared error.
    The transparent shades represent the distributions of the cumulative minimum objective at $\{50,100,150,200\}$ evaluations.
    The performance of Optuna v4.0.0 (brown dotted lines) is provided as a baseline.
  }
  \label{appx:experiments:fig:bandwidth-hpolib}
\end{figure}

\begin{figure}[p]
  \vspace{-10mm}
  \centering
  \includegraphics[width=0.98\textwidth]{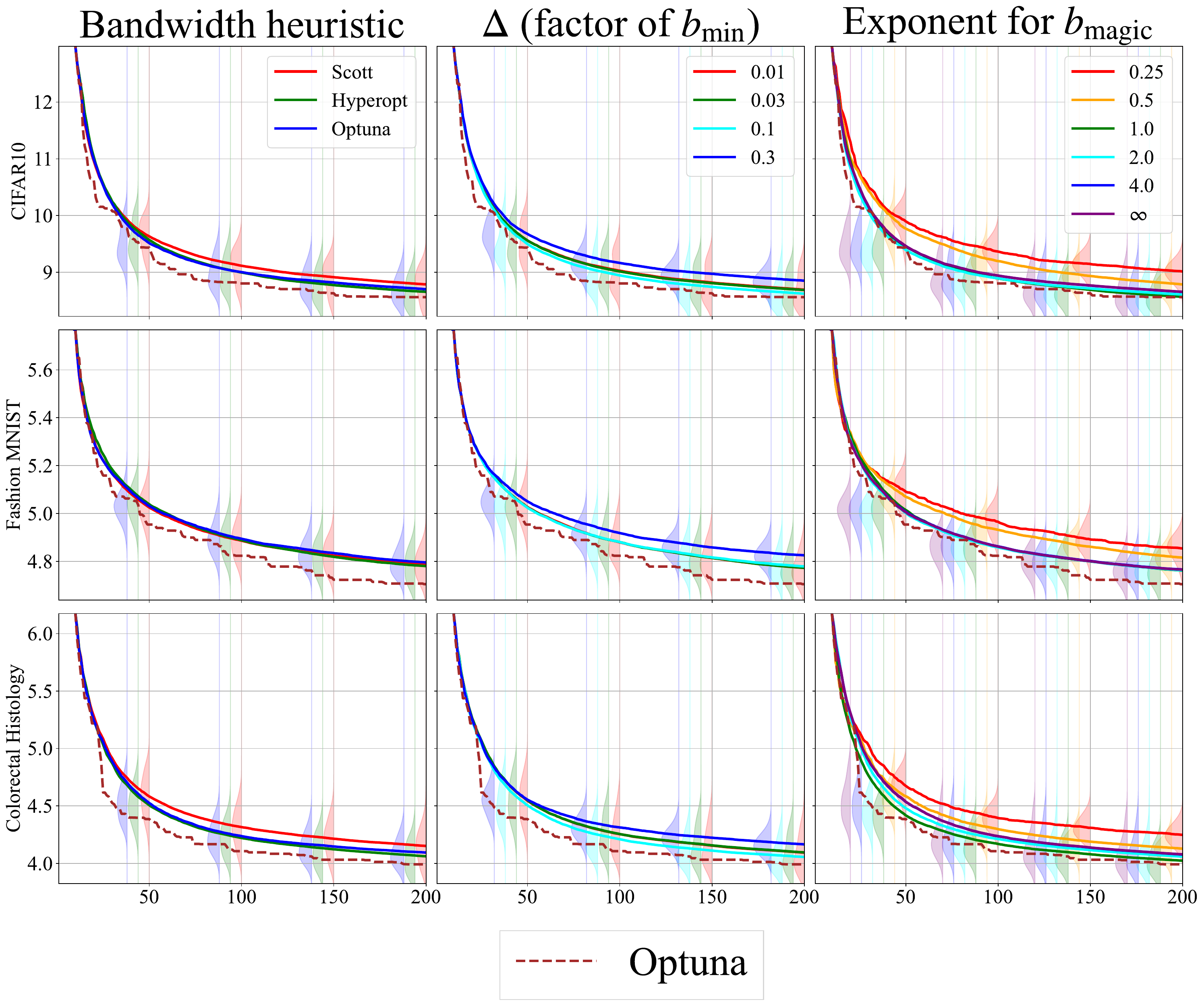}
  \vspace{-3mm}
  \caption{
    The ablation study of bandwidth algorithms on JAHS-Bench-201.
    The $x$-axis is the number of evaluations and the $y$-axis is the cumulative minimum objective.
    The solid lines in each figure show the mean of the cumulative minimum objective over all control parameter configurations.
    The transparent shades represent the distributions of the cumulative minimum objective at $\{50,100,150,200\}$ evaluations.
    The performance of Optuna v4.0.0 (brown dotted lines) is provided as a baseline.
  }
  \label{appx:experiments:fig:bandwidth-jahs}
\end{figure}

\begin{figure}[p]
  \centering
  \includegraphics[width=0.98\textwidth]{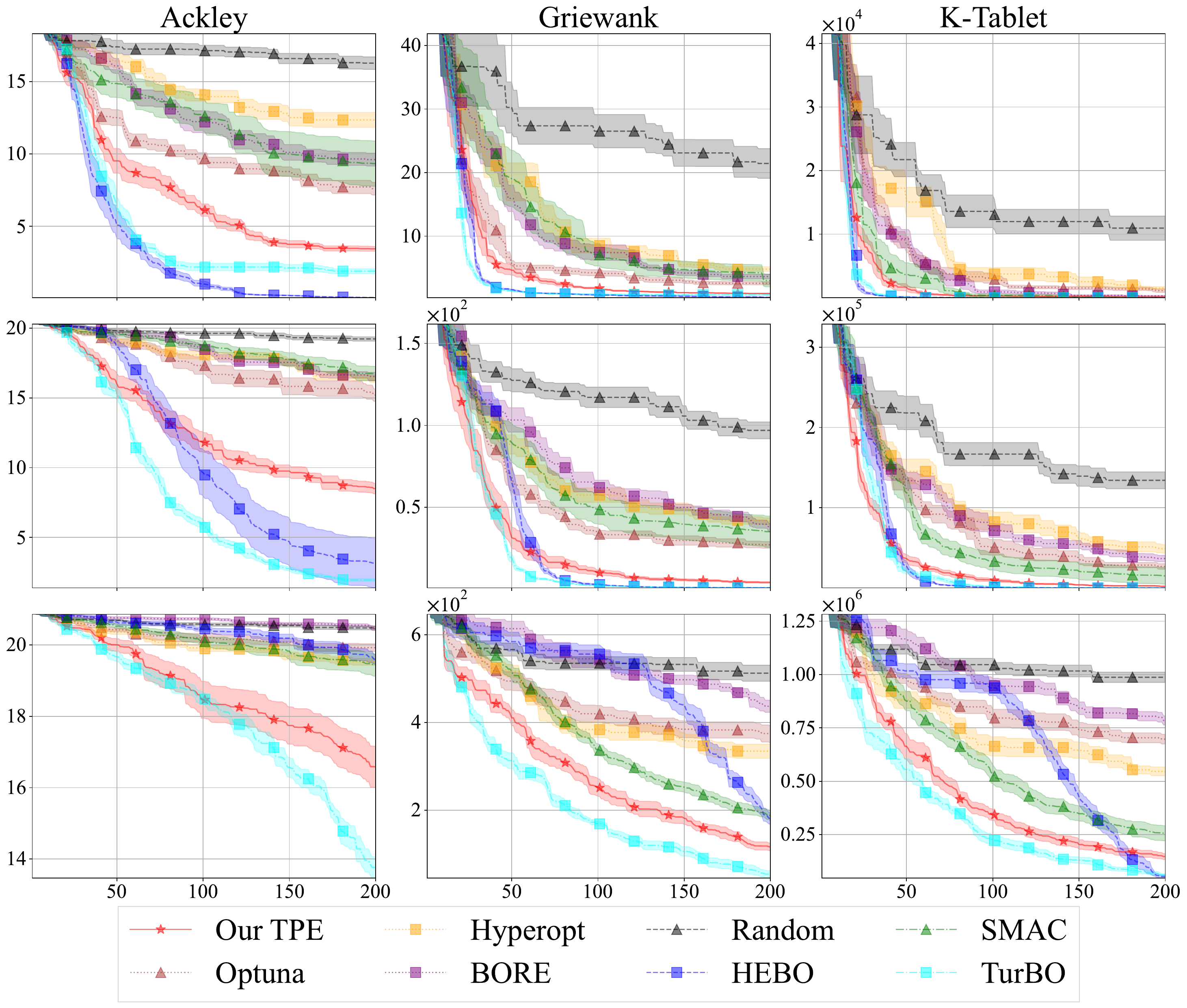}
  \caption{
    The comparison of optimization methods on benchmark functions.
    The $x$-axis is the number of evaluations and the $y$-axis is the cumulative minimum objective value.
    Each optimization method was run with $10$ different random seeds and the weak-color bands represent the standard error.
  }
  \label{appx:experiments:fig:comparison-bench0}
\end{figure}

\begin{figure}[p]
  \centering
  \includegraphics[width=0.98\textwidth]{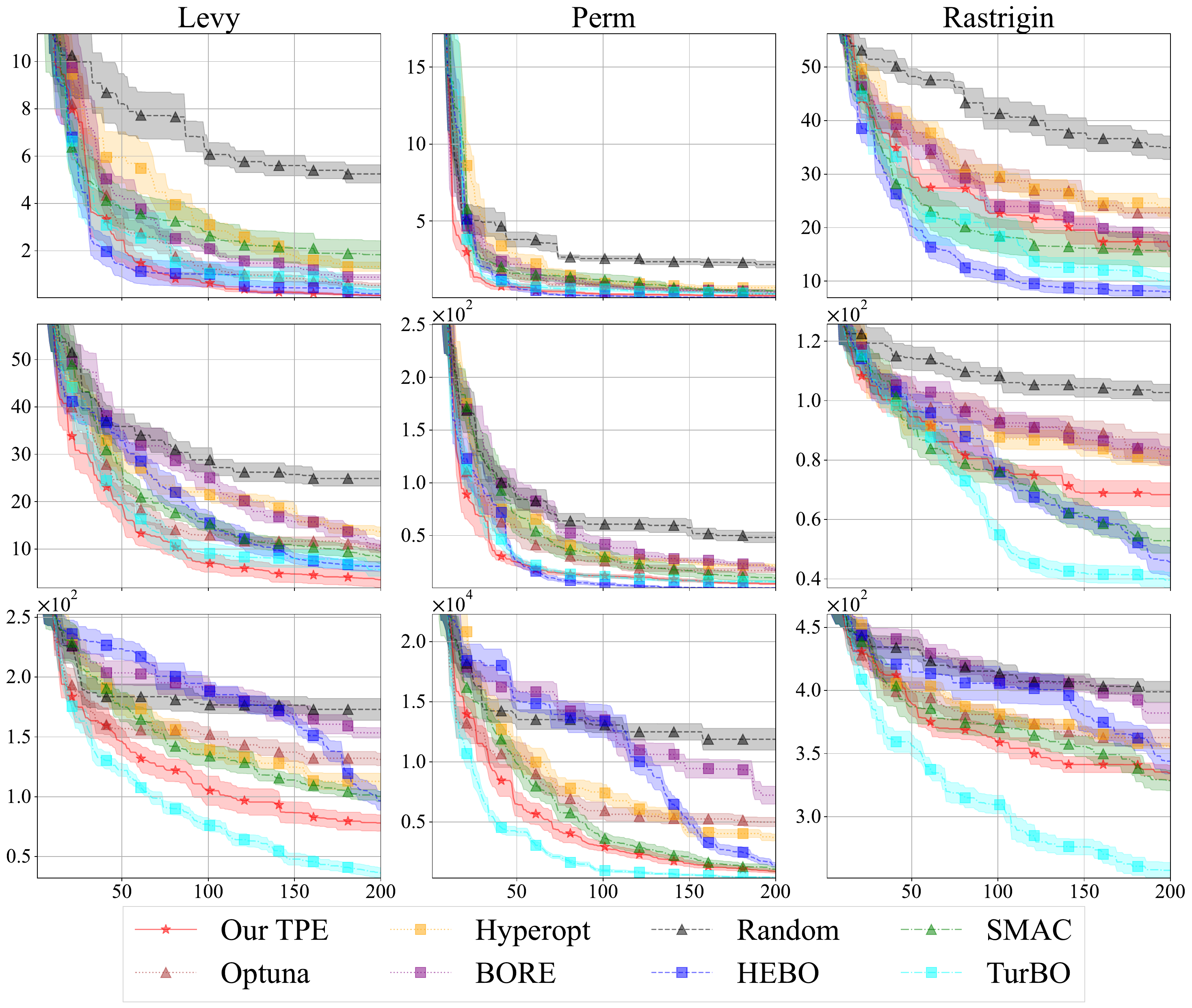}
  \caption{
    The comparison of optimization methods on benchmark functions.
    The $x$-axis is the number of evaluations and the $y$-axis is the cumulative minimum objective value.
    Each optimization method was run with $10$ different random seeds and the weak-color bands represent the standard error.
  }
  \label{appx:experiments:fig:comparison-bench1}
\end{figure}

\begin{figure}[p]
  \centering
  \includegraphics[width=0.98\textwidth]{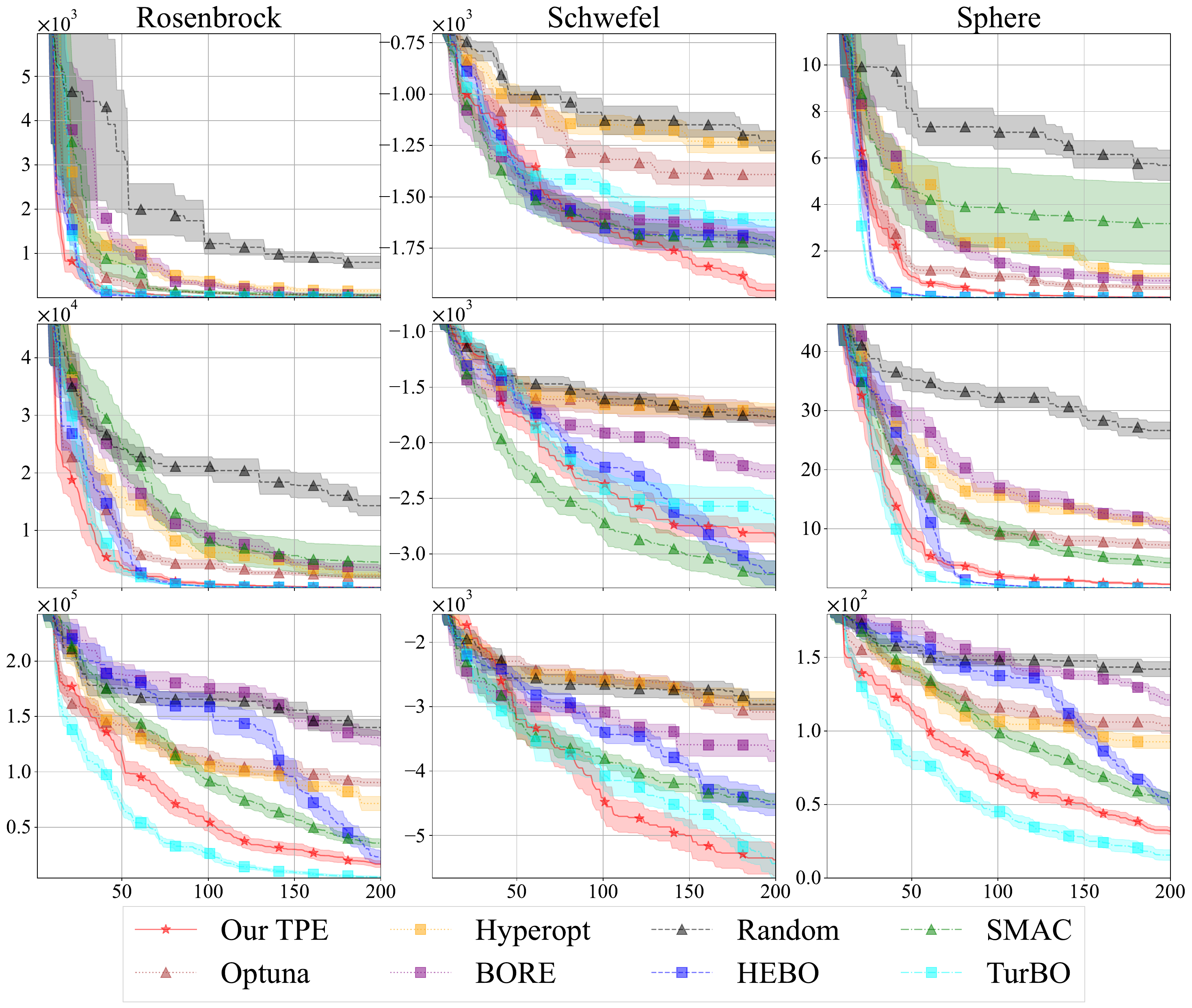}
  \caption{
    The comparison of optimization methods on benchmark functions.
    The $x$-axis is the number of evaluations and the $y$-axis is the cumulative minimum objective value.
    Each optimization method was run with $10$ different random seeds and the weak-color bands represent the standard error.
  }
  \label{appx:experiments:fig:comparison-bench2}
\end{figure}

\begin{figure}[p]
  \centering
  \includegraphics[width=0.98\textwidth]{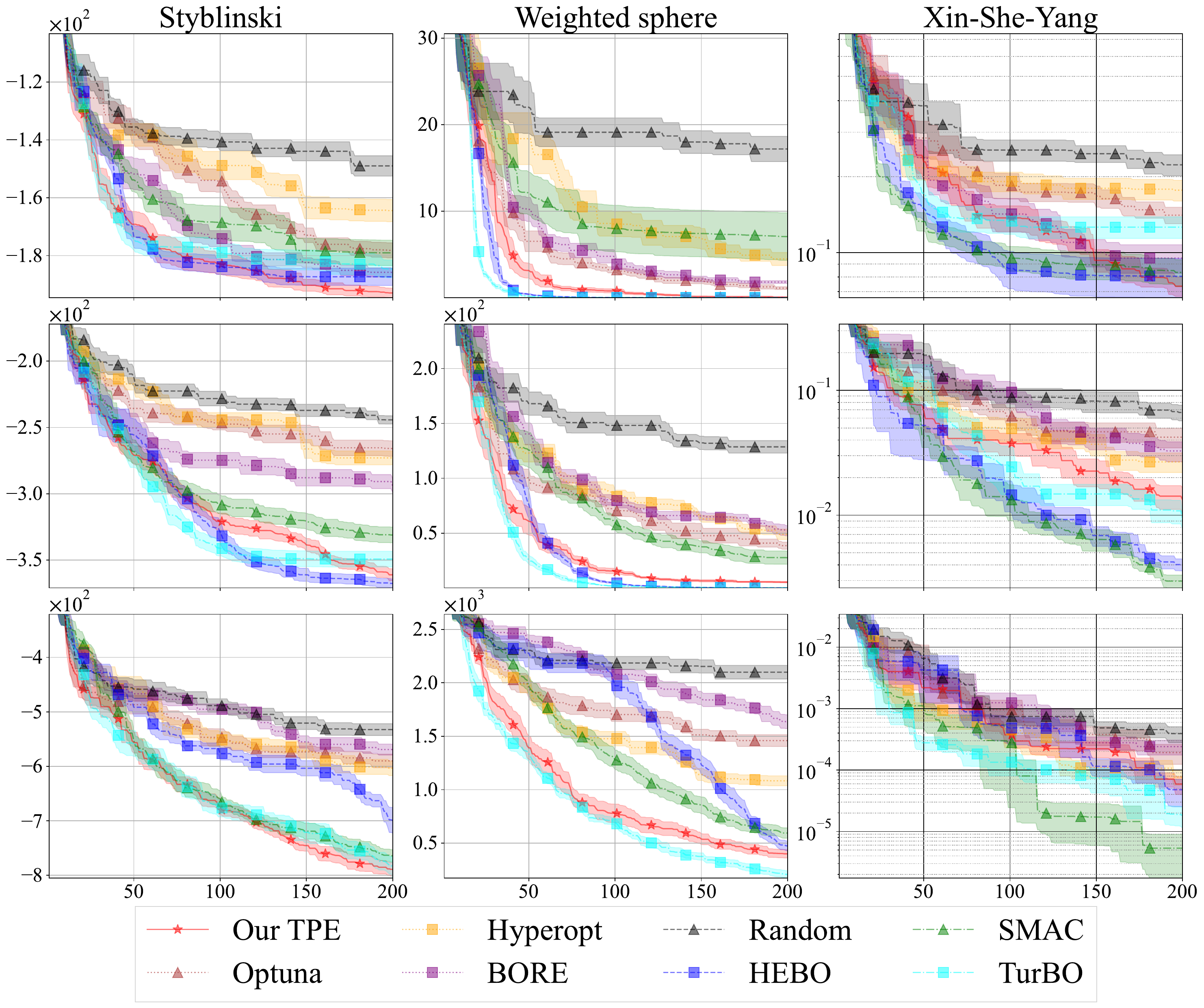}
  \caption{
    The comparison of optimization methods on benchmark functions.
    The $x$-axis is the number of evaluations and the $y$-axis is the cumulative minimum objective value.
    Each optimization method was run with $10$ different random seeds and the weak-color bands represent the standard error.
  }
  \label{appx:experiments:fig:comparison-bench3}
\end{figure}

\begin{figure}[p]
  \vspace{-10mm}
  \begin{center}
    \includegraphics[width=0.9\textwidth]{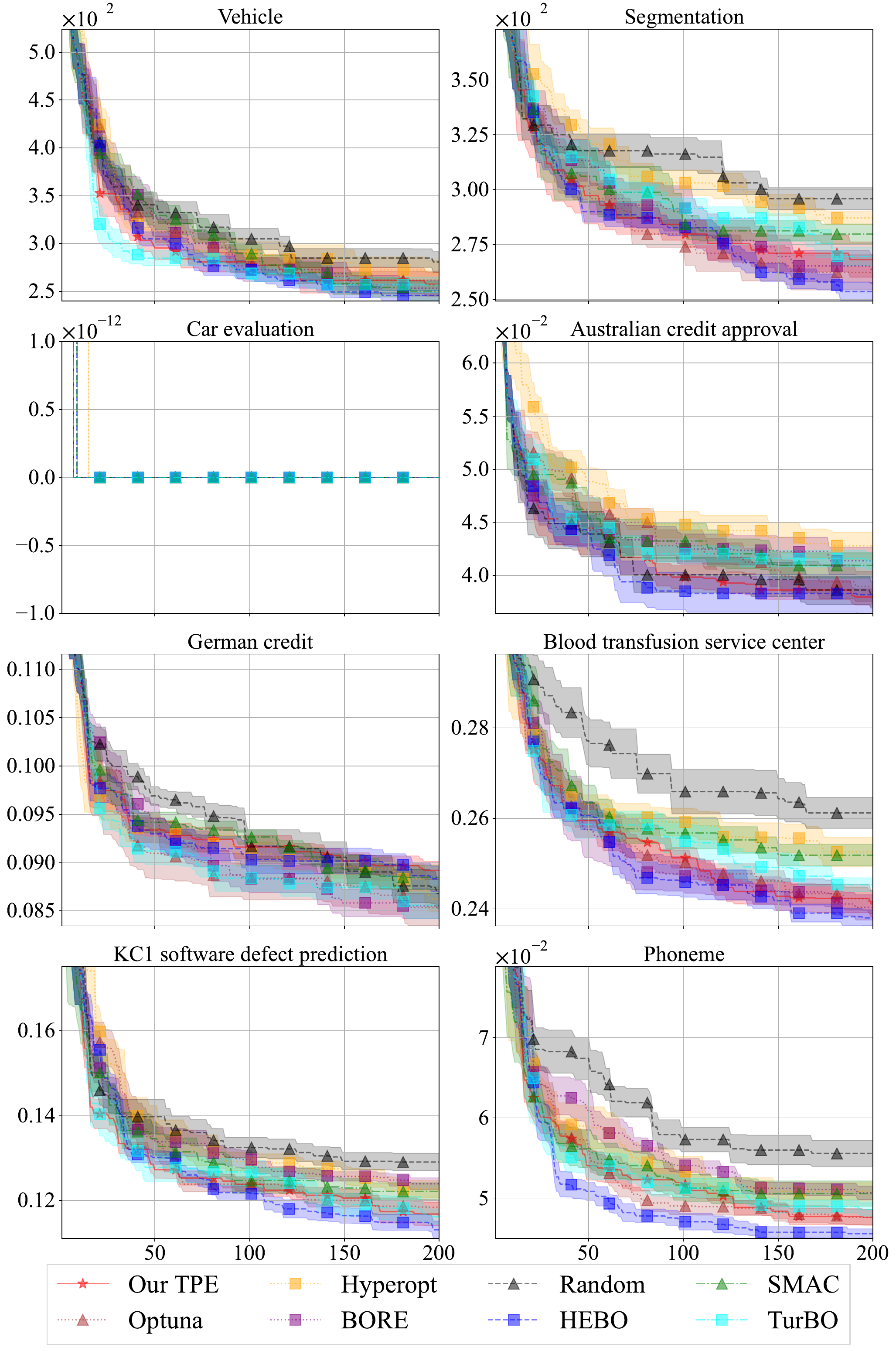}
    \vspace{-3mm}
    \caption{
      The comparison of optimization methods on HPOBench.
      The $x$-axis is the number of evaluations and the $y$-axis is the cumulative minimum objective value.
      Each optimization method was run with $10$ different random seeds and the weak-color bands represent the standard error.
    }
    \label{appx:experiments:fig:comparison-hpobench}
  \end{center}
\end{figure}

\begin{figure}[p]
  \vspace{-10mm}
  \begin{center}
    \subfloat[HPOlib\vspace{-3mm}]{
      \includegraphics[width=0.98\textwidth]{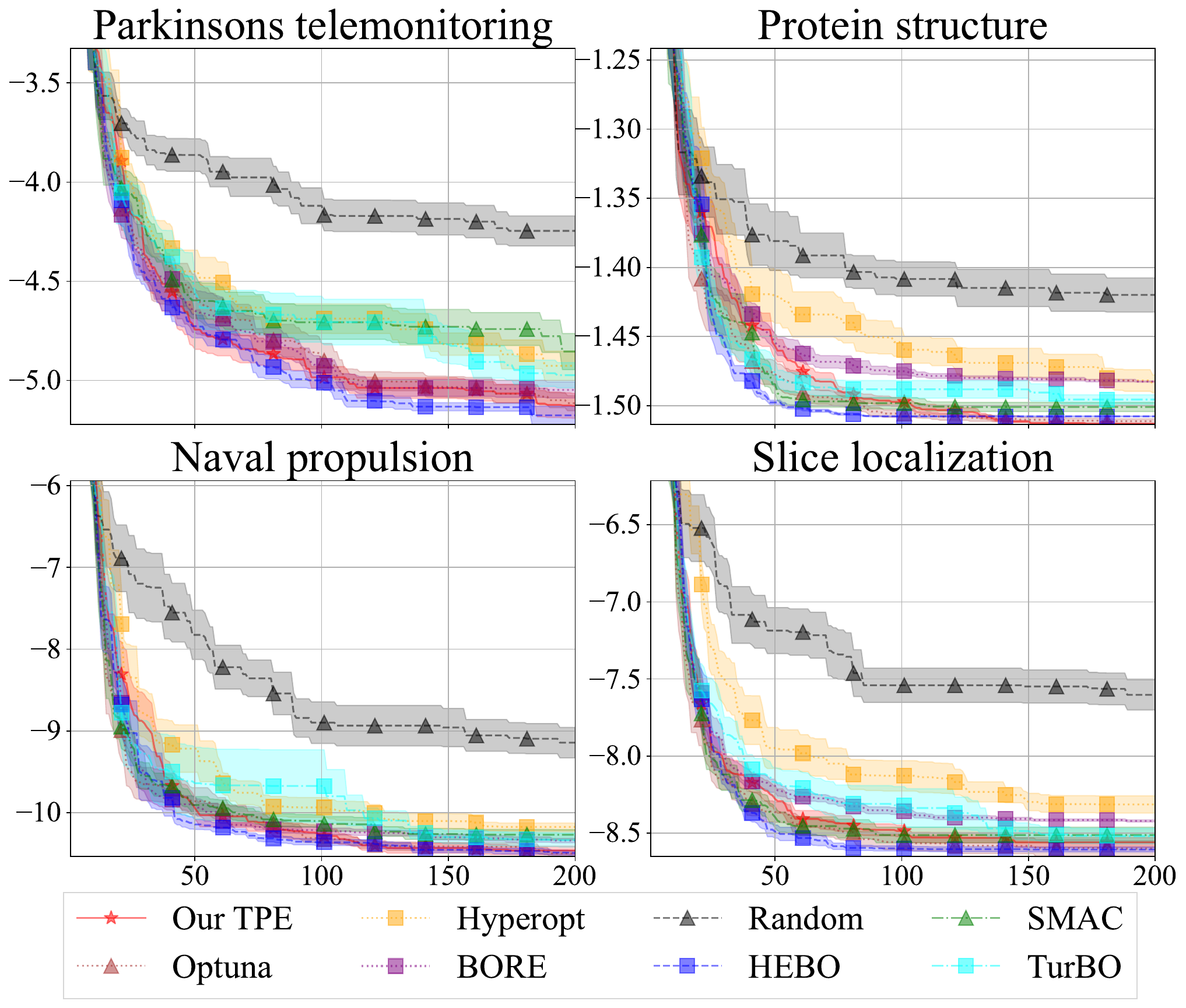}
    } \\
    \subfloat[JAHS-Bench-201\vspace{-3mm}]{
      \includegraphics[width=0.98\textwidth]{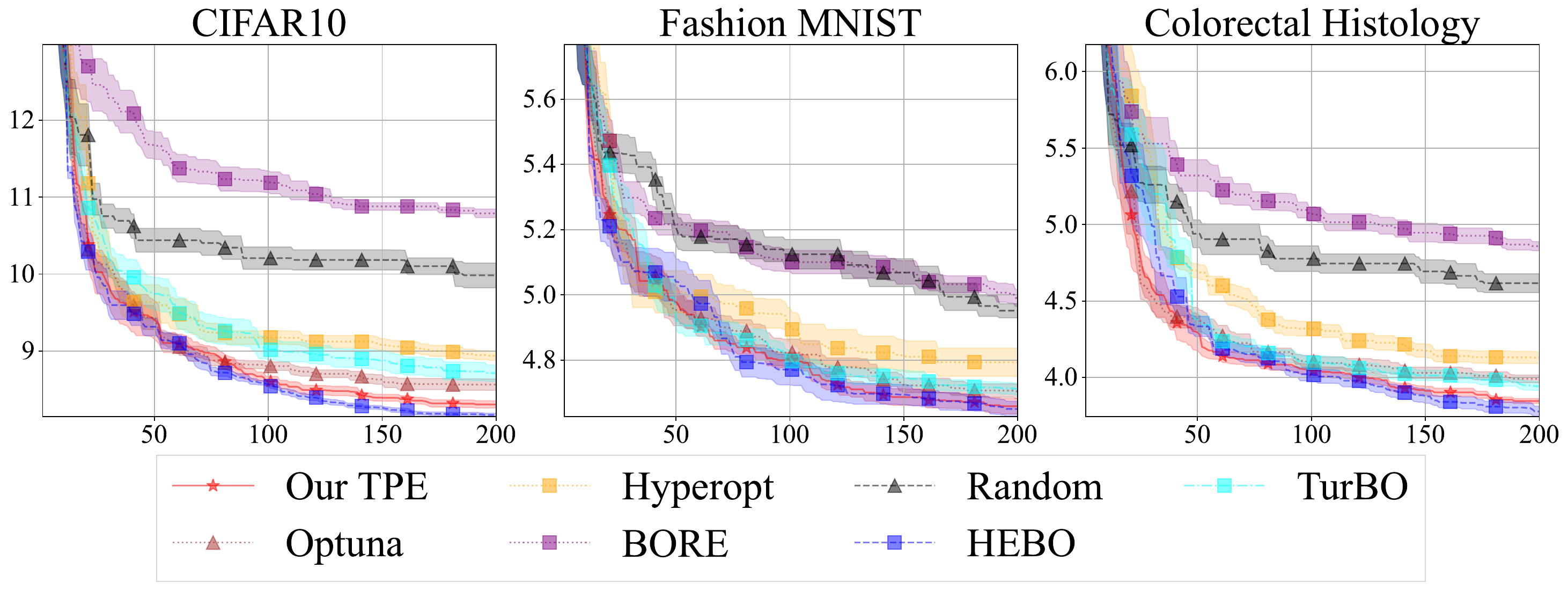}
    }
    \caption{
      The comparison of optimization methods on the HPO benchmarks.
      The $x$-axis is the number of evaluations and the $y$-axis is the cumulative minimum objective value (for HPOlib, we took the log-scale of validation MSE).
      Each optimization method was run with $10$ different random seeds and the weak-color bands represent the standard error.
      Note that \texttt{SMAC} are omitted for JAHS-Bench-201 due to the package dependency issue.
    }
    \label{appx:experiments:fig:comparison-hpolib-jahs}
  \end{center}
\end{figure}

\section{General Advice for Hyperparameter Optimization}
\label{appx:general-advice:section}
This section briefly discusses simple strategies to effectively design search spaces for HPO.
There are several tips to design search spaces well:
\begin{enumerate}
  \vspace{-1mm}
  \item reduce or bundle hyperparameters as much as possible,
  \vspace{-1mm}
  \item include strong baseline settings in initial configurations,
  \vspace{-1mm}
  \item use ordinal parameters instead of continuous parameters,
  \vspace{-1mm}
  \item consider other possible optimization algorithms,
  \vspace{-1mm}
  \item restart optimization after a certain number of evaluations.
  \vspace{-1mm}
\end{enumerate}
The first point is essential due to the curse of dimensionality in high dimensions.
For example, when hyperparameter configurations of neural networks are considered, it is simpler to bundle hyperparameters for each layer such as dropout rate and activation functions.
Such design significantly reduces the dimensionality.
The second point is to include strong baselines if available.
This follows the basic principle of warm-starting methods~\shortcite{feurer2015initializing,nomura2021warm,hvarfner2022pi}, which start near known strong baselines.
The third point is to define hyperparameters as ordinal parameters rather than continuous parameters according to intrinsic cardinality.
Recall that intrinsic cardinality is defined in Section~\ref{main:algorithm-detail:section:bandwidth}.
We illustrate an example below:
\begin{lstlisting}
low, high = 0.0, 1.0
# Definition as a continuous parameter
dropout = trial.suggest_float("dropout", low, high)

# Definition as an ordinal parameter
intrinsic_cardinality = 11
dropout_choices = np.linspace(
    low, high, intrinsic_cardinality
)
index = trial.suggest_int(
    "dropout-index", 0, intrinsic_cardinality - 1
)
dropout = dropout_choices[index]
\end{lstlisting}
Notice that the discretization is not preferred in some methods such as Gaussian process-based BO because they often optimize the acquisition function by gradient-based approaches.
The fourth point is algorithm selection.
If the search space contains only numerical parameters, a promising candidate would be CMA-ES~\shortcite{hansen2016cma} for large-budget settings and the Nelder-Mead method~\shortcite{nelder1965simplex} for small-budget settings.
On the other hand, if the search space contains categorical or conditional parameters, random search or BO becomes a strong candidate.
Note that \shortciteA{ozaki2022global} report that local search methods such as Nelder-Mead and CMA-ES consistently outperform global search methods.
Although \shortciteA{ozaki2022global} do not test TPE, TPE is a promising option considering its local search nature especially when the search space contains categorical or conditional parameters.
The fifth point is to restart optimizations.
Restarting is especially important for non-global methods such as TPE and Nelder-Mead method because the optimization often gets stuck in a local optimum, missing promising regions.

\section{Erratum}
\label{appx:erratum:section}
We state that the improvement-based weighting makes the acquisition function of TPE \texttt{EI}, and that the weighting of MOTPE is proven to be \texttt{EI}.
We found that these statements are not exact.
Let $p_{\mathrm{unif}}(\xv | \Dl)$ and $p_{\mathrm{imp}}(\xv | \Dl)$ be the densities of $\Dl$ estimated with uniform weights and with the improvement-based weights, respectively.
Then \shortciteA{song2022general} show that \texttt{EI} is proportional to
\begin{equation}
  \frac{p_{\mathrm{imp}}(\xv | \Dl)}{\gamma p_{\mathrm{unif}}(\xv | \Dl) + (1 - \gamma) p(\xv | \Dg)},
\end{equation}
whereas the density ratio $p_{\mathrm{imp}}(\xv | \Dl) / p(\xv | \Dg)$ only approximates it.
The same applies to MOTPE where its weighting based on the hypervolume improvement approximates.
Its acquisition function only approximates expected hypervolume improvement.
Note that \texttt{EI} in the weighting strategies (e.g., Table~\ref{main:algorithm-detail:tab:benefits-drawbacks-of-weighting-strategy}) refers to the improvement-based weighting and not to the exact \texttt{EI}.


\else

\customlabel{data1}{1}
\customlabel{data2}{2}

\fi

\clearpage
\bibliographystyle{theapa}  
\bibliography{ref}

\end{document}